\documentclass[conference]{IEEEtran}
\usepackage{setspace}
\usepackage{amsmath}
\usepackage{acronym}
\usepackage{bbm}
\usepackage[dvips]{color}
\usepackage{comment}
\usepackage{todonotes}
\usepackage{epsf}
\usepackage{siunitx}
\usepackage{epsfig}
\usepackage{enumitem}
\usepackage{times}
\usepackage{epsfig}
\usepackage{graphicx}
\usepackage{bbold}
\usepackage{geometry}
\usepackage{mathrsfs}
\usepackage{amssymb}
\usepackage{pdfpages}
\usepackage{epstopdf}
\usepackage{tcolorbox,bbold}
\usepackage{algorithm,algorithmicx,algpseudocode}
\makeatletter
\newcommand{\algmargin}{\the\ALG@thistlm}
\makeatother
\algnewcommand{\parState}[1]{\State%
  \parbox[t]{\dimexpr\linewidth-\algmargin}{\strut #1\strut}}

\usepackage[nomain,acronym,shortcuts]{glossaries}
\newcommand*{\acro}[3][]{\newacronym[#1]{#2}{#2}{#3}}
\glsdisablehyper
\acro{D2D}{device-to-device}
\acro{SIR}{signal-to-interference-ratio}
\acro{SINR}{signal-to-interference-plus-noise-ratio}
\acro{PCP}{Poisson cluster process}
\acro{CoMP}{coordinated multi-point}
\acro{BS}{base station} 
\acro{MD-CoMP}{macrodiversity CoMP transmission}
\acro{MAC}{medium-access-control}
\acro{JT-CoMP}{joint transmission CoMP}
\acro{CoMP-JT}{coordinated multipoint joint transmission}
\acro{SBS}{small base station}
\acro{MDSD}{multiple devices to a single device}
\acro{MDS}{maximum distance separable}
\acro{SCN}{small cell network}
\acro{PPP}{Poisson point process}
\acro{TCP}{Thomas cluster process}
\acro{CSI}{channel state information}
\acro{PDF}{probability distribution function}
\acro{PMF}{probability mass function}
\acro{RV}{random variable}
\acro{i.i.d}{independently and identically distributed}
\acro{MBMS}{multimedia broadcasting multicasting service}
\acro{EE}{energy efficiency}
\acro{HCP}{hard-core placement}
\acro{CCDF}{complementary cumulative distribution function}
\acro{CDF}{cumulative distribution function}
\acro{PC}{probabilistic caching}
\acro{RC}{random caching}
\acro{CPF}{caching popular files} 
\acro{PGFL}{probability generating functional}
\acro{KKT}{Karush-Kuhn-Tucker}
\acro{PGF}{point generating function}
\acro{SCA}{successive convex approximation}
\acro{HD}{high-definition}
\acro{FHD}{full-high-definition}
\acro{UHD}{ultra-high-definition}
\acro{VR}{virtual reality}
\acro{AR}{augmented reality}
\acro{5G}{fifth-generation}
\acro{QoS}{quality-of-service}
\acro{QoE}{quality-of-experience}
\acro{IoT}{Internet of Things}
\acro{MHCPP}{Matern hardcore point process}
\acro{LoS}{line-of-sight}
\acro{NLoS}{non-line-of-sight}
\acro{PSD}{power spectral density}
\acro{MEC}{mobile edge computing}
\acro{E2E}{end-to-end}
\acro{THz}{terahertz}
\acro{CLT}{central limit theorem}
\acro{HQ}{High Quality}
\acro{eMBB}{enhanced mobile broadband}
\acro{URLLC}{ultra reliable low latency communications}
\acro{mmWave}{millimeter wave}
\acro{EVT}{extreme value theory}
\acro{GEV}{generalized extreme value}
\acro{HR2LLC}{high-rate and high-reliability low latency communications}
\acro{EHF}{extremely high frequency}
\acro{TVaR}{tail-value-at-risk}
\acro{UE}{user equipment}
\acro{RIS}{reconfigurable intelligent surfaces}
\acro{MIMO}{multiple-input and multiple-output}
\acro{HMD}{head-mounted display}
\acro{AI}{artificial intelligence}
\acro{ML}{machine learning}
\acro{HITRAN}{high resolution transmission}
\acro{OFDM}{orthogonal frequency division multiplexing}
\acro{PAPR}{peak-to-average power ratio}
\acro{MR}{mixed reality}
\acro{EVaR}{entropic value-at-risk}
\acro{DNN}{deep neural network}
\acro{MDP}{Markov decision process}
\acro{RL}{reinforcement learning}
\acro{RNN}{recurrent neural network}
\acro{ANN}{artificial neural networks}
\acro{LSTM}{long short-term memory}
\acro{ReLu}{rectified linear unit}
\acro{VaR}{value-at-risk}
\acro{SNR}{signal-to-noise ratio}
\acro{RF}{radio frequency}
\acro{CVaR}{conditional value-at-risk}
\acro{AVaR}{average value-at-risk}
\acro{ES}{expected shortfall}
\acro{AoI}{age of information}
\acro{PAoI}{peak age of information}
\acro{QoPE}{quality of physical experience}
\acro{LCFS}{last come first served}
\acro{FCFS}{first come first served}
\acro{SC-FDM}{single carrier frequency-division multiplexing}
\acro{ToA}{time of arrival}
\acro{MUSIC}{multiple signal classification}
\acro{IoE}{Internet of Everything}
\acro{6DoF}{six degrees of freedom}
\acro{OFDMA}{orthogonal frequency-division multiple access}
\acro{AoSA}{array of subarray}
\acro{XR}{extended reality}
\acro{AoA}{angle of arrival}
\acro{ULA}{uniform linear array}
\acro{AoD}{angle of departure}
\acro{EM}{electromagnetic}
\acro{IoI}{Internet of Intelligence}
\acro{MISO}{multiple-input and single-output}
\acro{CRAS}{connected robotics and autonomous systems}
\acro{FL}{federated learning}
\acro{GAN}{generative adversarial network}
\acro{mMTC}{massive machine type communications}
\acro{Tbps}{terabit per second}
\acro{NTN}{non-terrestrial network}
\acro{SLAM}{simultaneous localization and mapping}
\acro{IS}{information shower}
\acro{OAM}{orbital angular momentum}
\acro{NOMA}{non-orthogonal mutliple access}
\acro{CPS}{cyber physical system}
\acro{IoNT}{Internet of Nano-Things}
\acro{TDD}{time division duplex}
\acro{CDMA}{code division multiple access}
\acro{OMA}{orthogonal multiple access}
\acro{VLC}{visible light communications}
\acro{UAV}{unmanned aerial vehicle}
\acro{LEO}{low earth orbit}
\acro{DL}{deep learning}
\acro{KPI}{key performance indicator}
\acro{NLP}{natural language processing}
\acro{SCM}{structural causal model}
\acro{O-RAN}{open-radio access network}
\acro{RIC}{RAN intelligent controller}
\acro{RT}{real-time}
\acro{O-RU}{open radio unit}
\acro{O-DU}{open distributed unit}
\acro{CU}{centralized unit}
\acro{OSI}{open systems interconnection}

\usepackage{amssymb}
\usepackage{url}
\usepackage{dsfont}
\usepackage{lettrine} 
\usepackage{cite}
\usepackage{amsmath,epsfig,amssymb,algorithm,algpseudocode,amsthm,cite,url}
\IEEEoverridecommandlockouts
\usepackage{subcaption}
\allowdisplaybreaks
\usepackage{csquotes}

\usepackage{verbatim}
\usepackage[english]{babel}
\usepackage{amsmath,amssymb}

\usepackage[justification=centering]{caption}
\usepackage{verbatim}

\newtheorem{proposition}{\bf Proposition}

\newtheorem{definition}{\bf Definition}

\IEEEoverridecommandlockouts
\begin{document}
	\title{Less Data, More Knowledge: Building Next Generation Semantic Communication Networks\vspace{-0.35cm}}
		\author{Christina Chaccour, \emph{Graduate Student Member, IEEE,} Walid Saad, \emph{Fellow, IEEE,} \\   M\'erouane Debbah,  \emph{Fellow, IEEE}, Zhu Han, \emph{Fellow, IEEE},
		and H. Vincent Poor, \emph{Life Fellow, IEEE}   
	\thanks{C. Chaccour and W. Saad are with Wireless@VT, Bradley Department of Electrical and Computer Engineering, Virginia Tech, Arlington, VA, USA, Emails: \protect{christinac@vt.edu}, \protect{walids@vt.edu}.}
	\thanks{ M. Debbah is with the Technology Innovation Institute, 9639 Masdar City, Abu Dhabi, United Arab Emirates and also with CentraleSupelec, University Paris-Saclay, 91192 Gif-sur-Yvette, France, Email: \protect{merouane.debbah@tii.ae}.}
	\thanks{Z. Han is with the Department of Electrical and Computer Engineering, University of Houston, Houston, TX 77004 USA, and also with the Department of Computer Science and Engineering, Kyung Hee University, Seoul 446-701, South Korea.}
	\thanks{H.V. Poor is with Department of Electrical and Computer Engineering, Princeton University, NJ 08544, USA, \protect{poor@princeton.edu}.}}
\maketitle
\vspace{-0.5cm}
\begin{abstract}
 Semantic communication is viewed as a revolutionary paradigm that can potentially transform how we design and operate wireless communication systems. However, despite a recent surge of research activities in this area, remarkably, the research landscape is still limited in at least three ways. First, the very definition of a ``semantic communication system'' remains ambiguous, and it differs from one work to another. Second, there is a lack of \emph{fundamental and scalable} frameworks for building next-generation semantic communication networks based on rigorous and well-defined technical foundations. Third, the question of what a ``semantic representation’’ means, and on how this representation can be used to instill meaning, significance, and structure to every information transfer over a wireless network remain unanswered. In this tutorial, we present the first rigorous and holistic vision of an end-to-end semantic communication network that is founded on novel concepts from artificial intelligence (AI), causal reasoning, transfer learning, and minimum description length theory. We first discuss how the design of semantic communication networks requires a move from \emph{data-driven} and \emph{information-driven} AI-augmented networks, in which wireless networks remain ``tied" to data, towards \emph{knowledge-driven} and \emph{reasoning-driven} AI-native networks in which wireless networks are AI-native and can perform versatile logic. We then distinguish the concept of semantic communications from several other approaches that have been conflated with it. For instance, we opine that effectively and efficiently building next-generation semantic communication networks must go beyond: a) creating a new type of encoder and decoder at the transmitter/receiver side, and b) designing a new ``AI for wireless’’ framework in which AI is used to extract some application features or to fine tune a wireless protocol or algorithm. Then, we identify the main tenets that are needed to build an end-to-end semantic communication network. Among those building blocks of a semantic communication networks, we highlight the necessity of creating semantic representations of data that satisfy the key properties of \emph{minimalism, generalizability, and efficiency} so as to faithfully represent the data and enable the transmitter and receiver to do more with less, i.e., computationally generate content via a minimally semantic representation. We then explain how those representations can form the basis a so-called \emph{semantic language} that will allow a transmitter and receiver to communicate at a semantic level. In this regard, we distinguish the concept of a semantic language from that of a natural language, and we present the pillars needed to gradually build a semantic language with fundamental structural content, yet tolerable complexity. We then show that, by using semantic representation and languages, the traditional transmitter and receiver now become a teacher and apprentice. The teacher can identify the semantic content elements in the raw datastream and learn its semantic representation. The apprentice can reason over a semantic representation, map its corresponding semantic content element, and further draw logical conclusions based on the cumulative knowledge base built. This phenomenon mimics the growth of a child’s language’s expressivity and reasoning in a more-or-less parallel fashion. We then concretely define the concept of \emph{reasoning} by investigating the fundamentals of causal representation learning and their role in designing reasoning-driven semantic communication networks. We particularly demonstrate that reasoning faculties are majorly characterized by the ability to capture causal and associational relationships in datastreams. This enables radio nodes to communicate minimal, generalizable, and efficient semantic representations, and ultimately perform versatile logical conclusions – doing more with less. For such reasoning-driven networks, we revisit the fundamentals of information theory, in order to emphasize the concepts that must be redefined to capture semantic reasoning. We then propose novel and essential semantic communication \emph{key performance indicators (KPIs)} and metrics that include new ``reasoning capacity’’ measures that could go beyond Shannon’s bound to capture the imminent convergence of computing and communication resources. Finally, we explain how semantic communications can be scaled to large-scale networks such as cellular networks (6G and beyond), and deployed in emerging environments such as open radio access networks (O-RAN). In a nutshell, we expect this tutorial to provide a unified and self-contained reference on how to properly build, design, analyze, and deploy next-generation semantic communication networks. 
\end{abstract}
\section{Introduction}\label{sec:Intro}
Future wireless systems, namely 6G systems and beyond, must cater to the complex and stringent requirements of emerging applications such as the metaverse, holographic teleportation, digital twins, and Industry 5.0 \cite{chaccour2021seven}. Nonetheless, delivering a disruptive leap in wireless technologies cannot be fulfilled by continuing to pursue incremental advances to conventional wireless system components such as spectrum and multi-antenna technologies. Instead, it is necessary to rethink the way in which the entire wireless system architecture and functions are designed and operated. Along those line, current 5G and 6G research efforts, have already demonstrated the efficiency of using AI-driven augmentation in addressing various challenges throughout the network layer stack. For example, in \cite{soltani2019deep, shlezinger2020viterbinet, cousik2021fast}, it was reported that \ac{AI} and \ac{ML} can provide robust, accurate, and reduced complexity for tasks like channel estimation, initialization, and symbol detection. Also, \ac{ML} has played an integral role routing protocol design \cite{li2020routing}, resource management \cite{jayaprakash2021systematic}, and network management \cite{ayoubi2018machine}, among many other specific wireless problems and protocols\cite{eldar2022machine}.\\
  	\begin{figure*} [t!]
	\begin{centering}
		\includegraphics[width=0.95\textwidth]{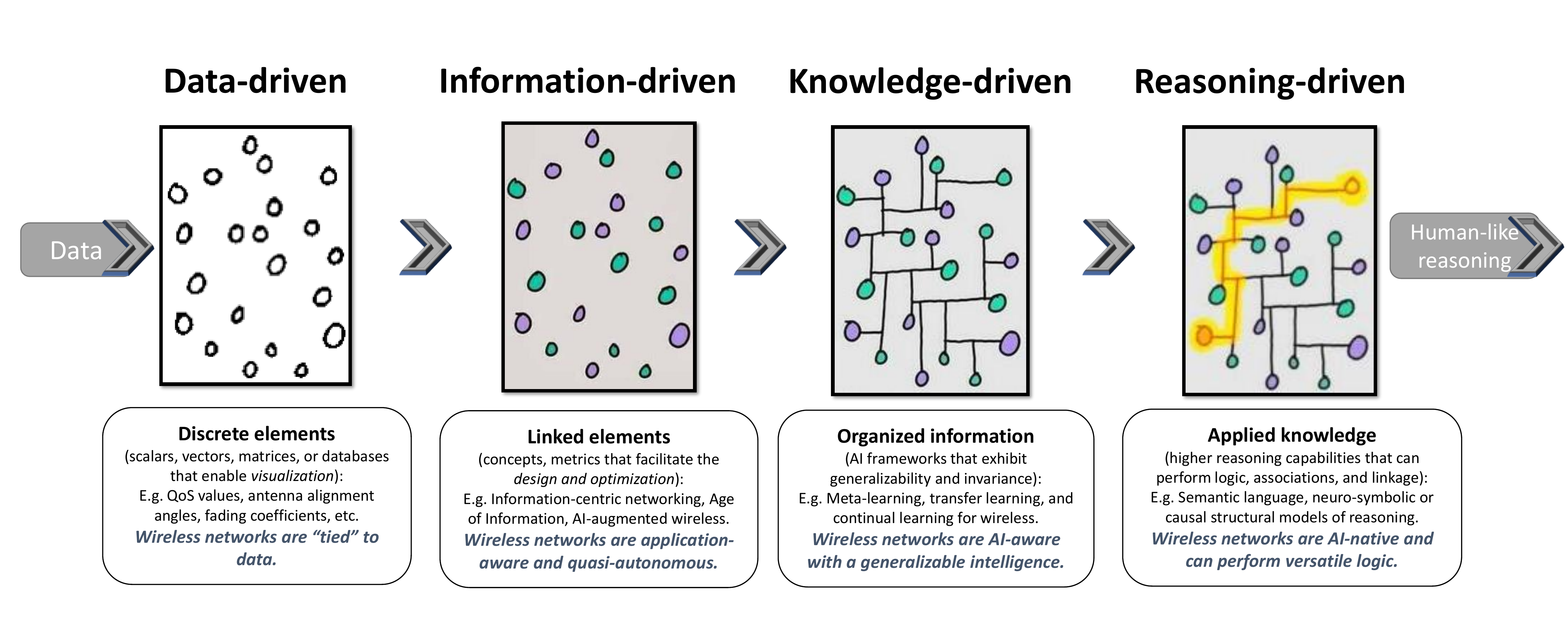}
		\caption{\small{Illustrative figure showcasing the evolution of wireless networks from data-driven ones towards reasoning-based ones.}}
		\label{fig:datatoknowledge}
	\end{centering}
\end{figure*}
\indent However, current AI-based wireless approaches~\cite{soltani2019deep, shlezinger2020viterbinet, cousik2021fast, li2020routing, jayaprakash2021systematic, ayoubi2018machine, eldar2022machine} remain limited in a number of ways. First, in the state-of-art, the design of wireless networks and protocols is either limited to \emph{data-driven solutions} or to \emph{information-driven approaches}, and, thus, it fails to \emph{leverage knowledge} accumulated throughout the operation of the system, as shown in Fig.~\ref{fig:datatoknowledge}. For instance, under the data-driven paradigm, wireless networks rely on discrete elements, (e.g., spectrum data, channel data, \ac{QoS} values) to fine-tune their operation. Here, \emph{wireless networks designs, and their performance will be tied very closely to data}. Meanwhile, in information-based approaches, wireless networks leverage \emph{information-centric} metrics, such as \ac{AoI}, value-of-information, application data, and reliability to make more informed network decisions (which links multiple data points into an important performance-evaluating metric). Nonetheless, both of these approaches fail to leverage, accumulate, and organize \emph{knowledge within the data and information sets}. Essentially, the decision making performed via these two paradigms remains largely training dependent, and exhibits \emph{limited generalizability}. In contrast, under our \emph{envisioned knowledge-driven} and \emph{reasoning-driven} (see Fig.~\ref{fig:datatoknowledge}) paradigms, the wireless network will be able to proactively make decisions and draw conclusions on its own based on the logical intent that can be extracted from built knowledge bases. In this case, the network will be able to acquire \emph{more knowledge} with \emph{less data}, compared to data/information-driven approaches. This, in turn, will facilitate the network's ability to achieve high-rate, low latency, and high-reliability, thereby becoming more adept in meeting the very stringent \ac{QoS} requirements of future 6G and beyond applications. In addition, by continuously exploiting accumulated knowledge, the network can now reduce reliance on the brittle spectral resources, enhance the way in which data is transmitted and recovered, and rely on computational resources to yield semantic content rather than reconstruct raw data. We envision that these two paradigms must be a stepping stone of the emerging concept of \emph{AI-native} wireless systems, that has attracted significant attention by academia, industry, and standardization bodies.\\
\indent Broadly, the concept of \ac{AI}-native systems envisions building the entire protocol stack and air-interface of a wireless system using \ac{AI} techniques. For instance, in~\cite{hoydis2021toward} a new research direction is attempting to transforming the air-interface into a full \ac{AI}-\ac{AI} integration, to reach the milestone of \ac{AI}-nativeness. However, existing works~\cite{hoydis2021toward, wu2022ai, carbone2021neuroran} in this regard, do not specify what type of \ac{AI} framework should be used to build such systems, but they still hint towards classical techniques (e.g., convolutional neural networks, reinforcement learning, etc). In this regard, we opine that \ac{AI}-native networks cannot continue to rely on mere \ac{AI}-augmented techniques, such as designing a transceiver via autoencoders or resource management protocols via deep \ac{RL}~\cite{soltani2019deep, shlezinger2020viterbinet, cousik2021fast, li2020routing, jayaprakash2021systematic, ayoubi2018machine, eldar2022machine}. In contrast, instead of \emph{augmenting} or merely \emph{replacing} existing network layers and components with \ac{AI} to achieve \ac{AI}-nativeness, we envision that, ultimately, \ac{AI}-native wireless systems should be intrinsically designed and structured based on \emph{applied knowledge}. In our envisioned \emph{reasoning-driven AI native wireless systems}, the \ac{E2E} design and operation of the network will be designed using next-generation reasoning \ac{AI} frameworks that can exploit causality and stochasticity in the data, identify structure, deduce logical connections, and extract (and mitigate) semantic noise (the source, root-cause, and mechanism of the noise are identified). In this scenario, the wireless system becomes a \emph{living, sustainable network} that can grow with its cumulative knowledge in order to execute operations that cannot simply be done when being overly reliant on existing data and re-training mechanisms. Thus, this opens the door to go beyond the use of ad-hoc \ac{AI}-augmentation techniques as is the case in today's data-driven and information-driven networks. This, in turn, poses a fundamental question: \emph{``How can we create reasoning-driven AI-native systems?''}. Our proposed reasoning-driven AI-native systems can thereby use less data, and more knowledge, in order to perform the various functions of the E2E wireless system. \\
\indent One key challenge that must be addressed when answering the aforementioned question is the need to fundamentally transform the way in which data is viewed, processed, transmitted, recovered, and exploited at the level of the a network's transmitter in receiver. In other words, creating reasoning-driven AI wireless systems must challenge the classical assumption that the wireless network's transmitter and receiver (even those designed with AI) are simple ``bit pipes'' that act as a simple conduits of data bits, without exploiting a knowledge base that is built on the structure, linkages, and relationships between multiple low-level data points. Indeed, somewhat remarkably, despite all the effort directed towards the \ac{AI}-nativeness of future wireless networks, at the level of transmitter and receiver, we still rely on very classical message construction and recovery mechanisms. One could argue that recent \ac{AI}-based transmitter and receiver designs (e.g., using autoencoders \cite{jakobauto}) are efficient in jointly learning transmitter and receiver implementations as well as signal encodings without any prior knowledge. However, although such mechanisms adopt deep learning to learn the transmitter, channel, and receiver, these learned building blocks fundamentally still perform the same classical information transmission tasks. In other words, the message construction and recovery mechanisms are \emph{``learned"} to combat the channel's uncertainty. Moreover, despite advances in \ac{AI}, the operational functionality of communication systems remains tied to data, dependent on classical \acp{ANN}, with weak generalizability and reasoning abilities. Such frameworks fail to \emph{organize information by characterizing the causal and associational characteristics of data} (See Fig.~\ref{fig:datatoknowledge}), thus, preventing the network from gaining knowledge that can be exploited in the future to operate with less input data. As a result, such \emph{knowledge-agnostic} \ac{AI}-native wireless network design approaches have a limited evolutionary potential and must re-engineered to mimic human reasoning, if we are to make a fundamental leap in future wireless technologies (6G and beyond).\\
\indent Next, we discuss why the transmitter, receiver, and air-interface have retained their classical Shannon functionality to this date. Then, why the path towards \ac{AI}-native, reasoning-driven networks requires a major leap from traditional communications to semantic communications, while leveraging the next wave of \ac{AI} frameworks.\\
\subsection{Why now? Why have we continued to rely on traditional communications so far?}
\indent Since its inception, the digital communication system problem, as posed by Shannon, has been a \emph{reconstruction problem} because of the paucity in computing capabilities needed for more intelligent \ac{AI}-guided tasks. That is, the fundamental goal of communications has hitherto been viewed as the capability of reproducing at one point either exactly or approximately a message transmitted from another point. Thus, the transmitter and the receiver have traditionally been designed in a fashion that relies solely on compression, transmission, and decompression. Moreover, the techniques used to encode the message only characterize the stochasticity stemming from the source, channel, and destination. For instance, the encoding performed at the physical layer characterizes the stochasticity at the transmitter, meanwhile, such encoding does not represent the characteristics of the message conveyed, nor its context, i.e., the encoding does not contain any information related to the significance or meaning of the message. When augmented with AI, such approaches remain confined to data-driven designs. \\
\indent In this conventional setting, one may not be able to efficiently convey the desired meaning of the transmitted messages (at the levels of the transmitter and receiver), thus leading to multiple repercussions on the overall \acrfull{E2E} communication system design. \emph{First}, in many use-cases continual repetitive, back-and-forth transmissions are needed to transfer the necessary application information. That is, in the standard communication setting, the transmitter does not attempt to \emph{automate the generation} of the message at the receiver. Also, the transmitter and receiver do not typically leverage their \emph{memories}, i.e., the history of previous message transmissions/observations or observed past patterns in the data. This knowledge, if exploited properly, i.e., if the structure within historical message transmissions is \emph{learned}, then the receiver can generate such a structure rather than \emph{continuously reconstruct it} (see knowledge-driven case in Fig.~\ref{fig:datatoknowledge}). \emph{Second}, in the current communication infrastructure, the receiver is, in general, \emph{passive} and the communication mode is \emph{asymmetrical}. In other words, the transmitter is typically in full control of the message generation and manipulation, and thus of its properties.\\
\indent This passive behavior of the receiver and the asymmetry of communication links make the receiver susceptible to adverse channel conditions and any erroneous hardware or air-interface impediments. Augmenting this communication with a sense of symmetry, whereby the receiver can \emph{learn} the structure of the messages and can leverage the history and context of the previously received messages, can potentially improve the robustness of communication with respect to channel and network irregularities. Broadly speaking, if the receiver is endowed with the ability to generate its own messages and make its own conclusions, the E2E wireless network becomes less reliant on and susceptible to the wireless channel and its impediments. To exploit such history and learn context, the concept of \emph{semantic communications}~\cite{lan2021semantic} can be leveraged. Semantic communication is a communication approach that promises to transform radio nodes into intelligent agents that can extract underlying semantics (meaning) in a datastream. That is, when communicating information, radio nodes leverage their reasoning faculties to identify the underlying structure of the message, and the role it plays in their knowledge base. Such a new form of communication can be very beneficial for scenarios in which the reliability of the link is intermittent. Examples of such scenarios include \acp{NTN} (whose links are unreliable due to dynamic obstacles in space) and \ac{EHF} (\ac{mmWave} and \ac{THz}) whose links are highly susceptible to blockage and the radio environment in general. Here, semantic communications can overcome the intermittent behavior with the generative capabilities (via computing) of the receiver. For instance, under a semantic communication paradigm, radio nodes can take advantage of the concept of \emph{semantic showers} to minimize back-and-forth communication with continual and reliable link (See Section~\ref{subsub_robustchannel} for more details). \emph{Finally}, when messages are not perceived as a mere bit-pipeline, various types of context-related information can be discovered and exploited. Here, context is exploited at the level of the \emph{data} itself (and its features) rather than at the level of the \emph{application}. Indeed, even though the information at the application level can enable intelligent decision making at the radio node, such intelligence remains insufficient. That is, performing machine reasoning on the low-level bit-wise data opens the door for can enable the transmitter and receiver to discover the \emph{causal roots of specific events in the messages}, regardless of the application-level information. Such causal information is a new input and knowledge that can be leveraged to steer the \ac{E2E} communication system to attain particular goals or to simply automated the \ac{E2E} wireless network operation. For example, if two robots are communicating with each other collaboratively, understanding the root cause of the messages sent from Robot $1$ (one of the collaborative robots), can enable both robots to reach their ultimate goal more efficiently.\\
\indent Clearly, it is desirable to transform today's communication systems to reasoning-driven semantic communication systems that intrinsically attribute meaning to the exchanged messages. In contrast to traditional communication systems that are driven by dynamic and uncertain communication resources, semantic systems leverage the computing resources and are founded, as will be evident from the rest of this paper, on the concepts of languages, reasoning, and causality. This transformation has the potential to substantially improve the efficiency and intelligence of future wireless networks, and ultimately achieve \emph{more with less} via the convergence of the computing and communication resources. In the following section, we will delve into the details of this transformation.\\
\subsection{From Transmitter/Receiver to Teacher/Apprentice}
The essence of semantic communication is to humanize the communication between a transmitter and receiver so as to mimic knowledge-driven human conversations, interactions, and discussions. At a high level, semantic communication systems are ones that can perceive \emph{the significance or the meaning} contained in a particular message. Semantic communication requires a rethinking of the communication problem with respect to the three levels introduced by Weaver \cite{weaver1953recent}. In particular, right after Shannon proposed information theory, Weaver posited that communication involves problems at three levels \cite{weaver1953recent}: i) \emph{Level A}: The technical problem which measures the accuracy of the symbols of communication to be transmitted; ii) \emph{Level B}: The semantic problem which concerns itself with the precision of the transmitted symbols with respect to the desired meaning; and iii) \emph{Level C}: The effectiveness problem which measures the effectiveness of the received meaning on the conduct of the overall system. Traditional communication systems operated solely within the confines of Level A. However, if properly designed, semantic communication system can take advantage of advances in AI and computing power, and thus, potentially create communication systems that not only encompass all three levels proposed by Weaver, but go beyond them, via \emph{a reasoning plane} (see Section~\ref{scaling}), as more with less can be achieved. To do so, it is necessary to transform today's transmitter and receiver pair into what we propose to designate as \emph{teacher and apprentice} nodes, whose capabilities are, the following:
\begin{enumerate}
        	\begin{figure*} [t!]
	\begin{centering}
		\includegraphics[width=0.8\textwidth]{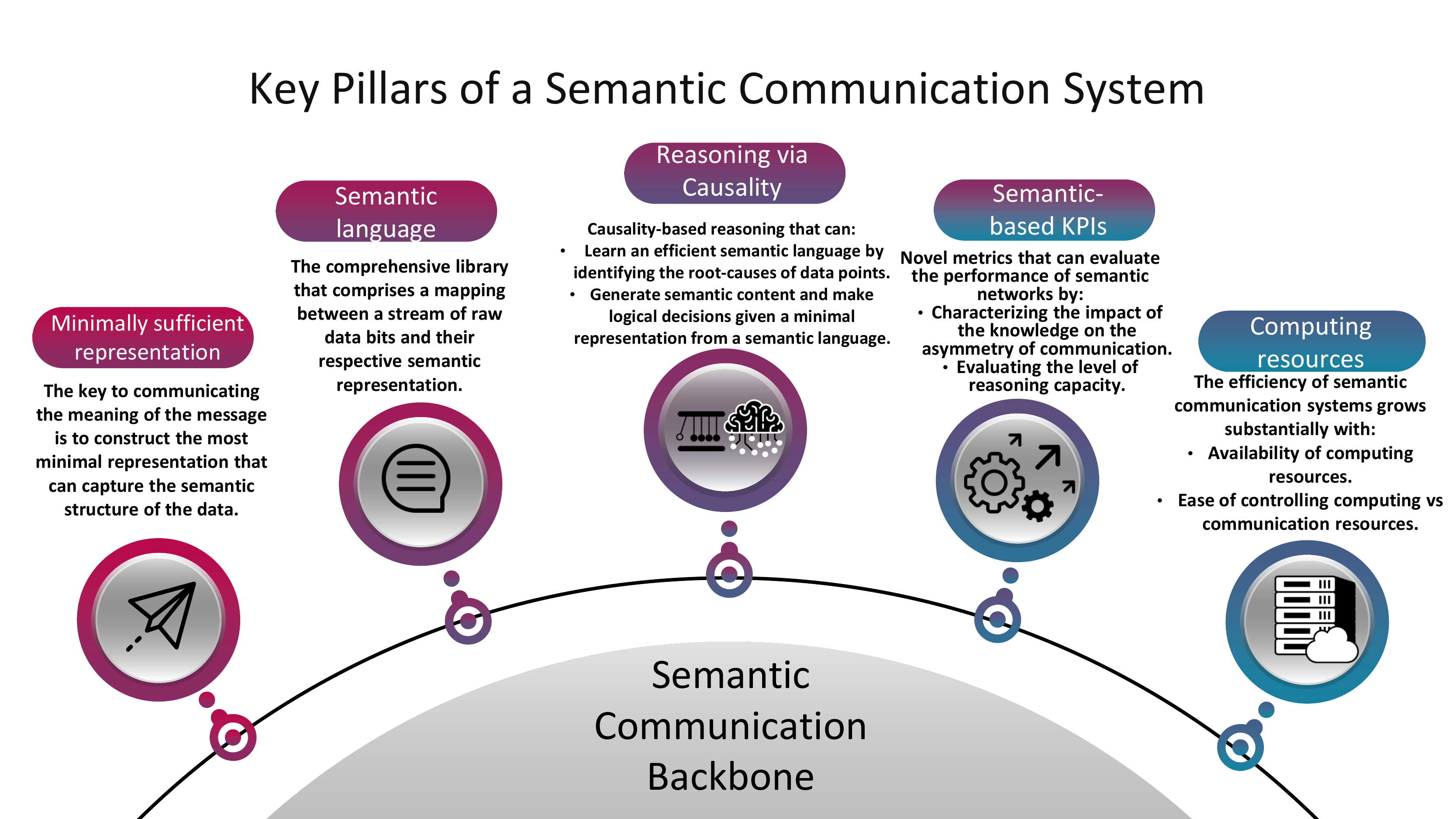}
		\caption{\small{Illustrative figure that showcases the key pillars of semantic communication systems.}}
		\label{fig:ingredients}
	\end{centering}
\end{figure*}
    \item \emph{From a bit-driven transmitter to a knowledge-driven teacher:} The transmitter must be transformed from a bit pipe into a \emph{teacher} capable of, first, disentangling multiple \emph{semantic content elements} within to source data, i.e., separating different meaning, i.e., semantics, contained within a message. Then, for every semantic content element identified, the teacher must craft a \emph{semantic representation} with desirable properties. Essentially, \emph{the semantic content is the ``meaningful" part of the data, and the semantic representation is the ``minimal way to represent this meaning"}. This is similar to the way human beings try to find suitable words to describe their observations and ideas. Also, different semantic content elements could map to different modalities in the data. For instance, when hearing an audio recording of someone's voice, the tone of the voice can be one semantic content element, while the words pronounced are another semantic content element. A human being can easily disentangle and separate those two, and they can also understand the meaning of the words pronounced. Remarkably, today's communication system transmitter cannot identify or separate any underlying semantic structure or modality. It is thus desirable to re-engineer the transmitter to mimic a human being's reasoning capabilities (to the extent possible). In this regard, on the transmission end there is a need for \emph{reasoning}\footnote{It is important to note that this is a broad definition for reasoning. In Section~\ref{section:howto}, we more concretely elucidate the definition of reasoning from a causal perspective.}: The facility that allows the transmitting agent to identify a semantic content element, distinguish it from others existing in the data, and devise an efficient representation of each of those identified contents. This is in stark contrast to the classical transmitter of today's networks that treats its input as a purely random and uncertain string of information and transmits it as a bit-pipeline that characterizes this uncertainty. 
    \item \emph{From a bit-driven receiver to a knowledge-driven apprentice:} Similarly, at the receiver end, \emph{reasoning} capabilities can transform the receiver into an apprentice capable of \emph{understanding the minimal semantic representation} used by the teacher, i.e., mapping it to a semantic content element. Furthermore, the apprentice must be able to generate, via their computing resources, the semantic content element that results from the communicated semantic representation with the highest fidelity possible, e.g. if one of the semantic content elements were part of a hologram transmitted, the apprentice must be able to generate such a hologram with the same resolution that the teacher transmitted it. Moreover, as a result of the developed reasoning capabilities used to understand a semantic representation, the apprentice can use causal and associational (statistical) logic to perform various projections and decisions across the networking stack. Such causal and associational logic is inferred from the progressively built knowledge base and exerted on the received semantic representation.
    \item \emph{From a bit-pipeline to a semantic language:} In semantic communications, the smallest distinct meaningful element is a semantic representation. Moreover, a series of representations constitutes a \emph{semantic language}. Semantic languages will mimic natural languages but they should be less focused to syntax and pragmatics in order to automate processes better (see Section~\ref{near-optimal sem}). Moreover, the semantic representations of a semantic communication language must satisfy three key properties:
    \begin{enumerate}
    \item \textbf{Minimalism}: The capability of characterizing the structure found in the information with the least number of language elements possible (and their equivalent bits). This characterization must be performed in a way to reduce the number of exchanged messages in the long run as well. 
    \item \textbf{Generalizability}: The capability of representing a particular underlying structure (or understanding one at the receiving end) while being invariant to changes in: a) distribution, b) domain, and c) context. Notably, here, context can be viewed as the theme encapsulating various semantics that share a common denominator (for example, the context of messages received by a robot can be a set of steering actions on a tennis court). Hence, generalizability means that, when a radio node has learned and established a particular representation $Z_i$ for a semantic content element $Y_i$, and it can then use such a mature and consistent representation to describe this semantic content element irrespective of the distribution, domain, or context it is extracted from. Hence, the node is now able to generalize its knowledge across multiple, previously unseen and unknown domains, distributions, and contexts. This mimics the behavior of words in a natural language that can \emph{generalize} and apply their knowledge base to describe any occurring event, even if previously unobserved. For instance, a human being can identify and describe a ``loud voice", regardless of what the voice is saying or the environment it is observed in. This generalizability capability represents the epitome of reasoning in regards to making logical conclusions.  
    \item \textbf{Efficiency}: The ability of the apprentice to re-generate the information with high fidelity, in the least time possible. In other words, the resolution of the data generated at the apprentice must be equal (or better) to that which could be recovered by a classical receiver. For instance, if the apprentice is trying to re-generate an audio clip, the resolution of the audio clip must be at the same resolution intended by the teacher. In other words, given that a \emph{reasoning} process has been added, there should not be a tradeoff between the quality of the content recovered and the minimalism achieved via semantic communications. It is important to note, that while the effects of efficiency can be measured via concrete metrics (which are defined via the semantic impact in Section~\ref{section_metrics}), it is more difficult to evaluate the efficacy of the system. That said, for the sake of brevity, in this tutorial when discussing ``efficiency", the term is considered to mean \emph{efficacy and efficiency} simultaneously, i.e., the upper right block of Fig.~\ref{fig:efficiency_efficacy}.
    \end{enumerate}
\end{enumerate} 
    \begin{figure} [t!]
	\begin{centering}
		\includegraphics[width=0.50\textwidth]{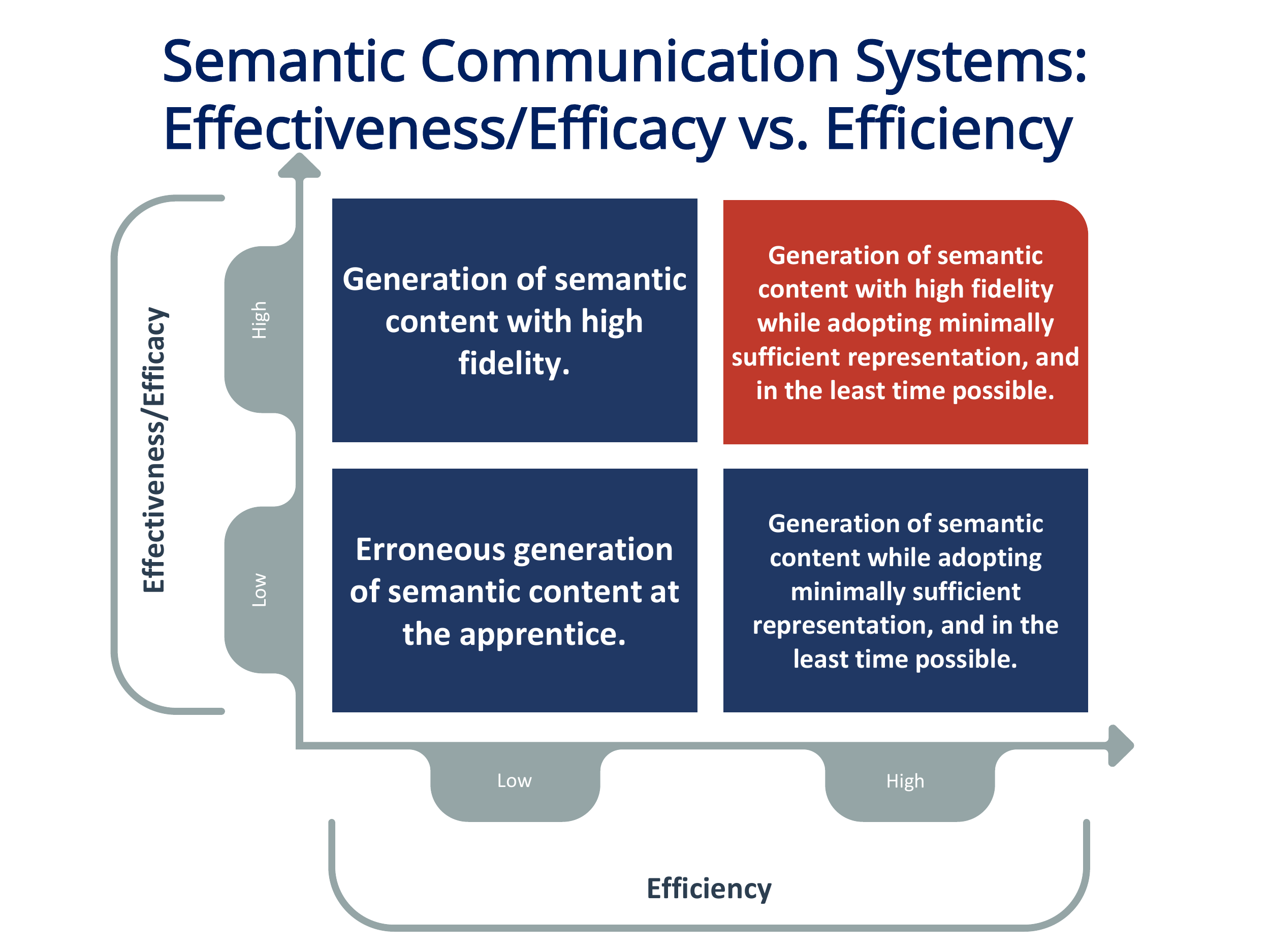}
		\caption{\small{Illustrative figure that showcases the differences between efficiency and efficacy/effectiveness.}}
		\label{fig:efficiency_efficacy}
	\end{centering}
	\vspace{-0.75cm}
\end{figure}
\indent As a result of the aforementioned desirable properties, a well-designed semantic language can achieve \emph{more with less}.\\
\indent A summary of the key pillars of semantic communication systems are shown in Fig.~\ref{fig:ingredients}. Essentially, a classical communication system revolves around the channel and the communication resources, which are dynamic and governed by uncertainty. In contrast, the major thrust of a semantic communication network is its reasoning (and knowledge) capabilities that can be achieved via causality. In essence, as shown in Fig.~\ref{fig:ingredients}, on the one hand, a reasoning radio node will communicate via a semantic language. A semantic language is a comprehensive library that maps every semantic content element in a raw datastream to a \emph{minimally sufficient representation}. A minimal sufficient representation is needed to achieve minimalism both on the short and long term. Ideally, such a representation would: a) minimize the communication resources on the short term, and b) enable the apprentice's reasoning capabilities to scale up, to ultimately generalize and perform versatile logic operations. On the other hand, properly deploying reasoning capabilities is at the helm of evaluating the semantic communication network via a suite of novel semantic-based \acp{KPI} which can properly capture the new reasoning dynamics of the performance. Furthermore, causal and associational logic via reasoning is not \emph{feasible} without abundant computing resources that exhibit a flexibility in control, in contrast to communication resources.\\
\\ In a nutshell, we have thus far elucidated the roles of the teacher and apprentice in a semantic communication system. We have also highlighted the need for a semantic language and overviewed its unique characteristics. Next, we highlight the main contributions of this article.
\subsection{Contributions}
The main contribution of this article is a novel and holistic vision that articulates fundamental principles necessary to build next-generation reasoning-driven, \ac{AI}-native semantic communication networks. In particular, we \emph{first,} investigate the key tenets necessary to extend today's classical information theory towards a semantic information theory. This extension is performed via a migration from today's \emph{bit-pipeline} to a \emph{semantic language}. Second, we scrutinize the reasoning foundations that are imperative for the communication of a semantic language. These foundations are centered around the migration from \emph{data-driven} networks towards \emph{knowledge-driven and reasoning-driven} ones. In essence, organizing \emph{information} is majorly influenced by the capability of a radio node to unravel causal and associational relationships and logic. In this regard, we propose rigorous reasoning techniques that must be adopted in gradually building a language, to ultimately reach a \emph{generalizable, minimal, and efficient} semantic language. Here, we particularly shed light on the significance of causality to rise up in the \emph{reasoning ladder} (see Fig.~\ref{fig:causal_logic}). Third, we propose a suite of novel \emph{semantic \acp{KPI}} for evaluating the performance of \ac{AI}-native, reasoning-driven systems, and optimizing future semantic networks. Finally, we discuss how one can build scalable semantic communication networks while bringing forth novel approaches and concepts to address several computing, control, and networking challenges. Furthermore, through our contributions, we answer the following fundamental questions:	
\begin{itemize}
    \item \emph{How do we extend classical information theory to capture semantic information?}\\
    The performance of today's communication systems is evaluated based on principles derived from Shannon's information theory. That said, information theory is built on the premise of defining ``information" as a mere ``uncertainty" that does not perceive meaning or structure\cite{weaver1953recent}. In consequence, we investigate and characterize the equivalents of today's ``information" and ``entropy" in a semantic communication system. Then, we discuss how these novel concepts modify the way communication is intrinsically viewed and evaluated.
    \item\emph{Why do we need a semantic language? How is it different from a natural language?}\\
    For a radio node to become capable of communicating using a semantic language, it must be able to: a) extract semantic content elements from the data, b) map into a minimal semantic representation, and c) understand a semantic representation occurring in various domains, contexts, and stemming from different distributions. We show that a semantic language is fundamentally different from a natural language. The atomic unit of a semantic language is a \emph{representation} that captures \emph{the structure and variability of the represented semantic content element}. Meanwhile, the atomic unit of natural language is a word. Limiting a semantic language to a natural one, would constrain it in syntax and wording which (unlike causal and associational logic) are governed by deterministic rules.
    \item\emph{How do we semantically process data, and how can we build a semantic language?}\\
    Many tools can enable extracting a semantic representation from data. However, the right approaches must be able to create a semantic representation that is minimal, efficient, and generalizable. To obtain such a representation, one must be able to characterize the causal and statistical properties of the data. Thus, after surveying the set of existing tools for representing semantic information, we expose the fundamentals of causal representation learning and its accompanying benefits, challenges, and future directions for building next-generation semantic communication networks.
    \item \emph{How do we move from data-driven intelligence towards knowledge-driven reasoning?}\\
    Moving from data-driven intelligence towards knowledge-driven reasoning requires engineering the semantic language based on a model that can characterize causal and associational logic. As a result, we demonstrate that mapping a language to a \emph{\ac{SCM}}, enables exploiting the concepts of \emph{interventions and counterfactuals} from causal logic. Such concepts allow building a semantic control plane, whereby instead of classical acknowledgements and non-acknowledgements, the apprentice can gather information about the structure of the previously conveyed representation. This process enables a gradual acquisition of a language at the apprentice, which ultimately leads to elevating the radio node in the causal reasoning ladder.
    \item \emph{How do we evaluate the performance of semantic communication systems?}\\
    The evaluation metrics of classical communication systems have heavily relied on Shannon's information theory, however, future evaluation schemes for semantic-based systems must capture the structure of the semantic representations and the reasoning capability of the teacher and apprentice. Thus, we propose three novel semantic-based metrics that enable characterizing the semantic impact a particular representation can generate (which characterizes the gain in time and resources when relying on a semantic representation versus classical data), the communication symmetry index, and the reasoning capacity of a semantic communication link.
    \item \emph{How do we scale semantic communication systems to current and future large-scale cellular communication networks?}\\
    Today's 5G cellular networks are  characterized with a separated control and user plane. With the introduction of semantic communications, many fundamental changes are necessary ranging from the need to integrate causality and reasoning to designing an expressive yet minimal semantic language. One key change is the need to introduce a novel reasoning plane, that would be sandwiched between the control and the user plane. Based on the real-time inference performed in the reasoning plane, the control plane is fed with information that enable radio nodes exchanging interventions and counterfacturals. These are queries that replace acknowledgements and non-acknowledgements and enable the teacher and apprentice to build and learn a language. Also, with the introduction of semantic communication, today's \ac{O-RAN} concept will evolve further to account for real-time, near-real-time, and non-real time intelligence. 
\end{itemize}
\begin{figure*}
    \centering
    \includegraphics[width=0.75\textwidth]{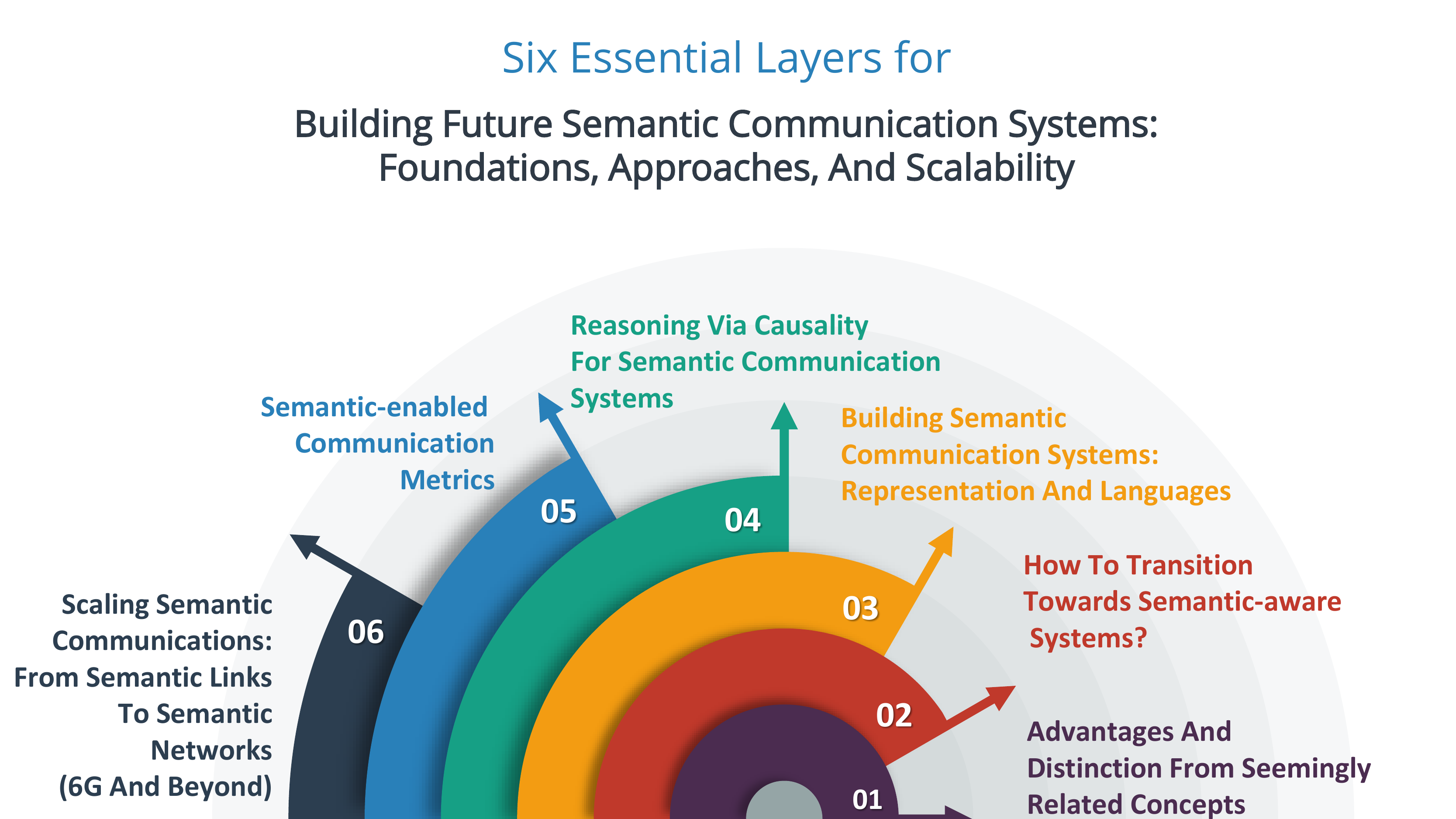}
    \caption{\small{Main sections of this tutorial via the 6 Essential Layers for Building Future Semantic Communication Systems.}}
    \label{fig:6layers}
\end{figure*}
\subsection{Prior Works}
Recently, a number of surveys and tutorials related to the concept of semantic communications have appeared in \cite{petarsem, qin2021semantic, kalfa2021towards, strinati20216g, gunduz2022beyond, luo2022semantic, niu2022towards, yang2022semantic}. The authors in \cite{petarsem}, presented a view on semantic communications within three communication modalities: human-to-human, machine-to-human, and machine-to-machine. The authors in \cite{qin2021semantic} present the developments of \ac{DL}-enabled semantic communications for multi-modal data transmission, including text, image, and audio. In \cite{kalfa2021towards}, the authors overview a semantic signal processing framework that can be tailored for specific applications and goals. In \cite{strinati20216g} and \cite{gunduz2022beyond}, semantic and goal-oriented communications were overviewed while highlighting the network benefits in terms of reliability and effectiveness. The work in \cite{luo2022semantic} presents key methods for performing feature extraction based on semantic communications. In \cite{niu2022towards}, the authors analyze semantic communications from an information-theoretic perspective. The authors in \cite{yang2022semantic} discuss how 6G technologies can drive the development of semantic communications.\\
\indent While the works in \cite{petarsem, qin2021semantic, kalfa2021towards, strinati20216g, gunduz2022beyond, luo2022semantic, niu2022towards, yang2022semantic} are interesting they have not considered various concepts and fundamentals that are necessary to define, design, and ultimately deploy semantic communications systems:
\begin{itemize}
    \item \textbf{Representation:} While the works in \cite{petarsem, qin2021semantic, kalfa2021towards, strinati20216g, gunduz2022beyond, luo2022semantic, niu2022towards, yang2022semantic} acknowledge the need to depart from a latent bit-pipeline, such works do not lay the language pre-processing techniques that are needed to move from entangled raw datastreams to \emph{learnable} datastreams that can be used to learn a semantic language. Moreover, the previous works in~\cite{petarsem, qin2021semantic, kalfa2021towards, strinati20216g, gunduz2022beyond, luo2022semantic, niu2022towards, yang2022semantic} fail to articulate the necessary measures that enable the apprentice to \emph{understand} a representation, i.e., leverage it to efficiently generate semantic content elements via their computing resources.
    \item \textbf{Semantic Language:} While some of the works in \cite{petarsem, qin2021semantic, kalfa2021towards, strinati20216g, gunduz2022beyond, luo2022semantic, niu2022towards, yang2022semantic} acknowledge the need for a language, these works fail to explicitly define a semantic language,  its corresponding characteristics, and how it can improve the efficiency of communications. Moreover, to mimic human conversations such works often confuse natural languages with semantic languages. In contrast, this tutorial will be the first to extend Shannon's information theory to the semantic communication domain by investigating the necessary measures to gradually build a semantic language. In fact, the main goal of a semantic language is to express a minimal semantic representation vis-à-vis the raw datastream and its contained semantic content element. Such a language is characterized with \emph{minimalism, generalizability, and efficiency}. Also, our work will be the first to further elucidate the fundamental tenets of this language by distinguishing from natural languages with respect to syntax, pragmatics, and semantics. 
    \item \textbf{Reasoning and Causality:} The prior art \cite{petarsem, qin2021semantic, kalfa2021towards, strinati20216g, gunduz2022beyond, luo2022semantic, niu2022towards, yang2022semantic} does not provide any concrete technical approaches to perform reasoning. In particular, they do not put into perspective the importance of \emph{causality} in the data. In many ways, existing works limit themselves to statistical and associational relationship in the data that fail to unravel the underlying structure of the data. Also, relying on such statistical relationships may lead to spurious representations. Instead, in this tutorial, we are the first leverage the concept of causality. In essence, ultimately enables communication nodes to recognize the root causes of specific datastreams via the notions of interventions and counterfactuals. 
    \item \textbf{Semantic \acp{KPI}:} All the prior works in this area in \cite{petarsem, qin2021semantic, kalfa2021towards, strinati20216g, gunduz2022beyond, luo2022semantic, niu2022towards, yang2022semantic} still rely on classical \acp{KPI} such as rate, reliability, and latency. These \acp{KPI} cannot capture the reasoning capability of radio nodes, nor can characterize the level of communication symmetry. In contrast, we propose a suite of semantic-based evaluation metrics that enable characterizing the performance bounds of any semantic communication system. For example, we derive a metric called ``reasoning capacity" that can ultimately have an impact that is higher than Shannon's capacity because its reliance on computing resources (in contrast to communication resources).
    \item \textbf{Scalable semantic communications:} The works in \cite{petarsem, qin2021semantic, kalfa2021towards, strinati20216g, gunduz2022beyond, luo2022semantic, niu2022towards, yang2022semantic} are limited to one source and destination, meanwhile the potential of semantic communications cannot be fully unleashed unless it is considered over large-scale cellular networks. As such, in this work we elucidate the challenges and the opportunities regarding scaling semantic communications over a wireless network.
\end{itemize}
\indent With the currently established literature, designing and building future semantic communication systems from the ground up is an extremely strenuous and difficult task. Essentially, there is a major lack in existing works that investigate novel knowledge and reasoning frameworks, which constitute a central pillar to propose a comprehensive framework for semantic communications. It is thus necessary to investigate the overarching measures needed to successfully usher the birth of semantic communication systems in beyond 6G systems. Next, we examine the preliminaries of semantic communications by highlighting its underlying benefits and distinguishing the concept of semantic communications from alternative concepts that have recently emerged.
\subsection{Organization}
\indent The rest of this paper is organized as shown in Fig.~\ref{fig:6layers} and as follows. In Section~\ref{section_preliminaries} we discuss some of the advantages of semantic communications and its relationship to existing techniques. Then, in Section~\ref{section_transitioning} we discuss novel concepts and definitions that enable a smooth transition from classical communication systems towards semantic communications system. Subsequently, in Section~\ref{section:howto} we thoroughly investigate novel views and techniques that enable establishing an expressive semantic representation and language via scrutinizing the structure in the data. Then, in Section~\ref{reasoning_via_causality}, we demonstrate how to leverage, for the first time, the concept of causality to ultimately equip radio nodes with a reasoning capability. Furthermore, in Section~\ref{section_metrics}, we propose a suite of novel semantic metrics that enable evaluating emerging semantic communication system. Then, in Section~\ref{scaling} we develop some of the key techniques that enable the design and deployment of semantic communication networks at scale.  Finally, conclusions and recommendations are drawn in Section~\ref{Sec:Conclusion}.
\section{Semantic Communications: Advantages and Distinction from Seemingly Related Concepts}\label{section_preliminaries}
Before delving into the main technical components of semantic communications, we first distinguish the concept of semantic communications from alternative frameworks and concepts that were recently proposed for next-generation wireless systems. Then, we overview the benefits of semantic communication networks.
	\begin{figure*} [t!]
	\begin{centering}
		\includegraphics[width=0.80\textwidth]{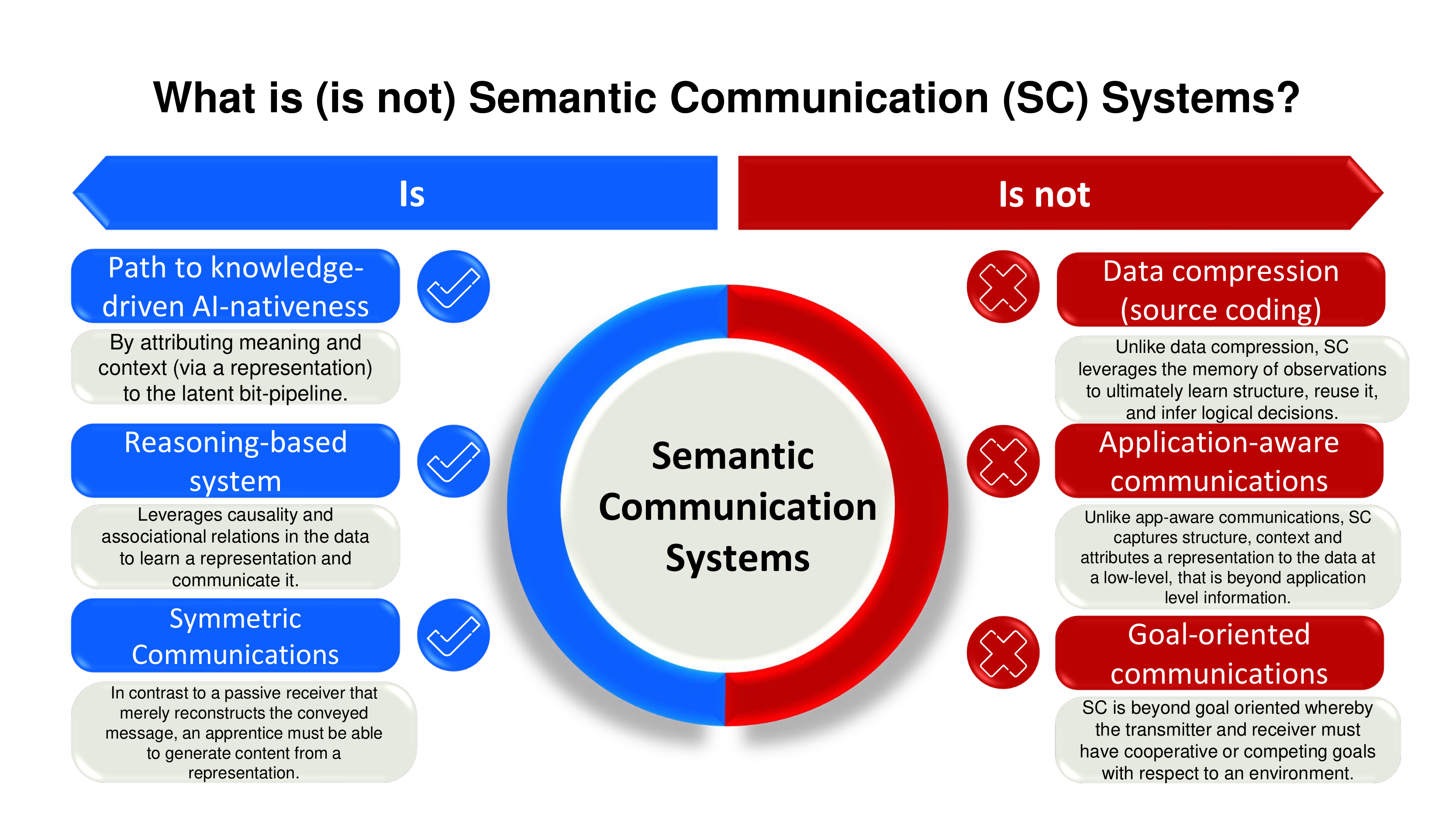}
		\caption{\small{Illustrative figure showcasing what is (is not) semantic communications.}}
		\label{fig:whatisnot}
	\end{centering}
\end{figure*}
\subsection{What is NOT Semantic Communications?}
At first glance, semantic communications can seem like an incremental variant of known approaches and techniques. In this subsection, we attempt to demystify this confusion by highlighting the fundamental differences between such techniques and semantic communications. In Fig.~\ref{fig:whatisnot}, we summarize our answer to the questions of what is and what is not semantic communications. 
\subsubsection{Semantic communications is not data compression}
In a classical communication setting, according to information theory\cite{shannon1948mathematical}, the process of data compression (also known as source coding or bit reduction) is the process of encoding information\footnote{Here information denotes the Shannon definition of information, and thus is designating the concept of uncertainty which will be elaborated in Section~\ref{semantic_substance}.} using fewer bits than the original datastream representation. This process is performed by exploiting the statistical redundancy in the data bits to ultimately represent data without any loss of information. As such, the process becomes reversible at the receiver. While data compression shares some common ground with semantic communications with respect to minimalism, i.e., minimizing the size of data transmitted, both concepts are fundamentally different:
\begin{itemize}
    \item Data compression achieves minimalism, i.e., shrinking the size of a particular datastream, by identifying and eliminating statistical redundancy. For instance, the most prominent lossless compressors employ probabilistic models such as prediction by partial matching \cite{zhang2008prediction}. Notably, there is a close connection between data compression and \ac{ML}, in that they both specifically attempt to predict the posterior probabilities of a sequence given its history. Therefore, a duality arises between data compression and \ac{ML}, to the extent that some works in \cite{sculley2006compression}, consider data compression as a key method that can be used in \ac{ML} for tasks like clustering and classification. That said, as a concept, data compression does not bear any learning ability that contributes to a particular training memory or a trained model. In other words, from an \ac{ML} perspective, data compression techniques often intend to \emph{overfit} since their only goal is to shrink the current datastream and not the future ones. Thus, while data compression can be a component within an \ac{AI} technique, it does not exhibit \emph{learning} characteristics, let alone reasoning. From a minimalist perspective, data compression could even be more beneficial than semantic communications on the short term. However, it is unable to instill contextual information and knowledge-driven memory on the receiver. In many ways, on its own, data compression is restricted to the realms of data-driven and information-driven networks from Fig.~\ref{fig:datatoknowledge}.
    \item In contrast to data compression, in semantic communications, instead of identifying data redundancies and compressing them, patterns that map to structure and semantic content are identified, learned, and then represented with a semantic representation. In essence, in source coding and data compression, the goal is to \emph{overfit} to the statistical characteristics of the datastreams. Meanwhile, semantic communication's ultimate goal is to characterize structure, and, thus, the focus is not on the pure randomness exhibited in the data. In fact, these random data points are better transmitted classically as we explain in Section~\ref{section:howto}. Furthermore, semantic communication achieves ``minimalism" as a byproduct of the semantic representations transmitted which: a) Comprise a fewer number of bits in total, b) Serve to teach the apprentice to learn, generate, and ultimately automate the task or message at the receiver. Consequently, the characteristics of the adopted representation and the acquired reasoning capabilities enable minimalism via: a) Minimizing the number of bits per transmission, and b) Minimizing the total number of transmissions necessary to convey a message or achieve a task. Hence, semantic communication networks achieve minimalism via different mechanisms that go beyond a compression of the number of bits in a packet. Finally, when radio nodes operate based on organized knowledge, such nodes can make more informed logical conclusions across the networking stack, a feature only possible with reasoning-driven semantic communication networks.
\end{itemize}
\subsubsection{Semantic communications is not only an ``AI for wireless" concept}
 \ac{AI} has been used for many wireless-related problems in the past few years. This includes \ac{AI} and \ac{ML} for channel estimation, beamforming, network management, receiver design, etc. In essence, in all of these tasks, the design, performance, or optimization of the task was procured via an \ac{AI} or an \ac{ML} tool instead of traditional numerical methods. \ac{AI}-enabling such tasks was shown to improve the accuracy, precision, or relevant \ac{KPI}, however, the fundamental functionality and dynamics of the corresponding wireless task was kept the same. For example, \ac{AI}-enabled channel estimation is fundamentally an improved approach to perform classical channel estimation solutions (e.g. as performed in  \cite{soltani2019deep}), yet the core task to be performed remains the same. In contrast, semantic communication systems will not be an additional \ac{AI}-enabled layer on top of an originally existent task. In other words, semantic communications is not an \ac{AI}-improved air interface.\\
\indent With semantic communications, \acp{UE} and \acp{BS} do not need to rely continuously on the channel. For instance, for scenarios in which the radio nodes have established a solid knowledge base, the need for for channel estimation becomes minimal. Moreover, instead of relying on bit-driven signaling messages to sense the channel, a continual understanding of the context of previous messages can potentially establish an awareness of the physical environment. Such awareness can be leveraged to \emph{learn} the channel characteristics while relying on computing resources. In essence, the \emph{mechanism of communication tasks fundamentally changes with the introduction of semantic communications} (as previously explained for channel estimation as an example). This can be observed via the following insights: 
 \begin{itemize}
     \item In contrast to a classical AI-based transmitter that still relays bits ``as is'', the teacher in a semantic communication network can now attribute a semantic language to the raw bit-pipeline previously observed. Meanwhile, an \ac{AI}-layer can only add a prediction margin to the classical tasks of compression, transmission, and reconstruction.
     \item When communication relies on a semantic language, context becomes of relevance. As such, the more consistent the theme of the conversation is, and hence the context, the larger the improvement in the reasoning capability at the source and destination. In contrast, classical \ac{AI}-augmentations alone are not aware of the concept of context, i.e., the logical theme surrounding the data structures learned by exploiting knowledge base. Instead, data/information-driven \ac{AI} algorithms are ultimately reliant on their data input and their corresponding statistical properties.
     \item Classical \ac{AI} for wireless attempts to improve the performance after being trained a priori via large datasets, or via a trial-and-error phase (e.g. \ac{RL}) over the inputs from a specific environment (e.g. channel, spectrum, \ac{QoS} values). Instead, semantic communications \emph{gradually} builds a language between the teacher and the apprentice. This gradual language construction enables the teacher and the apprentice to organize and build their knowledge base. Thus, the radio nodes now acquire \emph{human-like} reasoning faculties. That is, a radio node can now: a) make conclusions according to its knowledge base (not data), and b) communicate its needs based on such conclusions. 
 \end{itemize}
 Clearly, based one the above key observations one can conclude that semantic communications is beyond a simple use of an \ac{AI} algorithm for a wireless task. Given that semantic communications enables radio nodes to build a knowledge base and communicate a language, the mechanism of communication fundamentally improves.
\subsubsection{Semantic communications is not only goal-oriented communications}
A goal oriented communication system involves a number of agents that interact and exchange messages to achieve a joint goal or separate goals that include the same environment. For example, two robots can interact with each other to execute a common mission. Here, in contrast to sending the information gathered by sensors bit-by-bit, the robots can exchange multiple feedback messages of their current semantic action, their next expected outcome, all while achieving a unique joint goal. In a goal-oriented framework, the nodes, e.g., the teacher and apprentice can also be achieving two separate goals. Much of the early-on work on semantic communication has equated it with such goal-oriented communication systems \cite{xie2022task, binucci2022dynamic, farshbafan2022curriculum, zhang2022goal}. However, there are fundamental differences between the two concepts. In some sense, goal-oriented communications falls under the umbrella of semantic communications. For instance, in every goal-oriented communication system, the nodes will have to embed semantic representations to ultimately achieve a particular goal. In contrast, under the broader auspices of a semantic communication system, the generation and communication of semantic representations is not necessarily done for the purpose of serving a system-wide goal \\
\indent In this regard, limiting the concept of semantic communications to the confines of goal-oriented systems will therefore unnecessarily limit its use to a subset of use-cases that have a competitive or cooperative nature. Meanwhile, there are many instances in which the teacher and the apprentice do not necessarily share any joint goals nor interact with a common environment. For instance, the teacher can be a server that is transmitting highly-data intensive content (e.g. \ac{XR} content) to a particular user. Here, every standalone content transmitted can have an entirely different goal, and there are no cooperative or competing goals between the teacher and the apprentice. Yet, in this case, semantic communications can still be used to: a) rely less on the channel to transfer massive information content, b) empower radio nodes with reasoning to make versatile decisions, which can enhance network's capability in meeting the stringent requirements of future applications. 
\subsubsection{Semantic communications is not only application-aware communications}
Implementing the context of information within the transmission of messages may seem at first glance similar to the traditional concept of application-aware communication. In fact, there are many prior works (e.g. \cite{appawareclancy, chendatacorr, temdee2018context, silva2022semantic, contextaware2020, contextQoS2019}) that have fine-tuned the network optimization process to address application-level requirements. For example, in \ac{XR} applications, the \ac{XR} content transmitted by users may exhibit a particular correlation. Here, some works such as  \cite{chendatacorr} exploit this correlation to ultimately improve the management of uplink and downlink wireless transmissions. Notably, it is important to distinguish between the ``context-awareness" concept defined by such frameworks and the one granted with semantic communications systems, as the former is a mere application and use-case specific awareness. In contrast, in semantic communication systems, \emph{``context" is a concept defined with respect to the low-level structure of exchanged datastreams between the transmitter and the receiver}. Such low-level intelligence opens the door for an inter-application, intra-application, and out-of-domain generelizability. In other words, a radio node can leverage the meaning attributed to low-level data corresponding to service $A$ by using it to improve the \ac{E2E} performance for service $B$.\\
\indent On top of gaining generalizability, the ``awareness" gained from application requirements, as done in classical application-aware works \cite{appawareclancy, chendatacorr, temdee2018context, silva2022semantic, contextaware2020, contextQoS2019}, majorly relies on statistical learning and cannot be easily extended to solve multiple challenges across the \ac{OSI} model. Meanwhile, semantic communications intrinsically relies on causal and statistical relationships in the context of the data. For example, after acquiring the structure of the data, the radio node can track the root cause of a mismanaged resource orchestration to ultimately predict and proactively prevent a beamforming error from happening. The radio node can also conclude information about the type of terrain in which communication is taking place (e.g. rural, urban).\\
\indent Furthermore, some works \cite{stamatakis2022semantics,  uysal2022semantic, maatouk2022age} view semantic communications through a \emph{significance perspective}. Such works \cite{stamatakis2022semantics,  uysal2022semantic, maatouk2022age}, consider metrics like \ac{AoI}, value-of-information, and other time-oriented metrics to be indicative of \emph{significance} of the use-case. While such metrics are useful in enhancing the performance of time-critical communications, they are still limited to certain use cases, and they do not possess the generalizability and reasoning capabilities needed for semantic communications. That is, \ac{AoI} is a networking metric that does not unravel low-layer information (which contains the structure of the data). Such networking metric alone, do not allow a radio node to build a knowledge base, thus such radio nodes fail to perform any reasoning-driven tasks. Additionally, such metrics are still highly dependent on communication resources and are mainly constrained to the framework of particular use-cases and time-critical communications. Thus, for instance, such significance measures cannot universally enhance the performance irrespective of the use-case or its time-criticality. In many ways, \ac{AoI} and its variants still lie in the scope of information-driven networks from Fig.~\ref{fig:datatoknowledge}.
 Clearly, semantic communications equips nodes with an intelligence that has breadth and depth that is beyond the one gained with application-aware communications.
 	\begin{figure} [t!]
	\begin{centering}
		\includegraphics[width=0.45\textwidth]{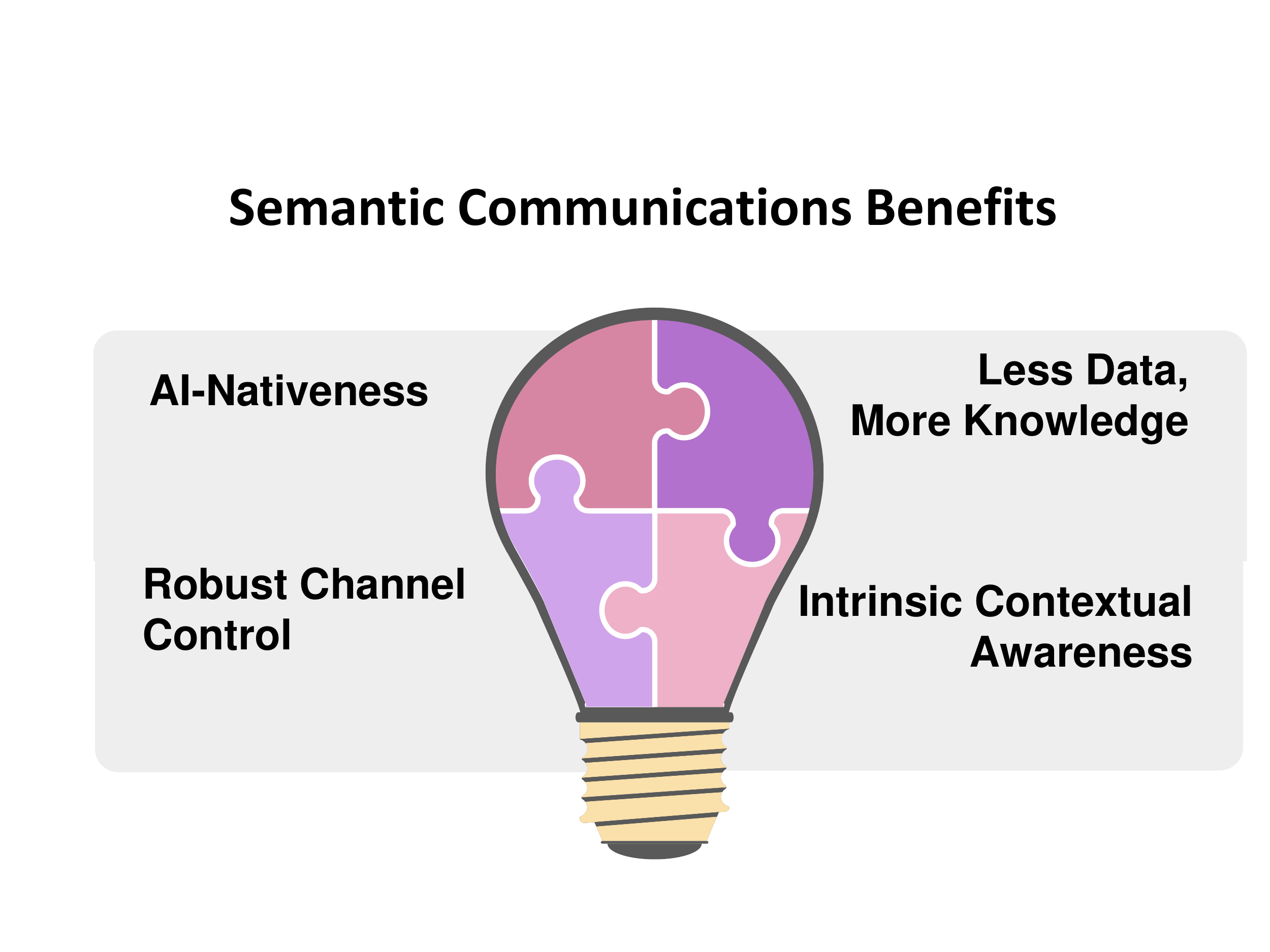}
		\caption{\small{Illustrative figure showcasing the benefits of semantic communications.}}
		\label{fig:benefits}
		\vspace{-0.25cm}
	\end{centering}
\end{figure}
 \subsection{Benefits of Semantic Communications}
 Semantic communications will bring forth many key benefits for future communication systems, in what follows we detail those benefits.
\subsubsection{Achieving Reasoning-Driven AI-Nativeness} \label{AI-nativeness-section}
6G and beyond systems must be \ac{AI}-native across their protocol stack, in the sense that every single component, layer, and structure in the network must be designed, deployed, and optimized via \ac{AI}. Nonetheless, today's industry inaccurately describes future \ac{AI}-native systems, as ones reliant on data or information, as shown in Fig.~\ref{fig:datatoknowledge}. Such classical \ac{AI} approaches are rigid, big-data dependent, and are knowledge-agnostic. Meanwhile, as we discussed in Section~\ref{sec:Intro} we define \ac{AI}-native networks, as ones that \emph{rely on less data but more knowledge}. That is, the overarching goal is to create \emph{reasoning-driven} systems as shown in Fig.~\ref{fig:datatoknowledge}. Effectively, semantic communication can provide a path to a next-generation of knowledge-driven and reasoning-driven radio nodes. This can be done by: a) Transforming today's bit-pipeline communication to one that relies on a semantic language. In essence, to communicate a semantic language, radio nodes must utilize the input semantic representation to \emph{computationally generate} semantic content elements, and b) Performing reasoning which is attained by extracting/understanding semantic representations. This enables radio nodes to capitalize the context of the semantic language, as well as their knowledge base to make versatile decisions across the networking stack. Clearly, radio nodes that communicate a semantic language and are reasoning-driven, can potentially reach the real-time prediction, automation, and agility needed for 6G and beyond systems.
\renewcommand{\arraystretch}{0.9}
\begin{table*}[!t]
	\footnotesize
	\caption{Common lexicon used in semantic communication systems.}
	\centering
	\begin{tabular}{m{5.5cm}m{7cm}m{4cm}}
		\hline
		\textbf{Vocabulary} & \textbf{\hspace{0.4cm}Definition} & \textbf{\hspace{0.4cm}Mathematical Expression} \\
		\hline
		\textbf{Teacher}& A transmitter with reasoning capabilities. A teacher is capable of first disentangling multiple semantic content elements to be transmitted, i.e., separating different meaning contained within the message. Then, for every semantic content element identified, they will craft a semantic representation with desirable properties.& $b \in \mathcal{B}$\\
		\hline
		\textbf{Apprentice}& A receiver with reasoning capabilities. The apprentice can map the conveyed semantic representation to a semantic content, i.e., mapping the minimal representation to its corresponding meaning. Then, the apprentice can generate the content at the destination with the same fidelity initially produced at the source. & $d \in \mathcal{D}$.\\ 
		\hline
		\textbf{Semantic Content Element} & The meaning of a specific datastream  (or a label denoting this meaning). & $Y_i$\\
		\hline
		\textbf{Fidelity of Information} & A high fidelity corresponds to recovered information with an equivocal resolution of the data type and content transmitted, e.g. for image data, a high fidelity is an image recovered in its original resolution.& N/A \\
		\hline 
		\textbf{Semantic Representation} & A representation that has desirable properties, and that is capable of ``describing" the meaning of a datastream. This representation must be sufficient so that the apprentice can generate the semantic content without sacrificing the fidelity of information.  & $Z_i$ \\
		\hline		
		\textbf{Semantic Language} & A dictionary (in terms of data structures) that maps every raw datastream to its corresponding semantic representation.  & $\mathcal{L}$	\\
		\hline
		\textbf{Semantic Didactics} & A combination of a stream of a semantic representations, complemented with a raw datastream sent by the teacher to gradually teach the apprentice the semantic language.  & N/A\\
		\hline
		\textbf{Disentanglement} & The process of separating multiple semantic content elements in a single datastream.  & Eq.~\ref{disentangled}\\
		\hline
		\textbf{Reasoning} & \begin{itemize} \item On the transmitter's end: A reasoning-driven teacher capable of disentangling multiple semantic content elements within a datastream, and attributing each one a semantic representation \item On the receiver's end: A reasoning-driven apprentice, mapping the received semantic representation to a semantic content elements, and generating such content with high fidelity. \item Reasoning at both teacher and apprentice must allow them to use their built knowledge base to perform logic operations and draw logical conclusions across the networking stack. \end{itemize}& Proposition~\ref{reasoning_capacity}  \\
		\hline
		\textbf{Context} & A single context corresponds to a family of structures that is shared upon multiple recurrent datastreams. It can also be viewed as the theme encapsulating various semantics that share a common denominator.&  N/A \\
		\hline
		\textbf{Dynamic Reasoning} & The capability to perform reasoning over varying context. & N/A \\
		\hline
		\textbf{Minimalism} & The capability of characterizing the structure found in the information with the least number of bits possible. This characterization must be performed in a way to reduce the number of exchanged messages on the long run& N/A \\
		\hline
		\textbf{Efficiency} & The ability of the apprentice to re-generate the information with high fidelity, in the least time possible. That is, the resolution of the data generated at the apprentice must be equal (or better) to the one that could be recovered by a classical receiver. & Large semantic impact, i.e., $\iota_\tau>1$ (See Definition~\ref{sem_impact} and Proposition~\ref{com_symmetry}).\\
		\hline
		\textbf{Generalizability} & The capability of representing an underlying structure when dealing with datastreams of varying: a) distribution, b) domain, and c) context. This embodies the ability of a radio node to generalize and use its knowledge base to draw conclusions via a mature semantic language that can identify semantic content elements irrespective where they are drawn from.    & Definition~\ref{gen}.\\
		\hline
	\end{tabular}
	\label{table:Vision}
\end{table*}
\subsubsection{Intrinsic Contextual Awareness}
One key benefit of semantic communications is that it can grant radios with an intrinsic awareness of the contextual information of their data transmission. This means that radio nodes (e.g. \ac{BS} or a \ac{UE}) at any point in time, become aware of the context, i.e., the revolving communication theme of recurrent messages. For instance, if a person is sending pictures of themselves and their pet in a park, the spatio-temporal settings, as well as the revolving picture colors have an underlying \emph{family of structure}. This consistent family of data structures is the \emph{context} of the messages transmitted in this scenario. An awareness of context is particularly important for \ac{IoE} services as they require the co-design and control of different functional modalities of communication, computing control, sensing, and tracking. Herem one can for example improve the optimization of of control, localization, and sensing messages. The converse can also be true: Properly extracting the semantic information gained from sensing or tracking messages, can further improve the optimization of communication resources. Moreover, this contextual awareness enables the radio nodes to be self-governed and self-optimized while being offline, i.e., such radio nodes can rely on their knowledge base to make future logical decisions without a continuous connectivity. For instance, in many settings, the radio node can extract the spatio-temporal changes from the semantic content of a communication message. Hence, owing to the spatio-temporal changes causally learned, the network can minimize the number of control acknowledgements needed as well as frequent tracking and signals. This can ultimately improve the spectrum and energy efficiency of the sensors, radar \acp{BS}, and control \acp{BS}.
\subsubsection{Robust Channel Control}\label{subsub_robustchannel}
To date, wireless communication systems have always been governed by the channel and its uncertainty. In essence, most of today's wireless communications research efforts aim to analyze and optimize the network performance to ultimately render the information transfer task more efficient over the \emph{uncontrollable channel}. Nonetheless, in semantic communications, the communication link is no longer an asymmetrical link that mainly relies on decoding every single bit to recover the identical message transmitted. In fact, as a byproduct of the convergence of computing and communication resources, the teacher and apprentice's center of gravity will move more towards \emph{controllable computing resources}. This shift towards computing enables communication systems to rely less on the classical 3GPP concepts of reliability and continuity. Thus, a higher level of independence and robustness to various channel conditions can be gained. Such independence and robustness is exhibited by two mechanisms: First, given that communication now depends on a semantic language, the teacher can offload \emph{semantic showers} to the apprentice. Such semantic showers would contain the language that \emph{explains} to the apprentice the meaning of the message. Then, by utilizing their computing resources, the apprentice would generate the elements of the service content. This minimizes the back-and-forth exchange of user and signaling messages needed to continuously convey information via a communication channel. In fact, such semantic showers can be particularly exploited in future networks that will rely on intermittent \ac{THz} or \ac{mmWave}. Also, such showers enable improving the intermittent service of \acp{NTN} that are needed to \emph{bridge the digital divide} and provide global coverage. Second, on top of minimizing the back-and-forth messaging, radio nodes in a semantic setting can leverage their knowledge base to \emph{correct erroneous semantic representations}. Instead of solely relying on error correcting codes, when a semantic representation results in a \emph{spurious semantic content element}, i.e., inconsistent with the current context of communication, the radio node could deploy its logic to \emph{predict} what representation the teacher was intending to send. 

	\begin{figure*} [t!]
	\begin{centering}
		\includegraphics[width=0.75\textwidth]{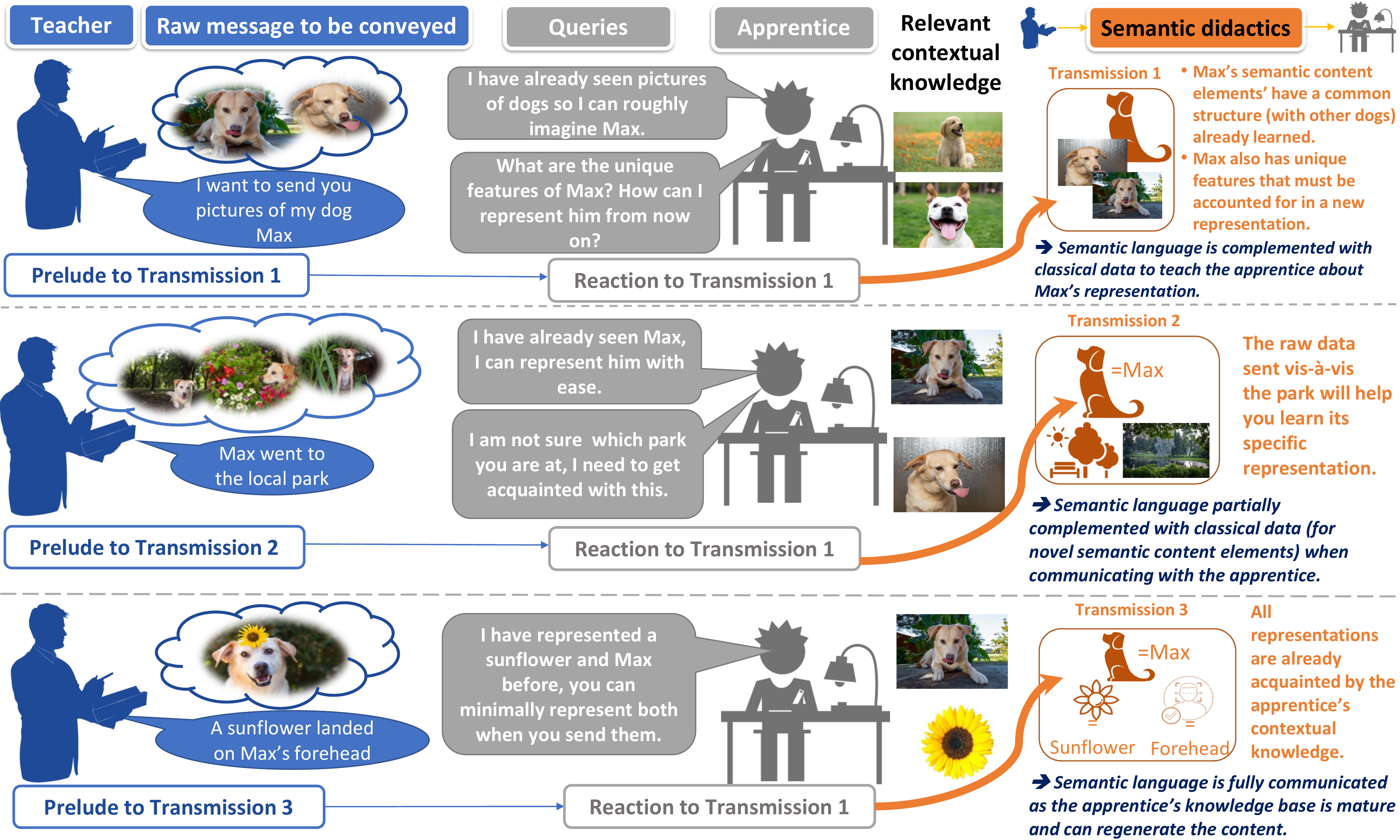}
		\caption{\small{Illustrative example of a reasoning-based communication framework that is gradually building a semantic language whereby they migrate from relying on discrete data elements to organized, linked, and logic inducing knowledge bases.}}
		\label{fig:semantic_convo}
	\end{centering}
\end{figure*}
\subsubsection{Less Data, More Knowledge}
 One key benefit and feature of the semantic language is \emph{minimalism}. This minimalism is exhibited via two mechanisms: First, as will be demonstrated in the sequel of this work in Section~\ref{section:howto}, communicating via a language with tolerable complexity, yet significant structure imposes the \emph{minimal sufficiency condition on representations}. In essence, a semantic representation is \emph{the teacher's minimal description with regards to the meaning held by particular content element in the datastream}. That is, such a representation must be characterized via the \emph{smallest number of bits}, while also being able \emph{accurately describe the semantic content}. Second, given that radio nodes can leverage their knowledge base to perform \emph{reasoning}, the apprentice can draw conclusions based on the context communicated. These conclusions can minimize the number of redundant back and forth messaging. For instance, in an autonomous vehicle setting, instead of tuning the car's direction in real-time and exhaust communications resources, the controller could send a representation that depicts the mode of driving in the future (e.g. drive straight for the next hour in high cautiousness). Here, the apprentice would leverage their computing resources to \emph{generate} how driving straight and in high cautiousness is exhibited (these exhibits are the semantic content elements). Thus, a radio node is operating in a ``less data, more knowledge", whereby the data stemming from representations and back-and-forth messaging is minimal, yet the knowledge base of a radio node is comprehensive and elevates the intellect of this node. Remarkably, this ``less data" realm \emph{minimizes the reliance on the spectrum}, and the need to open new spectrum bands in order to respond to the exponential data rate increases from one cellular generation to another. Thus, one long-term key benefit of semantic communications is a minimization in the need for more spectrum as well as complex dynamic spectrum sharing schemes (technical and regulatory) whenever new wireless technologies or use cases appear.\\
\indent Next, we overview the fundamental measures that must be revisited in classical information theory to transition towards semantic communication systems.

\section{How to transition towards semantic-aware systems?}\label{section_transitioning}
\subsection{From bit transmissions to knowledge-driven human conversation}
One caveat of a classical communication scenario is constrained to a bit agnostic representation of the message. For instance, if the transmitter were to send a photo of a dog to the receiver, then the photo must undergo signal processing steps to finally be represented on the bit and consequently packet level. Then, an erroneous reconstruction of the image at the receiver may result from any singular bit error. These bit errors can stem from errors at the transmitter, channel, or receiver. Also, such errors can have a hardware, software, or networking nature. In other words, a ``bit-error" can result from a hardware/software defect or a networking bottleneck. This phenomenon can be called \emph{data-blindness} at the transmitter, receiver, and air-interface levels. Moreover, given that the reconstruction process is oblivious to the context of the bits, e.g., whether a bit represents a dimension in the background or foreground; the reconstruction process is highly susceptible to these aforementioned bit errors. That is, current error correction schemes that attempt to minimize errors, lack an \ac{AI}-foundation that unravels the root-cause of the errors/bottlenecks in the network and attributes context to it. Hence, such error correction mechanisms are \emph{memory-less} and can only minimize errors based on ``current datastreams". Thus, these schemes cannot leverage recurrent and semantic errors to ultimately improve the long-term system performance. In contrast, if bits were to become aware of semantics and context, the robustness to errors and the intelligence of communication systems would evolve significantly. \\
\indent Furthermore, in a classical communication system structure, Shannon's information theory foundation is based on the \emph{asymmetry of communication} \cite{lopez2019no}. In other words, data at the receiver cannot be created in an ex nihilo fashion. Subsequently, to grant the receiver information generative and reasoning capabilities, one can reframe the communication problem as one in which a pattern or structure (which may or may not be repeated) needs to be realized or constructed in different instances within a limited time duration. This new definition enables reducing the asymmetry between the transmitter and the receiver as we envision for a teacher and apprentice. Here, we ask two important questions that must be answered in order to realize the prior semantic communication definition:
\begin{itemize}
	\item How should the teacher represent information minimally, without jeopardizing the apprentice's understanding of the representation?
	\item What are the steps needed from the apprentice to reason over the received semantic representation and make logical decisions out of it?
\end{itemize}
\indent Answering these questions necessitates defining the communication problem on the premises of human-like thinking, i.e., transforming the bit/data pipeline information exchange into a knowledge-driven semantic conversation. Answering those questions also requires gradually constructing a semantic language between the teacher and the apprentice. In Fig.~\ref{fig:semantic_convo}, we showcase how a teacher and apprentice perform three different transmissions over the course of converging towards a mature semantic language. Fig.~\ref{fig:semantic_convo} also shows the mechanism used by the teacher to \emph{explain} the meaning of conveyed representations. In Fig.~\ref{fig:semantic_convo}, we assume that the teacher has the reasoning capabilities needed to \emph{extract} the semantic content elements, and map them to a proper semantic representation. The details of acquiring this skill are discussed in Section~\ref{lanugae_preprocessing}. During the first transmission, we can see that the teacher complements the semantic representation of choice with raw messages. This enables the apprentice to recognize the variability of the transmitted semantic representations with respect to the apprentice's relevant knowledge. Here, such combination of raw messages and representations that are used during the pedagogic process are dubbed, \emph{semantic didactics}. Then, we can see that, after the apprentice has leveraged causality via queries\footnote{Queries are interrogations posed by the apprentice to learn more information about the causal and associational structure of a representation. In essence, this mimics the classroom learning process, whereby a student asks questions to build their knowledge base on a subject. Queries can be interventions or counterfactuals, and will be detailed in Section~\ref{reasoning_via_causality}.}, semantic didactics contain less raw messages during transmission 2. That is, transmission 2 mostly relies on previously acquired semantic representations, while marginally complementing the unseen representations with raw data. Finally, in transmission 3, we can see that the apprentice has posed the majority of their queries so far, and their relevant contextual knowledge enables them to understand the representations used by the teacher. In this case, we can see that in transmission 3, the teacher solely relied on semantic representations to transmit the information. Consequently, the apprentice can generate the intended image by reasoning over the semantic representations conveyed.\\
\indent Furthermore, we can observe that successfully representing a peculiar structure by the teacher while successfully understanding it at the apprentice depends on a number of factors: a) the relevant contextual knowledge of the apprentice vis-à-vis the current message to be conveyed, b) the capability of the apprentice to represent the current message while leveraging the relevant contextual knowledge, and c) the level of synchronization between the teacher and apprentice, i.e., how well they are acquainted with their representations. For instance, in the example of Fig.~\ref{fig:semantic_convo}, the apprentice has seen pictures of dogs before from other teachers (from previous information exchanges). Nonetheless, the apprentice does not know how to represent Max, and thus two scenarios are plausible here: 1) The apprentice needs to attempt to represent Max given that they have represented a dog before, however this depends on the richness of their knowledge base as well as their reasoning capability, 2) The teacher represents Max for the first time while also complementing the representation with classical data (via semantic didactics); after several transmissions, the apprentice learns how to construct the realization of Max minimally. From our example, we can see that the apprentice asked the teacher via query for more information, proving that scenario 1 might have led to further errors (if such information was not properly requested and delivered). Then, in transmissions 2 and 3, we observe that, within the overall message to be conveyed, Max's structure is \emph{well-defined} for the apprentice. However, we can also see that the apprentice lacked the representation structure to realize the local park. Here, the apprentice needs to conduct a series of queries to be capable of representing the local park appropriately. Finally, in their last interaction in Fig.~\ref{fig:semantic_convo}, we can see that the apprentice has the appropriate relevant knowledge that enables him to interpret the message. Consequently, in such a scenario, minimal representational information is only needed to construct these three well-defined patterns (dog, sunflower, and forehead), already acquired by the apprentice a priori. It is also important to note the following:
\begin{itemize}
    \item As the language gains more maturity, the communication link's symmetry increases. As such, in a symmetrical setting, the maturity of the language as well as the enhancement in the reasoning capabilities enable the link to be minimally reliant on the channel and communication resources. That is, as explained in Section~\ref{subsub_robustchannel}, the apprentice can leverage their knowledge base to understand the cause of particular errors caused by the channel. Also, they can rely on a few representations to generate the remaining content via their computational generative capabilities.
    \item Our example in this exposition was limited to a visual dog example, nonetheless, this is only for simplicity and for illustrative purposes. The same analogy can be extended and generalized to any \emph{structure} in the data.
\end{itemize}
Building on our intuition from this example, next, we investigate the necessary measures to migrate from information theory to semantic theory.
\subsection{From Information Theory to Semantic Information Theory}\label{subsection_informationtheory_tosemantictheory}
Fundamentally, as shown in our illustrative example in Fig.~\ref{fig:semantic_convo}, while the semantic didactics (a combination of semantic representations and classical data transmitted) are highly dependent on the capabilities of the teacher and apprentice, each unique semantic representation hinges on the \emph{complexity of the task to be described}. A task here is defined as the process of extracting the semantic content elements from a specific datastream, and then mapping each semantic content element to a corresponding semantic representation. In other words, there exists a relationship between the complexity of the task, the number of semantic representations that must be exchanged, and consequently the number of interventions required for an overall comprehension of the apprentice. Nonetheless, given that information theory is solely focused on characterizing uncertainty in the message rather than meaning, it is not the right tool for capturing or characterizing all of these semantic-based concepts. It is thus necessary to propose a suite of novel equivalent fundamentals that build on information theory, and that are tailored to reasoning-driven semantic networks. In essence, to characterize the complexity of a semantic language, we first investigate the way Shannon defined the concepts of information and entropy. This is fundamentally important as designing any communication network and evaluating its performance depend on the way Shannon defined these concepts. Building on these concepts, we similarly draw parallels from classical information theory to future semantic information theory. Such equivalent preliminaries and fundamentals enable us to understand the operation, design, and performance of future semantic communication system in terms of building a semantic language and reasoning over semantic representations. 
\\\begin{figure*} [t!]
	\begin{centering}
		\includegraphics[width=0.80\textwidth]{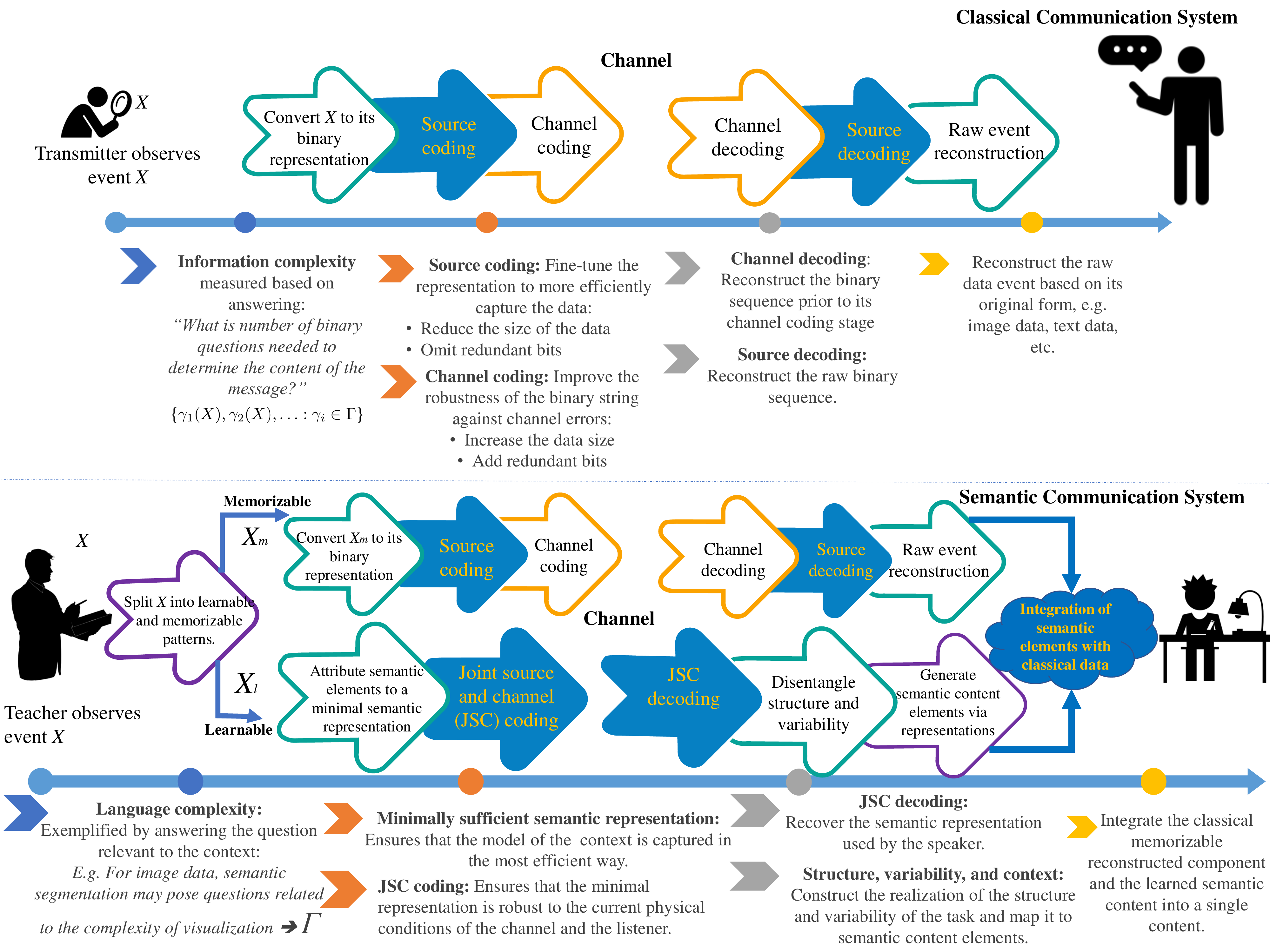}
		\caption{\small{Illustrative example showcasing the transformation of an \ac{E2E} communication system from a traditional setting to a semantic-aware one.}}
		\label{fig:comprehensive_transform}
	\end{centering}
	\vspace{-0.5cm}
\end{figure*}
\subsubsection{Information: From Uncertainty to Semantic Substance}\label{semantic_substance} Shannon defined information based on its combinatorial nature rather than on its meaning. In other words, under infromation theory, information characterizes what a transmitter, based on the way bits flow from one medium to another, \emph{could} say, bearing in mind source, channel, and destination errors. Hence, Shannon defined the amount of information according to the number of combinatorial choices that can be considered by the transmitter, i.e., in the simplest cases, the logarithm of the number of available choices in which one can transmit the message. If we let $X=x$ be the observed event by the transmitter, the information of observing $x$ is defined as \cite{ash2012information}:
\begin{equation}\label{information}
I(x)=-\log _{2}[p(x)],
\end{equation}
The logarithm function here enables measuring information in an additive fashion with respect to the number of states $p(x)$ of the system. For instance, if the transmitter wishes to transmit the number $5,555,555$, roughly $23$ bits are needed to represent this rather simple identical sequence of numbers. Interestingly, one can notice that, if a person were to say the exact same sequence to another, they would leverage the existing pattern and say: \emph{the number $5$, $7$ times}. Leveraging this pattern would inherently minimize the number of bits needed from $23$ to almost $10$ bits. Alternatively, if a string of numbers with the same length, such as $540,938,1326$ were to be transmitted then, again, the transmitter needs $23$ bits to send them across the channel. In contrast to the previous example, given that this sequence of bits does not exhibit any pattern, as humans we ought to memorize it in our short-term memory before we convey it to the listener\footnote{It is important to note that while the structure in the exhibited example is fairly simple and can be captured via source coding to minimize redundancy, it was used here for exposition purposes in conveying the existence of \emph{long-term and recurrent} structure in the datastream.}. Notably, from these simple, but insightful examples, we can make the following key observations:
\begin{itemize}
	\item When the message to be transmitted has inherent patterns that can be leveraged, relying on a reasoning mechanism enables to transform the process from a mere recovery process, to a human-like exchange that can potentially minimize communication resources substantially.
	\item While it was fairly easy to leverage the identical series of numbers in the first example, the second example seemed like an arbitrary sequence of numbers that does not exhibit any regularity. However, this was because the examples were taken \emph{out of context.}  The sequence could actually map to a phone number whereby the first three digits map to the state of Virginia area code, the following three digits map to the central office, and the last four digits map to the line number. Here, attempting to learn this task might not enable us to directly minimize communication resources. That is, in analogy to our dog example in Fig.~\ref{fig:semantic_convo}, the apprentice would require the teacher to complement the representation with raw data so that the context and the representations are acquired.  Then, moving forward from that point, the apprentice is cognizant of context, and, thus, they can utilize their knowledge base to correct any \emph{structurally inconsistent errors} in the semantic representations that map to a phone number or to information that should \emph{logically} accompany such numbers. This will also enable the communication link to be more minimal by not communicating every digit. 
\end{itemize} 
If we further apply the insights we have made on the illustrative example of Fig.~\ref{fig:semantic_convo} we can see that, in a classical communication system, a $3,300 \times 4,800$ image will still carry identical information regardless of the simplicity, redundancy, or complexity of the semantic content elements in the picture. This showcases how ``information" in Shannon's definition characterizes what the pixels \emph{could} represent in binary. Meanwhile, if one focuses on the semantic content elements of the image, the information transfer becomes a function of the complexity of the message as well as the maturity of the language established. That is, in the case of a message with an \emph{obvious} structure, like the repeated series of numbers, as well as a mature semantic language, the information transfer process becomes minimal, and it can be easily communicated. Meanwhile, in the case of a complex message, and a weakly-established language between the teacher and the apprentice, the information transfer process may \emph{first} waste communication resources to establish the language. Then, the teacher/apprentice benefit from the knowledge base and language built to ultimately reach an efficient, generalize, and minimal link.\\
\indent Consequently, Shannon's ``information" is technically quantifying what has been historically known as \emph{syntactic information} \cite{kolchinsky2018semantic}. Syntatic information quantifies how much the knowledge of the state of one system reduces the statistical uncertainty about the state of the other system, possibly at a different point in time. In a communication setting, Shannon's information theory measures how much knowledge about the transmitted datastream reduces the statistical uncertainty about the state of the received datastreams. While these statistical correlations between the transmitted and received datastream are important, Shannon's notion of information does not consider what such correlations mean \cite{kolchinsky2018semantic}. In contrast, in order to introduce a meaning and context to the definition of information one must: a) Highly correlate the link between the definition of information and the overall goal of the system (if and when such unified goal exists), and b) Capture the relationship between information and causality. Subsequently, we first define the concept of \emph{``semantic information"} for goal-oriented semantic communication systems and then generalize this concept for all semantic communication systems:
\begin{definition}
A particular datastream $x$ is said to be rich in semantic information if transmitting it improves the system's ability of pursuing its specified goals. 
\end{definition}
For example, the goal of a system of digital twins \cite{hashash2022} is to guarantee a high synchronization between the physical and cyber twin. Thus, the information exchanged is deemed to be a valuable semantic if it is capable of enhancing the performance of the digital twin in terms of reliability and synchronization. As discussed, in many use cases, a unified goal might not exist. In such settings, we extend the definition of semantic information to what follows:
\begin{definition}\label{semantic_info}
A particular datastream $x$ is rich in semantic information if it improves the reasoning capabilities of the teacher and the apprentice to ultimately expand the semantic language built between them.
\end{definition}
The concept of semantic language is explained, in depth, in Section~\ref{struct-var}. Essentially, Definition~\ref{semantic_info} captures the fact that a datastream is void of information if it does not contribute to enhancing the reasoning capabilities of the teacher and the apprentice. The improvement in reasoning capabilities can be achieved by leveraging causality, and properly identifying the root-cause of new semantic content elements, based on previously observed ones. Also, similarly to humans, when our learning improves, our capability to express our understanding also improves. Hence, a semantic language is a key element that captures the capability of a radio node to ultimately understand the meaning of information. As such, the previously established definitions of semantic information are the initial guide in the process of defining a semantic language and its complexity, compared to entropy. Next, we investigate the information theory fundamentals of entropy, then, we further discuss the necessary measures to define its counterpart in semantic communication systems.
\subsubsection{From Entropy to Language Complexity}\label{subsection_entropy}
The overarching role of entropy is to characterize the uncertainty of a micro-state on the macro-state of an overall considered physical or statistical system. Stated differently, statistical entropy is proportional to the number of ``yes/no" questions that must be asked, to determine the micro-state, given that the macro-state is known. In classical communications, the micro-state or atomic unit is a ``bit", consequently, if we let $\Gamma$ be the set of all possible binary functions on a domain $X$, characterizing the entropy requires asking the following question: \emph{What is the optimal sequence of yes/no questions needed to construct}: 
\begin{equation}
	\left\{\gamma_{1}(X), \gamma_{2}(X), \ldots: \gamma_{i} \in \Gamma \right\},
\end{equation}
with the goal of perfectly recovering $X$ from the shortest sequence of binary answers \cite{gray2011entropy}. \\  
\indent Furthermore, based on the previous definition of \emph{information} in \eqref{information}, answering this question requires computing the entropy over the domain $X$ in the classical sense, as follows:
\begin{equation}\label{classicalentropy}
H(X)=-\sum_{i=1}^{n} p\left(x_{i}\right) \log p\left(x_{i}\right).	
\end{equation}
From~\eqref{classicalentropy}, we observe that the entropy varies with respect to the freedom of choice of the transmitter. That is, $H$ tends to $1$ when a single probability measure $p(x_i)$ is equal to $1$, and all other probabilities are equal to $0$. In this case, it means that one can \emph{certainly} represent the message deterministically. However, the entropy is independent of the message itself. We can also observe that, when we are in an equiprobable setting, then entropy is at its highest. Meanwhile, if the states are not equiprobable, compression becomes more possible \cite{kolchinsky2018semantic}, and thus fewer yes/no questions need to be asked to identify a particular input of datastream. Nonetheless, in all cases, i.e. whether compression is considered or not, entropy remains a function of the choices of the transmitter and not a function of the message itself.\\
\begin{figure*}
	\begin{centering}
		\includegraphics[width=0.90\textwidth]{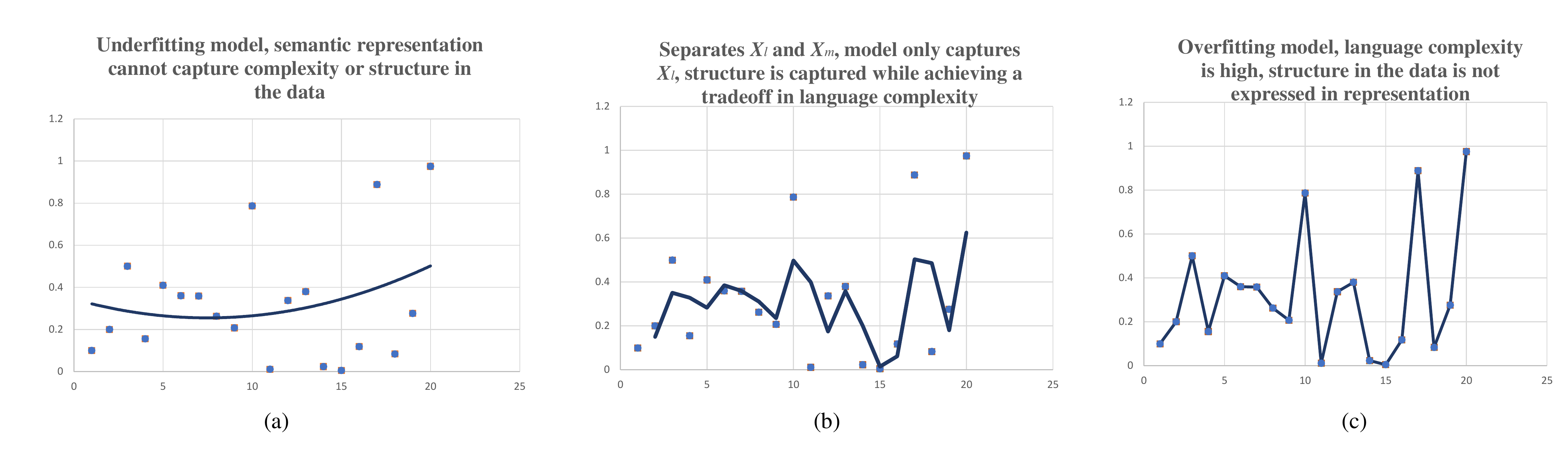}
		\caption{\small{Illustrative example showcasing the difference between (a) an overly simple model, underfitting over the raw data, (b) a reasonable model that separates $\boldsymbol{X}_l$ from $\boldsymbol{X}_m$, captures the semantic data, and leaves out random data, and (c) an overfitting model that captures all the characteristics of the data without exhibiting reasoning capabilities, and without capturing structure.}}
		\label{fig:overfitting_underfitting}
	\end{centering}
\end{figure*}
\indent Interestingly, if we consider the way humans communicate: when trying to explain something, we do not attempt to do so while considering the number of yes/no questions required to explain a particular story. Such yes/no questions fail to characterize the \emph{structure} of the story itself. Meanwhile, we tend to focus on the most \emph{meaningful} information that constitutes the \emph{core} of the story.  To do so, the set $\Gamma$ needs to be restructured so that it consists of questions regarding the content complexity (which highly correlates with the semantic content elements) rather than uncertainty (which is devoid of it). One can for example let $\Gamma $ represent the visual semantic entropy \cite{huang2021deep, gune2020generalized}, whereby the set of queries would indicate the presence or absence of an entity, and its relation to the other represented entities. From our Fig.~\ref{fig:semantic_convo} example, a simple question to semantically represent transmission 1 would be ``Is there a dog at the center of the picture?". Meanwhile, semantically representing transmission $3$ requires asking the question ``Is there a dog with a sunflower on its forehead?". We can see that the latter has a substantially higher reasoning complexity as multiple entities are involved, and they are related to one another via a certain function, and there exists a function between them (the sunflower is exactly on the forehead). Indeed, these questions enable us to characterize the complexity of a task to be semantically represented, nonetheless, the problem is that the set $\{\gamma\}_{\gamma \in \Gamma}$ cannot be computed tractably in this case.\\
\indent In essence, we need an alternative to entropy that enables exchanging questions between the transmitter and receiver to ultimately capture structure in the information and reflect the semantic content of the datastream. Consequently, this exchange reflects the necessity for engineering a semantic language between the transmitter. Here, \emph{the concept of language complexity in a semantic-based system is the equivalent of entropy in a classical one}. Next, based on the nuts and bolts we have elucidated to extend classical information theory to semantic information theory, we will investigate the fundamentals of semantic representations and languages for semantic communication systems.
\section{Building Semantic Communication Systems: Representations and Languages}\label{section:howto}
As shown in Fig.~\ref{fig:comprehensive_transform}, classical communications starts by first processing the raw data and converting it into binary representation. Then, this binary representation is source coded to minimize the number of redundant information bits. Then, this resulting datastream is channel coded to improve its robustness against the adverse channel conditions. At the receiving end, such processes are mirrored to ultimately reconstruct the raw data message originally transmitted. As discussed in Section~\ref{section_preliminaries}, all of these processes use Shannon's concept of ``information", viewed as ``uncertainty", and, thus they do not capture any core structure in the data. Moreover, they are not able to attribute any meaning to the datastreams processed. In this section, we investigate the concept of structure and variability in the data and, then, we highlight the necessity of splitting the data into \emph{learnable and memorizable} patterns. Then, we will discuss the properties of a semantic representation and language. We will further gradually explain the semantic communication system building blocks shown in Fig.~\ref{fig:comprehensive_transform} while we explore these novel concepts in this section.
\subsection{Language pre-processing}\label{lanugae_preprocessing}
\begin{figure}
	\begin{centering}
		\includegraphics[width=0.45\textwidth]{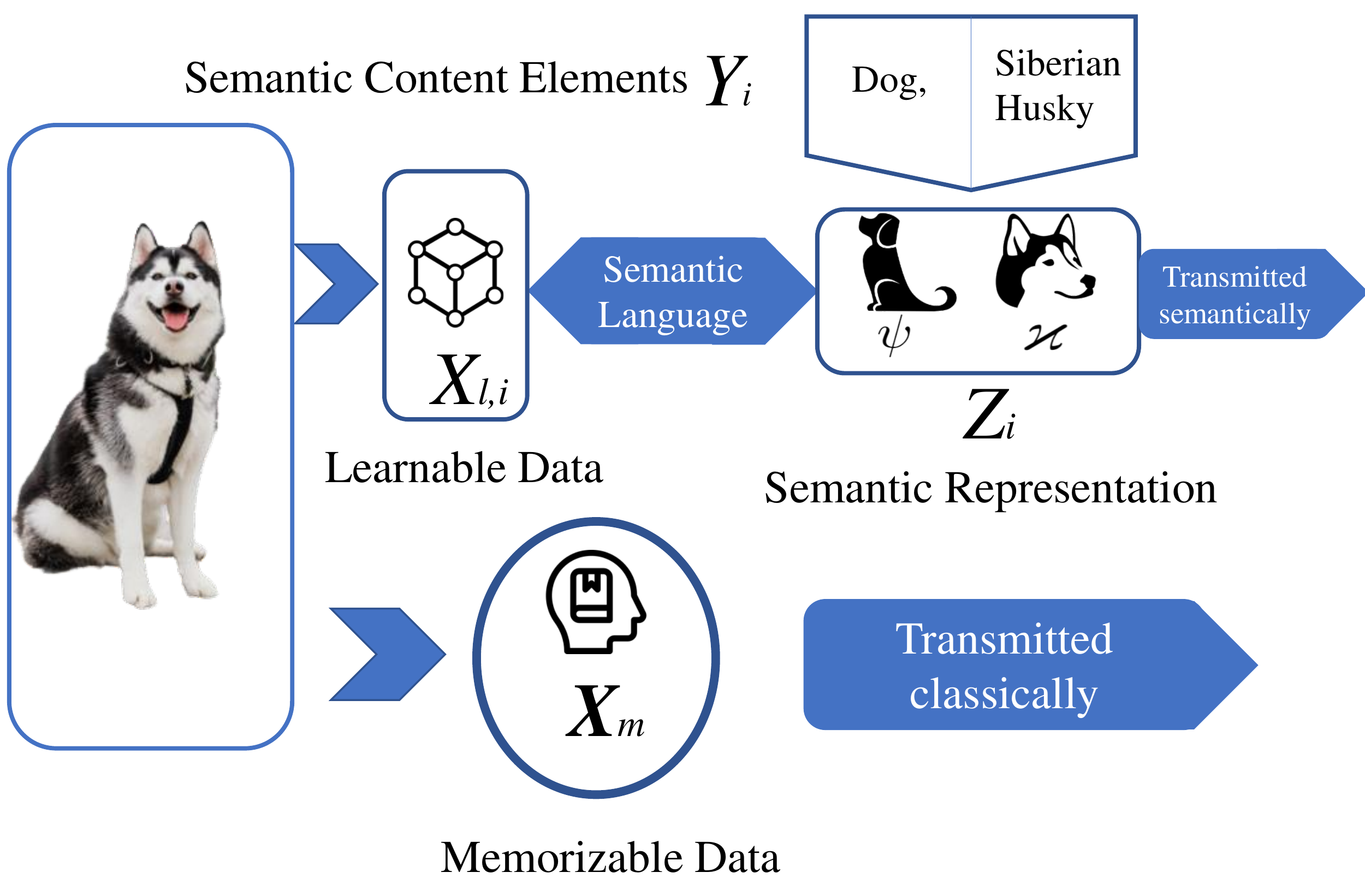}
		\caption{\small{Illustrative example showcasing the concept of semantic language.}}
		\label{fig:language_explained}
	\end{centering}
	\vspace{-0.25cm}
\end{figure}

In any communication system, the first step is to convert the highly dimensional raw data into binary data suitable for communication. While in classical communications this is customary and followed by source and channel coding, in a semantic communication network, we must scrutinize the data, and attribute a representation to every major structural part in it. Nonetheless, if one considers the whole datastream as a bulk, the learning process, i.e., identifying every semantic content element and crafting its corresponding semantic representation, becomes highly inefficient. This is due to the fact that raw data contains a lot of purely random information. This is problematic because such random information increases the complexity of the built language (which we will soon formally define), yet contributes to a language with spurious semantic representations. That is, one might think that the semantic language complexity stems from an inherently complex semantic structure (e.g. a complex high-dimensional hologram that must be described via a semantic language). In reality, the deceiving complexity measure here stems from an abundance of random information in the datastream. \\
\indent Hence, for the sake of efficiency, prior to learning a semantic language, we must first separate two parts of the data, learnable and memorizable, as defined next:
\begin{enumerate}
    \item The \emph{learnable part $\boldsymbol{X}_l$ of the data:} is the core of semantic information. That is, performing \emph{reasoning} on such learnable data points allows the radio node to capture the inherent structure in each semantic content element. Thus, reasoning over $\boldsymbol{X}_l$ eventually leads to a meaningful semantic language with non-spurious models. One can use $\boldsymbol{X}_l$ as an input to build a proper semantic language. This is shown in Fig.~\ref{fig:language_explained}.
    \item The \emph{memorizable part $\boldsymbol{X}_m$ of the data:}  is the one that is governed only by pure randomness. For example, these can be details in an image that do not contribute to any semantic content element. As such, attributing a semantic representation to $\boldsymbol{X}_m$ is a highly \emph{complex} learning process, but a very \emph{simple} memorization process. In other words, learning the structure of a purely random and structure-less datastream is \emph{complex} process and does not yield meaningful semantic representation. Thus, allocating computing resources to \emph{learning} random information is a wasteful process, thus $\boldsymbol{X}_m$ must be transmitted using classical communication resources.
\end{enumerate}
Furthermore, as shown in Fig.~\ref{fig:overfitting_underfitting}, separating data into memorizable and learnable components is crucial. If one attempts to learn all the data points as shown in Fig.~\ref{fig:overfitting_underfitting} (c), the resulting semantic representations learned will be spurious. While the curve here ``fits" the data, it does not learn any inherent structure.\footnote{In fact, the overfitting scenario shown in (c) is what source coding and data compression attempts to do, whereby they just aim to summarize data and then reconstruct it bit-by-bit. The compression does not identify any structure} In this case, reasoning on all the datastream, will not reflect the true semantic content elements of the data. This will lead to an inherently poor semantic language that cannot express structure in the data. Meanwhile, if one only captures the learnable part of the data as shown in Fig.~\ref{fig:overfitting_underfitting} (b) to build a language, the learning process would not overfit. It will capture all data points as its goal is to capture structure and not \emph{memorize} information. It is also important not to have an overly simplified language (Fig.~\ref{fig:overfitting_underfitting} (a)) that will fail to yield a minimal representation that characterizes the structure of data. We will further revisit Fig.~\ref{fig:overfitting_underfitting} when investigating the semantic language complexity. It is important to note that, in classical \ac{ML} frameworks, the data is not separated or pre-processed as the goal of \ac{ML} is fundamentally different from the overarching goal of semantic communications. In essence, in \ac{ML} the goal is \emph{learn} the data and then leverage this learned-model to make some predictions. Nonetheless, in semantic communication the goal is to perform \emph{reasoning} and dissect the structure of semantic content elements. A semantic language and a knowledge base cannot be built via a series of models and bulks of data (see Fig.~\ref{fig:datatoknowledge}), but they can only be built via organized knowledge (e.g. causal reasoning schemes and languages).\\
\indent Moreover, as shown in Fig.~\ref{fig:comprehensive_transform}, in contrast to a classical communication system, the first step in semantic communication systems is to split the observed data into learnable and memorizable parts. Then, the path followed by $\boldsymbol{X}_m$ is identical to the one of classical communications. To understand this  more clearly, $\boldsymbol{X}_m$ mimics data points in our daily lives that are \emph{wasteful} to learn. For instance, when trying to call someone, the digits following the area code are purely random, they do not follow a specific pattern nor do they result from a causal event. Thus, in such scenarios, our mind operates like a bit-pipeline that captures the information without performing any reasoning over it. Note that the separation between $\boldsymbol{X}_m$ and $\boldsymbol{X}_l$ is not unique, which makes this process of separation challenging. Here, one can resort to \ac{ML} schemes to perform this categorization \cite{lecoq2021removing, zhu2019heavy}, or can leverage causal models as discussed in Section~\ref{section:howto}. That said, these disentanglement methodologies remain outside of the scope of this work.\\
\indent We can also see in Fig.~\ref{fig:comprehensive_transform}, that for the semantic path, $\boldsymbol{X}_l$ first undergoes a \emph{reasoning} process whereby its semantic content element and representation are identified. Then, every semantic representation, and ultimately the semantic language, undergoes joint source and channel coding. This is intrinsically different to classical communications whereby the \emph{source data} undergoes joint source and channel coding. Here, the goal of joint is to leverage the correlation between multiple semantic content elements (or even for multiple transmitters) for cooperative transmission \cite{fresia2010joint}. In fact, as laid out in \cite{farsad2018deep} and \cite{bourtsoulatze2019deep} joint source and channel coding enable executing specific actions (in a goal-oriented scheme) directly rather than recovering the entire source messages. Similarly, in a broader semantic communication system, joint source and channel coding schemes enable re-generating (at the destination) the specific semantic content elements rather than the whole datastream. This is necessary as the receiver no longer decodes one bit stream at a time, but rather generates one semantic content element (meaning) at a time.

\subsection{On the Structure and Complexity of a Semantic Language}\label{struct-var}
Thus far, we have pre-processed the data and separated $\boldsymbol{X}_m$ from $\boldsymbol{X}_l$. In the sequel, and since $\boldsymbol{X}_m$ will undergo the path of classical communication, we focus on building a language from input $\boldsymbol{X}_l$. The concept of a semantic language can be seen in Fig.~\ref{fig:language_explained}. We first define a semantic language and then scrutinize its structure and variability to ultimately assess its complexity.
\begin{definition}
A \emph{semantic language} $\mathcal{L}={(X_{l,i}, Z_i)},$ is a dictionary (from a data structure perspective) that maps the learnable data points $X_{l,i}$ to their corresponding semantic representation $Z_i$, based on the identified semantic content elements $Y_i$.
\end{definition}
Representation $Z_i$ must be efficient in inducing the apprentice to \emph{generate} the originally conveyed task or semantic content $Y_i$. Thus, language $\mathcal{L}$ re-purposes the apprentice's task from a mere \emph{reconstruction} mechanism to a \emph{generative and automatic} process. That is, the language re-purposes information transfer via the convergence of computing and communications in semantic systems, thus yielding a \emph{generative apprentice and a dictating teacher} as follows:
\begin{itemize}
    \item \textbf{Generative Apprentice:} The apprentice relies on their \emph{computing resources} and reasoning faculties to \emph{generate} content from a \emph{representation}. This mimics our imaginative experience when someone mentions a particular term. That is, when someone says ``flower" we can ``imagine" and ``recreate" what a flower looks and feels like, based on its representation. For instance, in an \ac{XR} setting, if the transmitter is willing to augment the metaverse with new \ac{XR} content which is a flower on the sidewalk, they have to modify every bit surrounding the metaverse 3D space so as to reconstruct a flower on the sidewalk. Meanwhile, in a semantic setting, the teacher ``describes" to the apprentice that they need ``add a flower, on the left corner of the sidewalk". Based on their previous language, the apprentice maps findings in their knowledge base to what is ``flower", ``left", and ``sidewalk", to generate it via its computing resources.
    \item \textbf{Dictating Teacher:} The teacher does not have to exhaust multiple communication resources to convey a message in a \emph{bit-by-bit} fashion. Instead, the teacher must: a) identify the semantic content elements in the data, i.e., learn the meaning in the datastream (via their computing resources), b) based on this meaning, the teacher must attribute an existing representation or develop a new representation, and add it to their language, and c) convey the message based on the representations and language developed.  
\end{itemize}
Essentially, each semantic representation within language $\mathcal{L}$ must be able to:
\begin{enumerate}
    \item Characterize the \emph{structure}, $\psi$, that describes the data collectively. For example, this could be the main semantic element of an image. For instance, if a semantic representation describes a German Shepherd, the common denominator between the German Shepherd and all dog breeds represents the \emph{structure}. Naturally, this concept applies to any type of data, and not just images.
    \item Characterize the \emph{variability}, $\varkappa$, of the individual data points with respect to the shared structure. For instance, taking the example of a German Shepherd, variability characterizes distinctive features of a German Shepherd compared to other dogs. 
\end{enumerate}
\indent Furthermore, we can define the \emph{complexity of the language} $\Gamma(\mathcal{L})$. The goal of this complexity is to characterize the difficulty of identifying and learning the semantic content elements within $\boldsymbol{X}_l$. This process ultimately stems from the inherent structure of the data (which we will formally define shortly). This complexity also captures the difficulty of developing a language for this semantic content, i.e., a semantic represenation for every semantic content element. Our framework has been built by borrowing the definition of dataset complexity in transfer learning in \cite{achille2021information} and \cite{NEURIPS2021_6738fc33}. In the following proposition, we characterize this language complexity. 
\begin{proposition}\label{propp} The complexity of a specific language $\mathcal{L}$ adopted among a teacher and apprentice pair is given by: 
\begin{equation}
    \Gamma(\mathcal{L})=\min_{p(Z|\boldsymbol{X}_l)}L_\mathcal{L}(p) + K(p).
\label{complexity}
\end{equation}
\end{proposition}
Here, $L_\mathcal{L}(p)=\sum_{i=1}^N-\log p(Z_i|X_{l,i})$ is the cross-entropy loss, and $K(p)$ is the Kologomorov complexity of the distribution $p(\boldsymbol{Z}|\boldsymbol{X}_l)$.\\
\indent From Proposition~\ref{propp}, we observe the following:
\begin{enumerate}
    \item When learning random semantic-agnostic representations for the data, the complexity $\Gamma(\mathcal{L})$ will become very high. In essence, this is a result of the fact that the majority of the data \emph{lacks structure}. Thus, in that case, the ``task" of concern should be memorization rather than learning (using a language to communicate random information is a highly inefficient task). Hence, the teacher must prune data points from $\boldsymbol{X}_l$ and attribute them to $\boldsymbol{X}_m$, i.e., the pre-processing phase must be revisited. This will in turn minimize the complexity $\Gamma(\mathcal{L})$ and makes learning the language more efficient.  
    \item The language complexity is a metric that replaces the classical entropy. It is a function of the cross-entropy loss, whereby the fitness of the model is captured as well as the Kologomorov complexity of the data. Here, in contrast to entropy which only characterizes the uncertainty and the freedom of choice at the transmitter (which we highlighted in Section~\ref{subsection_entropy}), language complexity measures the complexity of the raw data and the language model at stake when communicating a particular message.
    \item Unlike Shannon's information-theoretic perspective whereby the encoding for a message is predetermined by the randomness of the source transmitting it, Kolmogorov's complexity enables us to characterize the \emph{individuality} of the semantic content elements in the message to be conveyed. In fact, Kolomgorov's complexity $K(x)$ is a measure of the shortest effective binary description of $X$. In other words, it characterizes the methodology that enable the apprentice to \emph{autonomously generate} the semantic content elements based on the conveyed semantic representations.
\end{enumerate}

Essentially, the \ac{PDF} $p$ in Proposition~\ref{propp} constitutes the learned model of the semantic representation of a particular semantic content element, based on input $\boldsymbol{X}_l$. Consequently, it is necessary to characterize the tradeoff between the \ac{ML} loss achievable by this model for a language $\mathcal{L}$ and its complexity. This tradeoff essentially is a \emph{measure} that indicates to the teacher/apprentice: a) the maturity of the established language, based on how expressive it is with respect to semantic content elements, and b) the language complexity which will indicate the level of reasoning required, as well as the amount of computing resources needed to achieve this language. This tradeoff is captured via the \emph{structure function:}
\begin{definition}
The structure function achievable by a model $p$ for a language $\mathcal{L}$ is given by \cite{achille2021information}: 
\begin{equation}
    \Psi_{\mathcal{L}}(t)=\min_{K(p) \leq t} L_{\mathcal{L}}(p).
    \label{structure}
\end{equation}
\end{definition}
The structure function $ \Psi_{\mathcal{L}}(t)$ will be zero for sufficiently high complexity. In particular, after all the shared structure is captured and characterized, the only methodology that minimizes the loss in it is by memorizing the variability of the leftover data bits, leading to an overall high structure function.\\ 
\indent To solve the optimization problem in \eqref{structure}, one can rewrite \eqref{structure} with respect to its associated Langragian: $L_\mathcal{L}(p)+ \lambda K(p)$, where $\lambda$ is the Langrangian multiplier. Furthermore, taking the minimum over $p$ leads us to obtain a family of complexity measures for our language $\mathcal{L}$ parameterized by $\lambda$:
\begin{equation}
    \Gamma_\lambda(\mathcal{L})=\min_{p(\boldsymbol{Z}|\boldsymbol{X}_l)} L_\mathcal{L}(p)+\lambda K(p).
\label{lagrangian}
\end{equation}
\eqref{lagrangian} is nothing but the Legendre transform of the structure function $\Psi(t)$ as a function of $\lambda$. Solving \eqref{lagrangian} can be performed by increasing the complexity $K(p)$ of the model until the return obtained is smaller than the constant $\lambda$ chosen by the teacher.\\
\indent Moreover, the model $p(\boldsymbol{Z}|\boldsymbol{X}_l)$ is considered a Kolmogorov sufficient statistic of the language $\mathcal{L}$ if it minimizes the complexity formulated in Proposition~\ref{propp}. This reconfirms our initial rationale in Fig.~\ref{fig:overfitting_underfitting} motivating the need for a model that acts as the smallest statistic that can characterize a particular semantic representation. Thus, to ultimately reach a semantic representation that is minimal and efficient, the statistic $p$ must be able to learn and build the language $\mathcal{L}$ without squandering computing and communication resources to model random, non-semantic information. Instead, the minimally sufficient statistic $p$ should be able to characterize the valuable and semantic information within the data, i.e., $\boldsymbol{X}_l$ as previously shown in Fig.~\ref{fig:overfitting_underfitting}.
\subsection{Why Semantic Languages? Why not Natural Languages?}\label{near-optimal sem}
In Section.~\ref{struct-var}, we have discussed the necessity to categorize the \emph{learnable and memorizable} data patterns in order to create a more efficient \ac{E2E} semantic communication system. Then, based on the learnable data categorized, the teacher and apprentice need to create a language $\mathcal{L}$ that yields semantic didactics that have a particular structure and variability. Essentially, the complexity of learning $X_l$ mainly depends on the structure function of choice. A semantic language with an extremely high complexity $\Gamma(\mathcal{L})$ can be a result of:
\begin{enumerate}
    \item A poor separation between $\boldsymbol{X}_l$ and $\boldsymbol{X}_m$. This leads to an  $\boldsymbol{X}_l$ with an inherently poor and unlearnable structure function. Thus, in this case, the teacher needs to revisit the methodology they are adopting to categorize $\boldsymbol{X}_l$ and $\boldsymbol{X}_m$. 
    \item Given a particular $\boldsymbol{X}_l$ that possesses a rich structure function, the teacher needs to revisit the methodology adopted to solve the optimization problem in \eqref{lagrangian}. In this case, the high complexity can be a result of not finding \emph{a proper sufficient statistic of the language $\mathcal{L}$}.
\end{enumerate}
We have so far investigated the fundamental notions of a semantic language and its complexity. We have discussed how a language re-purposes the information transfer with a \emph{dictative teacher} and a \emph{generative apprentice}. Here, given that this information transfer mimics human conversation, one might think that natural languages are the solution that can link the teacher and the apprentice. This misconception has been pervasive in the semantic communication literature~\cite{xie2021deep, xie2022task}, and \cite{zhou2021semantic}. Nonetheless, equipping radio nodes with a natural language constrains the information transfer process with various challenges that include wording and deterministic syntax rules. Also, a semantic representation must be characterized with features that are fundamentally different than words (which are the atomic unit of a natural language). Next, we formalize our ideas further by highlighting the common denominator and contrasting features between a natural language and a semantic-centric language. Then, we highlight the key characteristics of a semantic language and representation.
\subsubsection{Properties of a Natural Language}
A semantic representation is the atomic unit of a language. Thus, to formalize the concept of a semantic representation, we first contrast the differences between a natural language and a semantic language. In general, a \emph{natural language} must have the following properties:
\begin{itemize}
    \item \textbf{Syntax:} This is a system of rules constructing the possible grammatical and acceptable sentences out of words (symbols), and determining their sequential arrangement to create a well-formed sentence (expressions).
    \item \textbf{Semantics:} This is a system that attributes a meaning to a well-formed sentence (expression) constructed according to a particular syntax.
    \item \textbf{Pragmatics:} This is a system that that specifies how the semantically constructed syntax in a language can be used. In other words, pragmatics are the foundation of the context-dependent features of a language. For example, if someone asks the question, ``Are you wearing your seatbelt?'', this is a sentence that urges the passenger to exercise cautiousness, although the word ``cautiousness" has not be used in the sentence.
\end{itemize}
Indeed, every natural language requires the aforementioned properties to enable a smooth communication between human beings. Nonetheless, while semantic communication systems should mimic human communication, their goals and operation will be slightly different. Next, we emphasize the distinctive features of a semantic language, and we highlight why it is fundamentally different from a natural language.
\subsubsection{Goals and Properties of a Semantic Language}
Building a language between the teacher and the apprentice in a semantic communication system does not necessarily require it to be a natural language. Thus, in what follows we elaborate the standing of a semantic language with respect to the key properties of a natural one: 
\begin{itemize}
    \item \textbf{On syntax:} Relying on syntax and a set of grammatical rules to construct meaningful expression of semantic representations defies the initial purpose of semantic communication systems. In other words, the \emph{structure} of syntax is characterized by the properties of the deterministic rules set. Meanwhile, the goal of a semantic language is to characterize the causal and statistical properties of the datastream via a semantic representation. Thus, the core of a semantic representation is the structure of the data to be transmitted and not a set of deterministic rules. Hence, adding syntax will only add an overhead of deterministic rules to the established semantic language. 
    \item \textbf{On semantics:} Inherently, as the name suggests it, the goal of a semantic language is to transmit information by focusing on its meaning. Hence, as will be elaborated next, a semantic language aims to be simpler, by avoiding the strict formalities of a natural language. This similar to when two people are used to each other's vocabulary and avoid formal language. This ultimately \emph{minimizes back-and-forth messages exchanged.}
    \item \textbf{On pragmatics:} Relying on pragmatics asymptotically requires reading between the lines. While this is a consequence of any natural language, it is a sufficient condition but not a necessary one in a semantic language. In semantic communication networks, pragmatics are highly correlated to the context or general theme of communication related to the current semantic messages communicated. Indeed, this skill improves the ``dynamic" reasoning capability of the teacher-apprentice pair. Dynamic reasoning is characterized by generalizing the reasoning performed on one scenario to various different settings (e.g. extract location-based intelligence and use it for tasks such as environmental sensing). However, this skill can only exist when the language has matured. In other words, this skill emerges when the language's complexity has been minimized while capturing structure, i.e., the problem in~\ref{structure} has been solved. This also needs to be accompanied with a known (or non highly varying) context of communication. The sentence ``Are you wearing a seat-belt?" urged cautiousness in the language when the ``context" of communication is known to be in a vehicle. In fact, the concept of pragmatics justifies why a \emph{consistency in the context of subject} improves the dynamic reasoning capability of semantic radio nodes.
\end{itemize}
Consequently, given the fundamental differences between natural languages and semantic languages; next, we delineate the key properties of a semantic representation and its respective semantic language (representations are the atomic constituents of a language).
\begin{itemize}
    \item \textbf{Minimalism:} Based on our observations from Proposition~\ref{propp}, a semantic representation needs to be \emph{minimally sufficient}. That is, if a semantic representation is very simple, it has traded-off part of the structure of the message to minimize the communication resources. Meanwhile, a semantic representation that is overly complex, will have an overfitting \ac{AI} model, i.e., high complexity and low structure. Such a representation would be describing unnecessary random information in the data. 
    \item \textbf{Efficiency:} While a natural language does not mandate efficiency, a conversation between a teacher and an apprentice in a semantic communication network must be more efficient than a classical communication paradigm. Efficiency can be measured via the semantic impact (which is a new metric that we propose in Section~\ref{section_metrics}). In essence, efficiency is a measure of the time and communication resources that can be gained from adopting a semantic communication system, in comparison to a classical one. For example, assuming that the goal is to send large \ac{XR} content to the apprentice (contributing to better \ac{QoE}), if learning the structure of the data does not ultimately lead to a faster and more efficient generation of \ac{XR} content at the apprentice; classical communication is a better option.
    \item \textbf{Generalizable:} A specific semantic representation needs to be distribution, domain, and context invariant \cite{zhao2019learning, nguyen2021domain}. This invariance evokes generalizing to \emph{new and unseen, out of distribution, domain, and context} data points. In other words, the teacher must be able to extract particular features within the data whose structure is invariant to the context or domain. Context generalization can be seen in our example in Fig.~\ref{fig:semantic_convo}, here, the teacher needs to be able to semantically represent Max regardless of the context Max occurs in, and the correlations that such context might have with the content element. Meanwhile, domain generalization allows the reasoning node to learn a representation and generalize it to unseen target domains. For instance, assuming a semantic representation has been learned from an \ac{IoT} data source within image data, the reasoning node must be able to use this representation to describe the same semantic content element if it appears in the metaverse. 
\end{itemize}
Thus far, we have elucidated the foundational characteristics of a natural language and its contrasting features to a semantic-centric language. In addition to natural languages, there exists various forms of representations that are currently being considered to serve as a candidate semantic representation. In what follows, we overview the potential caveats of such representations.
\subsection{On Existing Forms of Semantic Representation}\label{existing_frameworks}
\renewcommand{\arraystretch}{1.0}
\begin{table*}[t]
	\footnotesize
	\caption{Characteristics contrasting potential approaches to reason semantic representations}
	\centering	
	\begin{tabular}{ m{2.8cm}  m{2.8cm}  m{2.8cm}  m{2.8cm}  m{2.8cm} m{2.8cm}}
		\hline
	\centering	\textbf{Key metric} & \textbf{NLP} & \textbf{ANN} &\textbf{Knowledge Graphs} &\textbf{Topos} & \textbf{Causal Representation Learning}\\
		\hline
		\textbf{Methodology Fundamentals}  & Read, understand, and decode human languages in a valuable fashion. Subsequently, use a human language to describe the semantic information contained in raw data.
& Perform semantic feature extraction, and then leverage a specific neural network structure to model the task complexity and its subsequent semantic features.  & Perform causal discovery of semantic features and represent such features and their corresponding relationships via vertices and edges in a knowledge graph. & Translate current data structures to a novel well-defined morphism, that enables extracting unobserved semantic information. & Learn a representation that can partially expose the unknown causal structure of data. This structure unveils the semantic content elements of the data and their relations, thus characterizing the context.\\
		\hline
		\textbf{Generalizability}  & Medium & Medium & Low & High & High \\ 
		\hline
		\textbf{Minimalism}  & Medium & Low & Medium & Medium & High \\ 
		\hline
		\textbf{Benefits}  & \begin{itemize}
			\item Easily understandable and decodable for design purposes.
			\item Suitable for specific data structures that heavily rely on text data.
		\end{itemize}& \begin{itemize}\item Easily integrated with existing ML and artificial neural network models of the \ac{E2E} network models.
		\item Has gained maturity and can be easily reparametrized. \end{itemize}& \begin{itemize}\item Characterizes intertwined parameters in simplified graphs. \item Provides the apprentice with a simplified basis for reasoning.  \end{itemize}& \begin{itemize} \item Capable of unraveling unobserved contextual patterns \item Can perform reasoning beyond statistical boundaries. \end{itemize} & \begin{itemize} \item Can leverage the concept of interventions and counterfactuals to understand the structure of the data beyond associative logic. \item The causes of semantic content elements implicitly characterize the context of the transmission.\end{itemize}\\ 
		\hline
		\textbf{Disadvantages}  &Limited by syntax, pragmatics, and wording. & Limited by statistical relationships within the data. & Can only represent simplified causal graphs, and restricted to the expressivity of a graph.  & Many concepts within Topos remain intractable and difficult to characterize.  & Limited by posing the proper interventions/counterfactuals at the apprentice. \\ 
		\hline
		\textbf{Major Challenge} & Transforms the bit-pipeline problem to a word-pipeline problem. & Limited reasoning capabilities with respect to contextual information that are not limited to statistical relationships. & Fails to characterize highly complex tasks despite its causal structure. & Non-scalable for the overall \ac{E2E} communication system. &  How to concatenate a structural causal model within an \ac{ANN} that characterizes both statistical and causal properties?\\
		\hline
	\end{tabular}
	\label{table:repmethods}
	\vspace{-0.25cm}
\end{table*}
Many efforts in the research community attempted to devise methods or mechanisms that yield a semantic representation of data. We have comprehensively detailed their contrasting characteristics in Table~\ref{table:repmethods}, and we explain them next.
\subsubsection{\Ac{NLP}}
The goal of \ac{NLP} is to transcend the capability of a computer and grant them the capacity to understand and speak human language. Recently, a number of works in \cite{xie2021deep, xie2022task}, and \cite{zhou2021semantic} have adopted \acp{NLP} in designing semantic communication systems. However, such works limited their communication to \emph{text data}, this is why a natural language was sufficient for \emph{describing} the data. In essence, describing complex messages like a hologram, or a high-precision manufacturing command using natural languages leads to more repetitive and redundant communication resources to describe the message. That is, our natural languages fail to ``describe" highly complex processes. For example if one tries to use the English language to describe the hologram of a particular entity, it would require a lot of time and sentences, that might not be able to ultimately generate the hologram in the intended fidelity. Meanwhile, a coded language has the capability to do so. Here, the semantic representation would be describing to the apprentice a mechanism similar to that code, which would automate the generation hologram via computing resources on the long-term. As outlined in the previous section, there are fundamental differences between a natural language and a semantic language. \acp{NLP} are limited to \emph{wording} in the same way that classical communication are limited to bit-pipeline. Additionally, \acp{NLP} require syntax and a set of deterministic rules, which are intrinsically not concerned with the meaning of the data. Hence, while \acp{NLP} are a good \ac{AI} tool for the \emph{service intelligence} of some tools like an automated chat bot or robot, if applied as is, they will fail at the helm of reasoning for low-layer data. 
\subsubsection{On \acp{ANN}}
\acp{ANN} mathematically exploit the universal function approximation theorem \cite{lu2020universal} to improve their generalization capability. Recent works in \cite{yang2021semantic, weng2021semantic, xie2021task, lu2022rethinking, dai2022nonlinear, wang2022performance} considered \acp{ANN} as a tool to design a semantic encoder and decoder. While \acp{ANN} are a powerful tool for data analytics, they are limited to the statistical structure of the data. Indeed, \acp{ANN} are unable to reason the cause, context, or effect surrounding the event or source of the data. Thus, the reasoning capability of \acp{ANN} is more or less limited by the statistical nature of the data $-$ \emph{they are not knowledge-driven}. In other words, if the data is not \emph{purely statistical}, an \ac{ANN} will not be able to capture the complexity of the data nor learn a proper representation. That said, \acp{ANN} remain powerful building blocks that are needed in our \ac{E2E} semantic communication framework. In essence, \acp{ANN} can potentially be used in modeling variables and parameters that describe purely statistical models, such as the channel, joint channel and source encoding processes, etc. While such processes will play a meaningful role in the ultimate semantic representation of the data, \acp{ANN} cannot be the main block in charge of yielding the semantic representation.
\subsubsection{Knowledge Graphs}
Several recent works \cite{zhou2022cognitive, jiang2022reliable, wang2021performance} developed semantic communication systems based on knowledge graphs. The approaches in \cite{zhou2022cognitive, jiang2022reliable, wang2021performance} adopt knowledge graphs as a result of their explainability and interpretability. However, in a realistic scenario, raw data is not structured into separate and well-defined categories and units. Thus, devising a semantic representation is a task that requires disentangling the raw data from memorizable information and scrutinize the structure and variability prior to any causal discovery. Moreover, as previously explained, the learnable data $\boldsymbol{X}_l$ must exhibit rich semantic structure. However, limiting the representation model to knowledge graphs limits the expressivity of the model, and, consequently, the capability of communicating semantically complex messages.
\subsubsection{Topos}
The works in \cite{babonnaud2019topos} and \cite{belfiore2021topos} used the mathematical framework of topos theory for semantic representation. Toposes originated from a purely mathematical foundation, in particular, homological algebra and algebraic topology \cite{heng2021mathematics}. In essence, this technique translates every data structure by a family of objects in a well-defined topos \cite{lawvere2009conceptual}. Here, a semantic representation stems on from the construction of a category set, a class of objects, types and stacks needs to constructed from the existing raw data. Nonetheless, adopting topos leads to multiple challenges: a) Deploying them requires a major overhaul on all existing machine learning frameworks running at any layer in the network stack. That is, they need to re-characterized and re-defined in the family of objects in the topos. Thus, topos cannot easily be integrated with co-existing \ac{AI} frameworks in the network stack, and b) Topos may have high computational complexity and intractability when handling raw data at the teacher's side prior to assigning it with a particular semantic representation. That said, despite these challenges, topos can play a role specific  building blocks of the \ac{E2E} network, as they are capable of characterizing a ``many-world view" interpretation in certain settings. For example, one can transcend basic probabilistic quantities in information-theoretic settings or semantic-aware ones to a cohomology class (or a cocycle) in toposses \cite{heng2021mathematics}. \\
\indent Having concretely defined the tenets of a semantic language, and explained the existing frameworks for semantic representations, we can now move our focus to the reasoning aspect, which is one of the most fundamental building blocks of a semantic communication network. Hence, in the next section, we first dwell on the concept of reasoning via causality. Then, we investigate the characteristics of causal representation learning that can lead to a minimal, generalizable, and efficient semantic representation and language. 
\section{Reasoning via Causality for Semantic Communication Systems}\label{reasoning_via_causality}
\begin{figure*}
    \centering
    \includegraphics[width=0.80\textwidth]{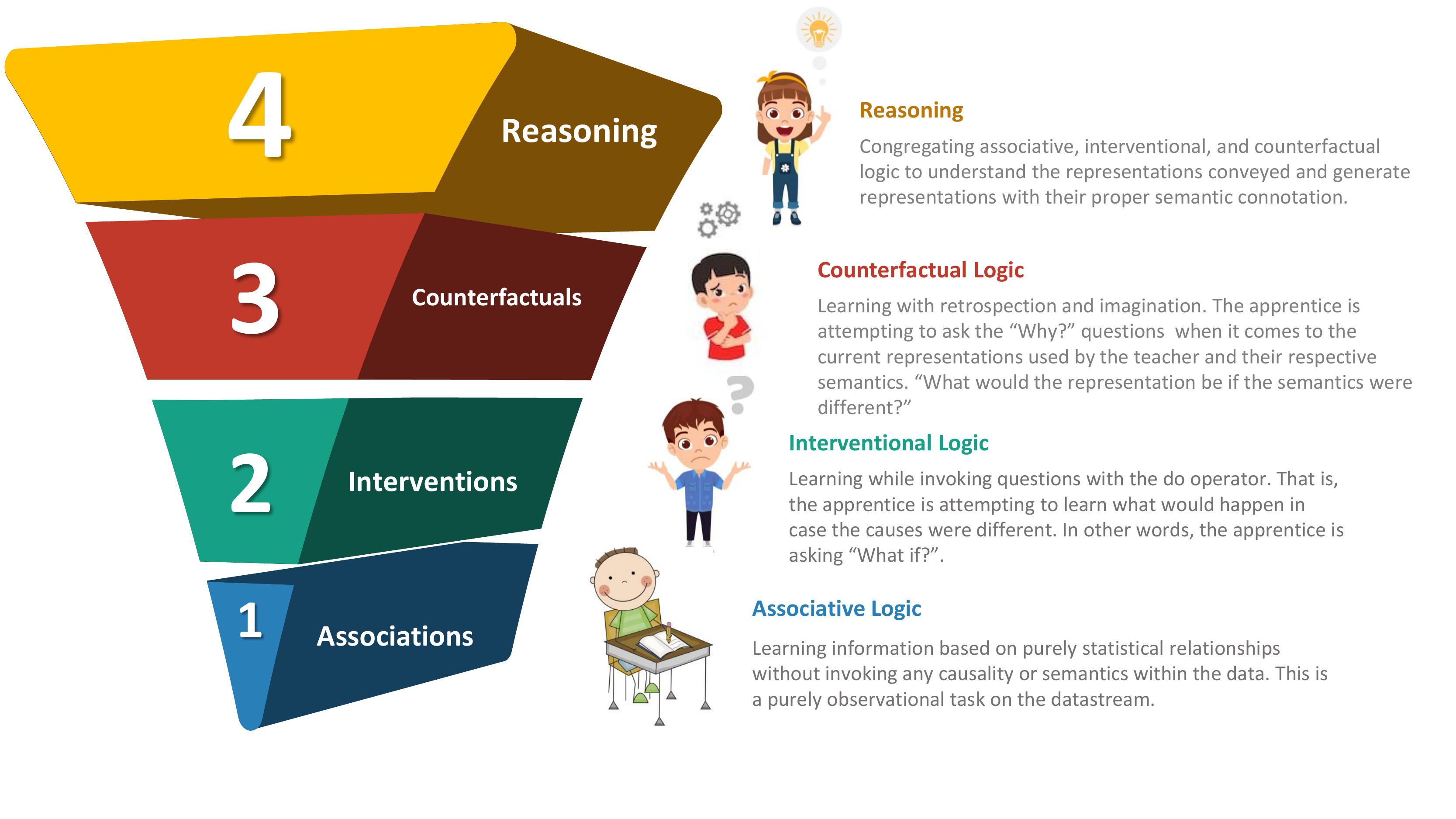}
    \caption{\small{Hierarchical Levels of Causality}}f
    \label{fig:causal_logic}
\end{figure*}
\subsection{Motivation and Preliminaries}\label{motivation_causality}
As shown in Table~\ref{table:repmethods}, existing generative methods have their own benefits and challenges in contributing to a proper semantic representation. In particular, one of the most fundamental challenges and drawbacks surrounding most of the current methods is that they excel at finding associations in \ac{i.i.d} settings. However, real-world data that is communicated between two wireless nodes is often not \ac{i.i.d}. For example, this confusion might take place when Max's representation in Fig.~\ref{fig:semantic_convo} is placed in a datastream that may be negatively correlated with its presence, i.e., background content that may negatively correlate with Max's data structure~\cite{scholkopf2022statistical}. In this case, a) the teacher will experience a difficulty in extracting the semantic content elements and b) the apprentice will face a difficulty in understanding the representation communicated by the teacher. Thus, adopting the previously surveyed \ac{AI} techniques (see Table~\ref{table:repmethods}) to extract or generate a semantic representation faces multiple challenges: \begin{enumerate}
    \item Existing frameworks (surveyed in Section~\ref{existing_frameworks}) cannot adapt to distribution shifts, i.e., the ability to generalize from one data point to the next, when sampled from a particular distribution that exhibits shifts. This is because such frameworks are mostly training and/or bias dependent. Thus, such frameworks exhibit \emph{weak vertical generalizability}.
    \item The aforementioned techniques do not possess the reasoning capabilities needed to draw logical conclusions out of datastreams, that go beyond generalizing to out of domain/distribution/context. The ability of the teacher/apprentice to perform associative logic and draw logical conclusions is a unique and necessary feature of a semantic communication system, that cannot be achieved by the previously discussed techniques.
    \item The various frameworks investigated in in Section~\ref{existing_frameworks} cannot adapt to novel domain and context settings. As a result of this caveat, such frameworks exhibit \emph{weak horizontal generalizability}. This is usually exhibited when the teacher and apprentice are communicating in a totally new setting, yet this setting exhibits new representations that have share analogous data patterns to previously exchanged and learned ones.
    \item Existing \ac{AI} frameworks generally have a large computational complexity. This drawback is more pronounced for tools such as knowledge graphs and Bayesian machines. Consequently, adopting such frameworks to build future semantic communication networks may not be scalable. 
\end{enumerate} 
\indent To remedy these technical shortcomings, it is necessary to move from \ac{AI}-augmented systems that are information-driven, to knowledge-driven or reasoning-driven systems as shown in Fig.~\ref{fig:datatoknowledge}. In particular, examining the \emph{``why and how?"} of the semantic content elements in an observed datastream or event, enables distinguishing between the set of meaningful information and the statistical accidents (e.g. identifying ``Type I error"). Essentially, a \emph{causal inference} framework consists of two parts: causal identification or discovery and statistical inference. In a semantic communication setting, \emph{causal identification} is the process of identifying the root causes of semantic content elements in the data. Furthermore, \emph{statistical inference} (in a causal-semantic setting) is the process of attributing a minimal semantic representation to that causal structural model. These two processes together create a semantic communication system with causal logic. Moreover, to efficiently perform \emph{causal identification} the following must be considered:
\begin{enumerate}
    \item If a radio node cannot characterize a correlation between the datastream to be conveyed and the established knowledge base (created through an accumulation of previous causal information), then the datastream does not exhibit any causality. Thus, this datastream contains pure random information, and no structure, and, thus, it is more efficiently transmitted classically. 
    \item If a correlation exists, between the datstream and the knowledge base, and there is a unique causal model that can rationalize this correlation, then true causality exists and it can be characterized via a representation.
\end{enumerate}
After evaluating the previously established correlation, the goal of \emph{statistical inference} is to asses the overall correlation and causality, as well as the corresponding stochastic changes in the system stemming from this causality. Furthermore, analyzing the reasoning behind the data pattern requires investigating the three levels of causal hierarchy: a) associations, b) interventions, and c) counterfactuals \cite{scholkopf2021toward} (which will be explicitly defined later in Section~\ref{fundamentals_causal}). Notably, the concept of ladder causation was first established in \cite{pearl2009causality} and is also called the Pearl's Causal Hierarchy. These conceptual levels constitute the foundational pillars of causal reasoning that we propose to use for equipping the teacher and the apprentice with a rationale beyond the statistical association between variables as shown in Fig.~\ref{fig:causal_logic}. Mathematically, these are logical queries that are posed via conditionals (for associational logic) or so-called ``do operators" (for causal logic). The ``do operator" is a mathematical operator that is performed on a causal model and its corresponding causal network which describe the intertwined cause-and-effect relationships in a particular datastream. It involves two tasks, a) setting a variable to a specific value (e.g. setting the parent node of branch within the causal network, to a certain value), and b) removing a particular branch in the causal network that contribute to a specific event. In a semantic communication setting, we can re-engineer the control plane and leverage such queries (via their ``do operators") to replace simple signaling messages. As will be discussed in more detail in Section~\ref{fundamentals_causal}, such queries can enable radio nodes to construct the causal model gradually, acquire reasoning capabilities, and ultimately learn a language.  \\
\indent In fact, causal inference has a long history in a variety of disciplines such as statistics, econometrics, and epidemiology \cite{scholkopf2021toward}. Thus, various frameworks exist for studying causality. That said, existing causality frameworks cannot be applied in a plug and play fashion over the data found at the teacher and apprentice of a semantic communication system. Henceforth, it is necessary to scrutinize the key aspects of causal inference that enable building a solid causal representation learning framework for semantic communications. Next, we shed light on definitions and fundamental concepts from causal representation learning that can be leveraged to establish a strong semantic connectivity between the teacher and apprentice, and subsequently a strong semantic connectivity in an \ac{E2E} network.
\vspace{-0.25cm}
\subsection{Fundamentals of Causal Reasoning}\label{fundamentals_causal}
Formalizing the concepts of causal representation learning in a semantic communication system requires introducing causality in the mechanism that builds and reasons over the semantic language and its components (semantic representations). One way to characterize such causality is via a \acrfull{SCM}. Essentially, mapping our language $\mathcal{L}$ into an \ac{SCM} creates many opportunities. In fact, \acp{SCM} enable three intrinsic causal concepts: a) graphical models, b) structural equations, and c) interventional and counterfactual logic. Henceforth, we map the semantic language between teacher and apprentice to an \ac{SCM} as follows:
\begin{definition}\label{def2}
Constructing a semantic language with causal reasoning capabilities requires mapping the language to an \ac{SCM} $\mathcal{L}:=(\boldsymbol{\psi}_L, p(\epsilon))$ where $\boldsymbol{\psi}_L=\{s_i\}_{i=1}^{N}$. The learnable data can now be written:
\begin{equation}
    X_{l,i}:=s_i(\epsilon_i, \rho_i).
\end{equation}
Here, $\rho_i$ is the set of direct causes leading to the specific data patterns in $\boldsymbol{X}_l$, $N$ is the number of semantic content elements contained in a datastream, and $p(\epsilon)$ is the joint distribution $p(\epsilon)=\prod_{i=1}^N p(\epsilon_i)$  over mutually independent exogenous noisy variables. Such exogenous variables map to the variability of the data previously defined in Section~\ref{near-optimal sem}. In a causal setting \cite{kaddour2022causal}, such ``noise" is thought to be an unaccounted source of variation. 
\end{definition}
Defining a language based on Definition~\ref{def2} enables us to leverage queries (interventions and counterfactuals). These queries (see Fig.~\ref{fig:semantic_convo} for illustrative examples queries) enable the apprentice to understand the reason a teacher used a particular representation (to describe a specific semantic content). In principle, as seen in Fig.~\ref{fig:causal_logic}, among queries, interventions on the causality ladder rank higher than associations. We can see that, under this definition, the language is now directly related to the structural assignments rather than being concerned with mapping the datastream to its corresponding task on an associational and statistical level. Furthermore, relying on interventions enables constructing a representation that does not merely rely on the \emph{observed datastream}. That is, the intervention proactively requires modifying the cause of events leading to a particular datastream, which enables inferring the ultimate technique to generate semantic content. If we look back at our intuitive example in  Fig.~\ref{fig:semantic_convo}, we can see that acquiring understanding requires the apprentice to intervene and ask questions to ultimately build the representation. Strictly speaking, in a causal setting, an \emph{intervention} represents a subset of the queries that the apprentice can ask to build their understanding, as defined next:
\begin{definition}\label{def3}
An intervention contributing to understanding, decoding, and eventually re-generating a representation communicated by the teacher is a question posed in the form of a ``do operator". In other words, given a model mapped to a representation $p(\boldsymbol{Z}|\boldsymbol{X}_l)$, the apprentice ought to ask questions that enable calculating and characterizing the quantity $p(\boldsymbol{Z}| do(\boldsymbol{X}_l=\boldsymbol{x}), A)$. Here $A$, is a latent variable that might be affecting the outcome of the representation. The ``do operator" enables the apprentice to understand the changes induced on the overall representation in case the datastream is different.
\end{definition}
It is important to note when intervening on a distribution with the \emph{``do operator"}, we are not considering a certain sub-population for which we observe $\boldsymbol{X}_l=\boldsymbol{x}$, but rather we are reasoning over the changes occurring on the datastream to be conveyed after taking an action on $\boldsymbol{X}_l$, namely $do(\boldsymbol{X}_l)$. Moreover, these interventions (and the counterfactuals, which will be discussed next) can:
\begin{enumerate}
    \item Replace control and signaling datastreams classically transmitted by radio nodes. This is particularly the case when the intervention is simple.
    \item Be transmitted via classical or semantic channels, that is, they can be: a) transmitted classically by the receiver (if they lack all reasoning foundations); or b) transmitted using a semantic representation. This can asymptotically lead to reverse mentorship whereby the apprentice is teaching the teacher their semantic language and vocabulary.
\end{enumerate}
\indent Furthermore, an \ac{SCM} model also enables the teacher-apprentice pair to leverage the concept of counterfactuals in a semantic communication system. Counterfactuals are at the highest level of hierarchy in causality to ultimately reach reasoning as shown in Fig.~\ref{fig:causal_logic}. That is, counterfactuals enable the apprentice to ask questions that include ``why" (not only ``what if") to understand the representation, and build their knowledge base. Here, formalizing counterfactuals goes beyond the \emph{do operator}. It incorporates factual data and an intervention (in which parts of the environment remain unchanged).   
To characterize the language via counterfactuals, one can modify its corresponding model as follows \cite{kaddour2022causal}:
\begin{definition}\label{def4}
Counterfactual questions can be reflected in a language $\mathcal{L}$ by replacing the prior distribution of variability $p(\epsilon)$ with the posterior $p(\epsilon|\boldsymbol{x})$:
\begin{equation}
    \mathcal{L}_x:=(\boldsymbol{\psi}_L, p(\epsilon|\boldsymbol{x})).
\end{equation}
\end{definition}
\indent In this case, mapping a language to an \ac{SCM} of this form, enables the apprentice to ``interrogate" their teacher with ``why"s and ``what if"s. The answers to these questions equip the radio node with a better understanding and knowledge base, enhancing their logic to ultimately tend to the human's brain. This definition further showcases the benefit of building a language according to an \acp{SCM}. \\
\indent So far, we have mapped a semantic language to an \ac{SCM} and we have proposed a suite of queries (interventions or counterfactuals) that can operate in the control or data plane to build a common language and understanding between teacher and apprentice. Another key benefit of mapping a language to an \ac{SCM} model is \emph{the disentanglement property}. That is, the semantic representation and its respective content element can be \emph{easily} separated from each other (the same way each word captures a standalone meaning in a natural language). Next, we highlight the disentanglement property and its respective consequences:
\begin{definition}\label{def5}
Building a semantic language $\mathcal{L}$ that can be mapped to an \ac{SCM} model enables disentangling each datastream and its respective representation from other established representations. In other words, the model describing the language can be written as \cite{kaddour2022causal}:
\begin{equation}
P(\boldsymbol{X}_l)=P(X_{l,1},\dots, X_{l,N})=\prod_{i=1}^M P(X_{l,i}|\rho_i),
\end{equation}
where $M \leq N$. 
\end{definition}
Definition~\ref{def5} implies that, when causality is present in the model, one can separate the parent root causes from each other, and consequently disentangle the semantic content elements. Furthermore, from Definition~\ref{def5}, we make the following observations:
\begin{enumerate}
    \item Performing an intervention or a counterfactual on one mechanism $P(X_{l,i}|\rho_i)$ does not change any of the other mechanisms $P(X_{l,j}|\rho_j)$, where $(i\neq j)$. This is inherently important because it creates a foundation that eases disentangling one learned task from another within a datastream at the teacher. For instance, in Fig.~\ref{fig:semantic_convo}, by performing queries on Max via mathematical operators (interventions or counterfactuals), can separate Max from the background or other transmitted information.
    \item Acquiring information about a specific mechanism $P(X_{l,i}|\rho_i)$, does not give us any information about $P(X_{l,j}|\rho_j)$. Intuitively speaking, information acquired about Max does not give further information about the sunflower in Fig.~\ref{fig:semantic_convo}. 
\end{enumerate}
\indent In light of this, the ``independent causal mechanisms" (independent causal mechanisms might be statistically correlated) principle~\cite{pearl2009causality} in Definition~\ref{def5} characterizes the dynamics of information shared between two distinct semantic representations (which represent two distinct semantic content elements). In essence, this is the principle that enables causal semantic representations to be invariant, autonomous, and independent as will be discussed in Section~\ref{invariance}. The importance of this principle can be highlighted by the following observation: if semantic representations were instead modeled via non-causal and purely statistical techniques \cite{scholkopf2021toward}, once the apprentice poses a new query on one semantic representation, the others will also be affected. In such cases, the factorization is known to be \emph{entangled}.\\
\indent As a result, in the converse, i.e., to evaluate the causality of a language, we first define disentanglement in what follows:
\begin{definition} For a set of representations $\boldsymbol{Z}$, s.t. $\boldsymbol{X}_l=g(\boldsymbol{Z})$ for some mapping $g$, a representation is known to be causally disentangled if the following factorization is possible \cite{kaddour2022causal}:
\begin{equation}\label{disentangled}
    p(Z_1,\dots, Z_M)= \prod_{i=1}^M p(Z_i|\rho(Z_i)),
\end{equation}
where $\rho(Z_i) \subset {Z_j}_{i\neq j} \cup \epsilon_i$ and $\epsilon_i$ is the exogenous causal factor of $Z_i$.
\end{definition}
From Definition~\ref{disentangled}, one can infer the following: Given a language $\mathcal{L}_t$, whose causality is not yet proved, if this language $\mathcal{L}_t$ admits multiple representations that can satisfy~\eqref{disentangled}, one can claim that this $\mathcal{L}_t$ is proven to be a \emph{a causal semantic language}. Furthermore, the presence of causality in the system opens the door for utilizing causal logic and queries like counterfactuals and interventions to perform reasoning and extract further information from the exchanged and generated tasks. Meanwhile, a non-causal model would only permit inferring information in a limited \ac{i.i.d} scenario. \\
\indent Furthermore, one can also \emph{leverage the principle of independent causal mechanisms to disentangle the structure and variability} of a specific datastream. In fact, one can adopt frameworks like contrastive learning \cite{chuang2020debiased }, which is a discriminative self-\ac{ML} framework that performs positive and negative sampling to ultimately create a semantic equivalence between the datatreams that need to yield the same semantic representation within a specific variability. It also establishes a distance between semantically different samples within a representation space. In fact, as shown in \cite{mitrovic2020representation}, performing contrastive learning on a causal model proved to be capable of invariantly learning representations. We will further elaborate the techniques and enablers of causal invariant representation learning next. 
\subsection{Causality for Generalizable Representation Learning}\label{invariance}
Thus far we have scrutinized the fundamentals of causal representation learning. In particular, we have identified its peculiar features that grant the teacher and apprentice the proper tools to reach the reasoning foundation needed to extract a minimal and efficient semantic representation. That said, given that raw data can result from heterogeneous sources, and can exhibit horizontal and vertical shifts, the representation must be \emph{invariant} to such changes. Such an invariance enables the teacher/apprentice pair to generalize the learned semantic language to \emph{new, unseen, and out of domain, distribution, or context} datasteams. Thus, this generalizability is characterized by the universality of the semantic language.
\vspace{-0.2cm}
\begin{definition}\label{gen}
A semantic representation is dubbed, generalizable, if it fulfills the general causal invariant prediction criterion. That is, despite different ``what if"s posed on the causal model, the same representation results in describing its respective content elements in data:
\begin{equation}
p^{do(\kappa_i)}(\boldsymbol{Y}|\boldsymbol{Z})=p^{do(\kappa_j)}(\boldsymbol{Y}|\boldsymbol{Z}) \forall  \kappa_i,  \kappa_j \in \mathcal{K},   
\end{equation}
where $\mathcal{K}$ is the set of queries at the apprentice which pose interventions on the considered \ac{SCM}.  
\end{definition}
 Definition~\ref{gen} is formulated based on the following observation: When considering a particular representation, and its corresponding content element, if a set of two different queries leads to the \emph{exact same learned causal model}, then such a representation is \emph{generalizable}. That is, irrespective of the queries that the apprentice has asked, the semantic language used by the apprentice remains consistent. This means that the representation used by the teacher can be applied irrespective of the data, distribution, and context. One analogy that one can draw is to ``words" that we use in our daily lives: A \emph{word} (which mimics a representation in a semantic language) consistently describes the same meaning in any context, time, and space limits. Moreover, in light of this, to guarantee that the yielded representations from our causal model are \emph{invariant}, we further detail two approaches that leverage the invariance principle in the design of the causal reasoning faculty:
\begin{enumerate}[label=(\roman*)]
    \item \textbf{Contrastive Causal Learning}:\\
    Given that contrastive learning is a form of self-supervised learning, one can leverage the approach adopted in \cite{mitrovic2020representation} to identify the invariant structure properties of a semantic representation. In particular, one can train or bias the apprentice, prior to any information exchange to a neutral causal structure of the data, then the apprentice is causally taught to disentangle the structure and variability, which map to the content and style of an \ac{SCM}. Thus, the structure $\boldsymbol{\psi}_L$ and the variability $\varkappa_L$ are assumed to be independent of each other. For instance, if the structure denotes the semantic representation of a dog, that will not have a bearing on the breed of the dog. They complement each other yet remain independent. Building on this contrastive learning setting, invariance can be achieved by applying Definition~\ref{gen} to the specific disentangled setting herein focused on the structure rather than the representation:
    \begin{equation}
        p^{do(\varkappa_i)}(\boldsymbol{Z}|\psi)=p^{do(\varkappa_j)}(\boldsymbol{Z}|\psi), \forall \varkappa_{i,j}\in \mathcal{V}.
        \end{equation}
    That is, the knowledge base of a radio node has robustly acquired a particular structure $\psi$ irrespective of the variability in which such structure might appear in. In other words, assuming a particular representation $\boldsymbol{Z}$ describes a structure $\psi$ mapping to a dog, and a variability mapping to $\varkappa$ German shepherd, the radio node can robustly generalize the structure $\psi$ to any breeds of dogs, irrespective of where they appear (the dog example is for illustrative purposes only and this can be expanded to any data type).
    \item \textbf{Counterfactual Invariance}:\\
    So far, we discussed how counterfactuals can enable the apprentice to gather higher reasoning capabilities. Moreover, counterfactuals enable leveraging the framework of \emph{counterfactual invariance} \cite{veitch2021counterfactual}. Adopting this framework can construct predictors that are invariant to particular perturbations in the raw data $\boldsymbol{X}_l$. Based on Fig.~\ref{fig:causal_logic}, such form of invariance must be stronger than the one imposed via interventions. This is important for semantic communications because for this causal model, despite different ``why"s posed on the causal model (in contrast to the weaker ``what if"s), the semantic language remains consistent. Essentially, this framework is built on the premise of identifying an additional variable, say $\Upsilon$, that captures information that must not influence the semantic representation $\boldsymbol{Z}$, nor its semantic content $\boldsymbol{Y}$. If we take the intuitive example in Fig.~\ref{fig:semantic_convo}, in the process of building a semantic representation for Max, the background or the position of Max, must not affect the semantic representation chosen for Max. As such, the background does not have a causal effect on the covariates of $\boldsymbol{X}_l$. More formally, the definition of counterfactual invariance is given as follows:
    \begin{definition}
    A semantic representation $\boldsymbol{Z}$ is counterfactually invariant to $Q$ if for any give sample $\boldsymbol{X}_l$, a counterfactual ${X}_{l,q}\sim p(\boldsymbol{x}_{l,q}|\boldsymbol{x}_l)$, and $\forall q \in \mathcal{Q}$, we have $z(x_{l,\upsilon})=z(x_{l})$ \cite{veitch2021counterfactual}.
    \end{definition}
    Adopting this logic at the apprentice requires us to \emph{first} identify the causal directions surrounding the raw datastream bits $\boldsymbol{X}_l$ and their corresponding semantic content $\boldsymbol{Y}$. \emph{Second}, the apprentice must be capable of capturing the attributes belonging to $Q$. Furthermore, the apprentice must be able to identify the associational relation between $Q$ and $\boldsymbol{Y}$ to reason whether this relationship is due to confoudning or selection bias \cite{kaddour2022causal}. That said, acquiring all of this knowledge at this apprentice requires either a knowledge map or a set of labeled data, i.e., the raw data and their respective semantic content (not the representation). Henceforth, it is worth exploring techniques that enable leveraging this concept further while freeing the apprentice from the aforementioned restrictions. 
\end{enumerate}
\subsection{Challenges and Future Directions}
In this section we first highlight the main challenges facing causal models. Then, we discuss the potential opportunities that can be leveraged to address such challenges and subsequently build semantic communication systems with robust reasoning faculties.
\begin{itemize}
    \item \textbf{Causality alone is not enough:} Causal models enable building a semantic language with intrinsic causality. Such causality enables the teacher/apprentice to acquire an understanding via counterfactuals and interventions. These queries and their corresponding answers equip radio nodes with robust knowledge bases. That said, implementing \acp{SCM} into a semantic communication system faces some challenges such as a: a) difficulty in initializing such \acp{SCM}, b) difficulty in expressing causal and statistical relationships within the data. Here, an incorrect initialization might lead to a \emph{biased} semantic language. Meanwhile, statistical relationships on top of causal ones expand the universality of a semantic language, and subsequently improve the generalizability of wireless networks. In essence, an \ac{SCM} needs to be embedded into a large \ac{AI} or \ac{ML} whose inputs and outputs may be unstructured or naturally entangled. This also further enables expressing the statistical relationships in the data that are not characterized via causality. Here, we envision that one avenue that can potentially mitigate the aforementioned challenges is to adopt an \ac{AI} system that merges connectionist \ac{AI} and symbolic \ac{AI}~\cite{smolensky1987connectionist}. One form of this integration is dubbed, \emph{neuro-symbolic \ac{AI}}. Neuro-symbolic \ac{AI} is an emerging concept that merges data-driven neural architectures which extract statistical structures from raw datastreams with symbolic AI representations of logic. In fact, in our recent work in \cite{thomas2022neuro}, we showed that a neuro-symbolic AI framework based on generative flow networks \cite{bengio2021flow} and logic-based symbolic components can be used to design an end-to-end semantic communication systems. Our results in \cite{thomas2022neuro} show that causality-based neuro-symbolic AI can indeed help achieve minimal representations, create symmetric communication channels, enable generalizability, and reduce the amount of data transmitted. Naturally, this early work can further be extended to accommodate several of the key concepts that we presented here, including the use of more elaborate \ac{SCM} models, fully exploiting the causality laddder in Fig.~\ref{fig:causal_logic}, designing novel neuro-symbolic architectures that go beyond generative flow networks, and the use of our newly defined metrics in Section~\ref{section_metrics}.
    \item \textbf{Highly complex causal models are problematic:} Expressing all the complexity of a task via an \ac{SCM} leads to a challenge in solving the optimization problem in \eqref{complexity}. That is, if the data is very complex, the reasoning radio node might not be able to find a tradeoff between the structure and complexity of the data. Also, this will jeopardize the explainability of the established knowledge base and the expressed semantic language. One key aspect to investigate here would be extending \acp{SCM} via novel \ac{AI} techniques that can represent highly complex causal models while maintaining their expressivity.
    \item \textbf{How to pose the proper interventions and counterfactuals via queries?} Theoretically speaking, and given a particular datastream that exhibits structural characteristics, the apprentice needs to pose the \emph{right} interventions and counterfactuals. In other words, the apprentice needs to have \emph{minimal reasoning capabilities} that enable them to pose the right questions. On the one hand, if the apprentice poses questions that are irrelevant of the context, the learned structural model will be wrong. On the other hand, if the apprentice has been accustomed to a particular context, and is therefore \emph{biased}, they might ask questions that only enable learning selective semantic content elements of the data. Here, one aspect that is worth exploring is to send the apprentice information that enable them to build proper context and subsequently pose the proper queries. For instance, given that the teacher uses raw data to complement their semantic representations, the teacher can build on this raw data to guide the apprentice vis-à-vis context and the queries they must ask to understand the language. Here, one can adopt the game-theoretic scheme proposed in \cite{thomas2022neuro} to initialize the gradual design of a semantic language. Clearly, this is a nascent research direction that needs to be investigated more closely.
\end{itemize}
Thus far we have scrutinized the fundamental reasoning concepts needed to design a semantic communication system. Next, we will investigate novel metrics that enable a proper evaluation of future semantic communication systems.
\section{Semantic-Enabled Communication Metrics}\label{section_metrics}
Based on the reasoning foundations we have built so far, in this section, we will establish a set of new semantic communication metrics that enable evaluating the performance of next-generation semantic communication systems.
\subsection{Index of Communication Symmetry}
In classical communications, the setting of communication was governed by asymmetry. This asymmetry had its bearing weight at the transmitter side, whereby transmitting nodes had the utmost power in terms of knowledge with respect to the datastream. That is, the transmitter is either generating or observing the datastream. Meanwhile, the receiver's main goal was to only \emph{identify}, at the destination, the message produced at the source. Thus, the receiver did not have an \emph{active} role given that it could not \emph{generate} anything from minimal information. This phenomenon is in alignment with the data processing inequality that explicitly states the fact that ``information" (per Shannon's definition) cannot be created in an ex nihilo fashion. However, in a semantic communication system, the overall situation changes given that the apprentice can leverage reasoning and causality to \emph{generate} the originally sent message. In fact, asymptotically, one can think there could be a channel between two points, even if there is no direct physical (wired or wireless) link joining them. In other words, in semantic communication networks, the concept of communications is \emph{governed by manipulation}. In fact, the authors in \cite{lopez2019no} claim that, \emph{``there is no communication without manipulation"}. Thus, whatever the teacher manipulates at the transmission end, must be re-generated within its semantic context identically at the apprentice.\\
\indent Consequently, it is necessary to introduce a suite of novel metrics that characterize the level of symmetry between a teacher and an apprentice. Such a metric must:
\begin{enumerate}
    \item Understand the reasoning capability of the teacher and the apprentice.
    \item Investigate whether an equilibrium has been reached, whereby the teacher and apprentice can majorly rely on semantic-based transmissions.
    \item Scrutinize whether a reverse-mentorship situation is taking place, whereby the apprentice's reasoning capabilities are superior to the teacher's.
\end{enumerate}
Prior to proposing such a metric, we first need to define a novel concept, called \emph{semantic impact}. This stems from the fact that the significance of a particular semantic representation $Z_i$ can be characterized by the equivalent number of data packets one would have needed to convey the exact same message. Considering a particular semantic representation, $Z_i$, and its respective semantic content  element ${Y}_i$, one must ask: ``If we transmitted the data classically, during a time duration $\tau$, how many data packets would be needed to generate the same semantic content?". We answer this question via a novel metric, dubbed, semantic impact:
\begin{definition}\label{sem_impact}
The semantic impact $\iota_\tau$ generated by a semantic representation $Z_i$ during a time $\tau$ is the number of packets that would have been needed to be transmitted to regenerate the semantic content element $Y_i$.
\end{definition}
\begin{proposition}\label{com_symmetry}
The communication symmetry index between a teacher $b$ and apprentice $d$, for a transmission session $\tau$ is given by:
\begin{equation}
\eta_{b,d,\tau}= \frac{\zeta_{d,\tau}}{\nu_{b,\tau}}\times \iota_{\tau,Y_i},    
\end{equation}
where $\iota_\tau$ is the semantic impact, $\zeta_{d,\tau}$ is the number of query packets (interventions, counterfactuals, etc.) requested by the apprentice to reason over the message transmitted, and $\nu_{b,\tau}$ is the number of raw data packets transmitted by the teacher to accompany the semantic representation sent. We also note that, if the queries are communicated via a semantic representation to the teacher, one can find $\zeta_{d,\tau}$ by applying the concept of semantic impact on the representation used.
\end{proposition}
Thus, this communication symmetry index along with the semantic impact enable us to characterize the reasoning state of the teacher and apprentice and the equilibrium reached by the teacher and the apprentice:
\begin{itemize}
    \item If $\eta_{b,d,\tau} \le 1$ and $\iota_\tau>1$: This is a setting in which the apprentice has little to no knowledge base. Here, the apprentice has not acquired the reasoning capacity to intervene and interrogate the teacher on the semantic representation sent. This setting asymptotically mimics the pure classical communication scenario, whereby the majority of the data is still being sent in its raw form to complement the semantic representation. 
    \item If $\eta_{b,d,\tau} \to \iota_\tau$ and $\iota_\tau>1$: This is a setting in which the apprentice has considerable knowledge and reasoning faculties. The teacher is still complementing their conveyed information with raw data, however the apprentice is intervening often to understand the causal structure of the data and gradually rely on semantic representations.
    \item If $\eta_{b,d,\tau} = \iota_\tau$ and $\iota_\tau>1$: In this setting, the apprentice is intervening the same way current receivers transmit an acknowledgment, meanwhile the teacher is only relying on raw data transmissions to characterize the unlearnable part of the data (that is solely governed by randomness).
    \item If $\eta_{b,d,\tau} > \iota_\tau$ and $\iota_\tau>1$: This is a very peculiar setting, whereby the datastream is mostly learnable and not memorizable, thus the teacher is relying mostly on semantic representations. Also, the apprentice is actively intervening to generate the conveyed message from the set of semantic representations conveyed.
    \item If $\eta_{b,d,\tau} > \iota_\tau$and $\iota_\tau\leq1$: Here, the number of queries requested by the apprentice is larger than raw data transmissions by the teacher. This settings represents a challenging scenario in which the teacher is unable to extract a proper semantic representation to be communicated with the apprentice.
\end{itemize}
We can see that parameter $\eta_{b,d,\tau}$ does not go significantly below $1$ unless a defect in reasoning is observed. Next, we build on the concept of communication symmetry index to propose a novel \emph{reasoning capacity} metric.
\subsection{From Information Capacity to Reasoning Capacity}
In classical communications, most of the network metrics are built on the concept of information capacity. In essence, information capacity characterizes the maximum achievable data rate of a particular communication link, under specific bandwidth allocation. This metric is fundamentally important because it enables the system designer to quantify the range of communication rates possible, and thus understand the type of applications or use cases that can benefit from such a \ac{QoS}. That said, it is important to note that this characterization is based on the classical concept of ``information", whereby information is merely quantifying uncertainty. In other words, the capacity will not describe the number of symbols that can be contained in a communication channel, but rather the number of choices that the ``\emph{information}" can take, i.e., the number of choices one could transmit in a particular communication channel. Shannon proved that the channel capacity is equal to the mutual information of the channel maximized over all possible choices at the transmitter:
\begin{equation}
C=\max _{P(X)} I(X, \Tilde{X}),
\end{equation}
where $\Tilde{X}$ is the output of the classical communication channel.\\
This definition is built on the method Shannon utilized to quantify information via uncertainty (see Section~\ref{subsection_informationtheory_tosemantictheory} for a more elaborate discussion about information via uncertainty). Building on this definition, there exists a need to propose a novel ``capacity" metric that is capable of accounting to the novel nature of a semantic communication system. To do so, the following key aspects must be scrutinized prior to the design of a novel notion of \emph{reasoning capacity}:
\begin{enumerate}
    \item Information must be defined as a semantic substance rather than a mere uncertainty measure \cite{9844779}, whereby the quantification of ``information", characterizes the semantic representation communicated between the teacher and the apprentice.
    \item The concept of ``information transmission" must not be viewed as a mere propagation between two points of physical space \cite{lopez2019no}. Instead, it must be seen as a generative process at the apprentice that requires reasoning, and is controlled via manipulation.
    \item Reasoning capacity cannot be high in the presence of an asymmetry between the teacher and the apprentice. In a classical setting, the communication is highly manipulated by the transmitter and the receiver does not perform any tasks on the data beside mere decoding and reconstructing. Furthermore, in a semantic setting, having a very well-versed teacher but a non-trained apprentice cannot yield a high reasoning capacity, as the approach ends up leading to a scenario similar to the classical asymmetrical one. That said, in contrast to the classical approach, the apprentice is in the gradual reasoning stage as they are incrementally building on their knowledge. 
    \item Similarly, having a well-versed apprentice but a teacher with weak representational capability leads to the use of redundant computing and communication resources over the course of connectivity. Even though the main course of information transmission must flow from the teacher to the apprentice, the teacher must gradually acquire new tools from the apprentice. That is, the teacher will establish a subset of their semantic language based on the apprentice's previous experiences and knowledge base. 
    \item Henceforth, a high reasoning capacity necessitates a teacher with well-versed representational skills (one that can easily build a semantic language based on a particular datastream), and an apprentice with good reasoning capabilities (one that can understand a representation and use it to computationally generate semantic content). 
\end{enumerate}
Our novel, proposed reasoning capacity measure must, unlike the classical capacity measure, correlate with the message complexity proposed in Proposition~\ref{complexity}. That is, the classical capacity measures would not change as a function of the \emph{message complexity} but rather as a function of the \emph{channel conditions}. If the goal is to load a large \ac{XR} content, a high data would be needed as the information is recovered a the receiver side in a \emph{bit-by-bit} fashion. Instead, if the receiver becomes an apprentice that relies on learning the content rather than simply recovering it, the \emph{understanding} of the apprentice and its \emph{impact} on the reconstruction process would be the \ac{KPI} of a reliable communication link between teacher and apprentice.\\
Essentially, characterizing the understanding of the apprentice depends on:
\begin{enumerate}
    \item Reasoning as measured by the number of queries posed by the apprentice.
    \item Efficiency and minimalism as measured by the number of raw messages supplemented to the semantic representation, as well as the impact of the semantic representation.
\end{enumerate}
\begin{figure*}
    \centering
    \includegraphics[width=0.80\textwidth]{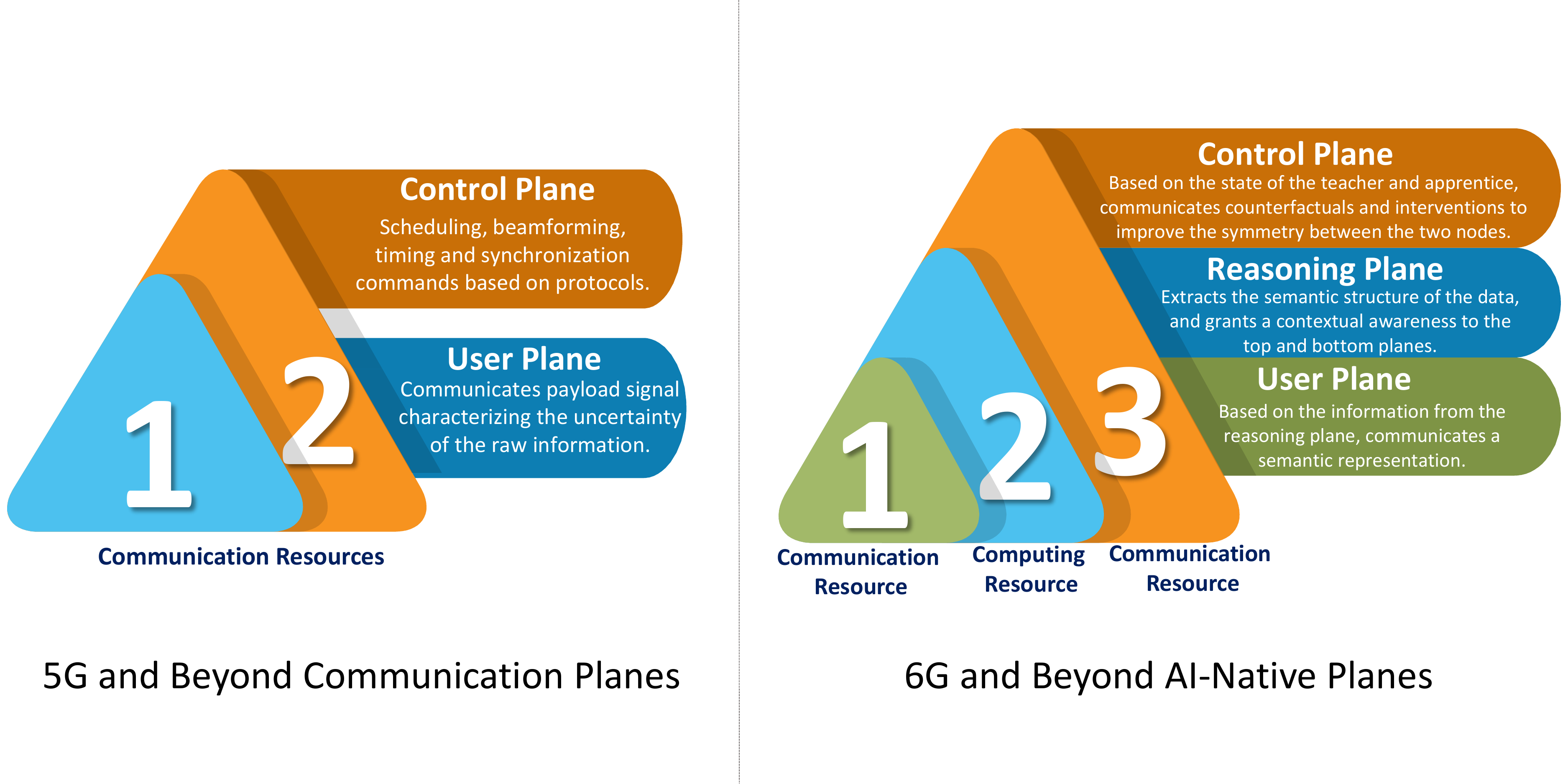}
    \caption{\small{An illustrative figure showcasing the transition from communication-only planes in 5G and beyond to AI-Native Planes in 6G and beyond.}}
    \label{fig:semantic_planes}
    \vspace{-0.25cm}
\end{figure*}
Consequently, we propose the following \ac{KPI} for semantic communication links:
\begin{proposition}\label{reasoning_capacity}
The reasoning capacity between a teacher $b$ and an apprentice $d$ is given by:
\begin{equation}
C_R= \Omega \log_2(1+ \eta_{b,d}), 
\end{equation}
where $\Omega$ is the maximum computing capability of the server used to represent or generate the semantic representation, and $\eta_{b,d}$ is the communication symmetry index per second.
\end{proposition}
Here, $\eta_{b,d}$ is computed per second rather than transmission session, thus $\tau$ is integrated out. Furthermore, the reasoning capacity is still a binary log function since the representation is still sent via bits over a channel. This proposition is universal in a sense that it is independent of the type of semantic representation utilized. It is also additive to the classical capacity metric given that $\iota_{\tau, Y_i}$ enables the classical conversion. It is thus important to find techniques that can accurately measure $\iota_{\tau, Y_i}$. Furthermore, given a datastream $X$, with $\boldsymbol{X}_l$ learnable information and $\boldsymbol{X}_m$ memorizable ones, one can simply use addition to compute the capacity. That is, the total achievable capacity would be given by:
\begin{equation}
C_T=C_C+C_R= W \log_2(1+ \gamma)+\Omega \log_2(1+ \eta_{b,d}),
\end{equation}
where $W$ is the bandwidth and $\gamma$ is the \ac{SINR}. One important thing to note here is that while the $C_C$ is limited by Shannon's bound, and the bandwidth of operation; $C_R$ is essentially bounded by the reasoning bound and the computing resources. Notably, claiming an \ac{E2E} capacity that might achieve an impact that is higher than what Shannon could have is possible because we are alternatively relying on computing. In other words, if the apprentice is capable of regenerating the message (via a semantic representation) faster than it could be classically transmitted, the \ac{E2E} capacity could asymptotically reach beyond Shannon's limit. \\
\indent Hence, we have so far elaborately investigated the principles of reasoning to elicit a semantic representation. Subsequently, we have proposed a set of novel semantic-based communication metrics to ensure that future systems are evaluated efficiently. Next, we will look into the considerations needed when scaling the semantic communication system from one teacher and apprentice, to an overall large scale network.
\section{Scaling Semantic Communications: From Semantic Links to Semantic Networks (6G and Beyond)}\label{scaling}
Thus far, we have discussed the concept of semantic communication systems for a link a teacher and an apprentice, i.e., two radio nodes. In this section, we further discuss how those can be used to build large scale semantic communication networks.
\subsection{Early deployments of semantic communications: challenges and facilitators}\label{co-evolution}
In this section, we will first discuss the techniques that enable developing the reasoning skills and the language with nascent and uninformed teachers. Then, we will shed light on the early use-cases and forms of semantic communication systems.
\subsubsection{Uninformed Teachers}
A semantic communication system essentially depends on the reasoning faculties of the teacher and apprentice. While the symmetry between the teacher/apprentice pair increases as the apprentice learns the semantic language, we have assumed thus far that the teacher has acquired the faculty to provide and teach a semantic language. However, in many instances, the teacher lacks the capability to develop their reasoning to ultimately teach the apprentice this language. This is the case when a radio node has never communicated via semantic-language before, or when this node is trying to communicate a novel structure that is considered ``novel vocabulary" with respect to its acquired knowledge. In this case, this ``uninformed" apprentice must use alternative techniques to develop their reasoning capability and learn the foundations of a language, given a structural datastream. This can be performed via:
\begin{itemize}
    \item \textbf{Reverse mentorship:} If the apprentice has powerful reasoning and representational capabilities, the apprentice can through, interventions and counterfactuals, teach the teacher specific data representations. The teacher may in fact ``meet'' such an informed apprentice if it moves in a network. 
    \item \textbf{Data showers:} The teacher can resort to standalone cloud services that offload standalone libraries that can complement their knowledge and ultimately build a basic semantic language.
\end{itemize}
\subsubsection{How to prepare today's networks so as to deploy semantic communications?}
Thus far, we have investigated how the concept of semantic communications can pave transform wireless networks from \ac{AI}-augmented, data driven networks, to \ac{AI}-native reasoning-driven networks. While this leap can be revolutionary, moving towards this direction will require:
\begin{itemize}
    \item \textbf{Readiness of computing resources:} In semantic communication systems, producing/understanding a language on the transmitter/receiver end requires a high abundance of computing resources at the end devices and \acp{UE}. In light of the \ac{AI}-wave for wireless networks, an exponential growth has been observed in the computing resources of modern devices, yet, this remains heterogeneous and disparate among devices. Henceforth, an open problem in this direction is to scrutinize: a) the level of reasoning that can be achieved with limited computing resources, and b) the evolution of a semantic language built based on this reasoning, and whether it can yield substantial benefits to the \ac{E2E} system. In addition, it would be important to investigate how a semantic communication network can be built out of devices that have very different reasoning/computing capabilities, and what the performance gains would be for semantic communication in such heterogeneous systems. 
    \item \textbf{Maturity of reasoning-intensive applications:} As a result of the characteristics of semantic communication systems, some applications might benefit more than others from using semantic communication mechanisms. For instance, services that require a high level of service intelligence (in contrast to operational intelligence which is used to optimize the network to fulfill the \ac{QoS} of the application), i.e., the execution of the service is complex and/or might have a high level of autonomy; can highly benefit from the semantic structure to produce content. For example, the production of holograms in holographic teleportation and high-precision manufacturing in Industry 5.0 are tasks that require multiple \ac{AI} mechanisms (this is independent of how the content will/can be transmitted). Here, using a semantic language to communicate the semantic structure modifies the content production process at the receiver. That is, instead of decoding bit-by-bit such complex messages, the receiver must reproduce (using their computing resources) the described task. Instead of transmitting the complex content in a message, the message would be comprised of a language that describes the key aspects of the complex content. This enables teaching the receiver to \emph{automate and produce the content based on their knowledge base}. 
    \item \textbf{Highly cooperative systems:} Semantic communication systems highly depends on the concept of a \emph{language}. Services that require continuous and real-time data or control messaging between homogeneous nodes can benefit from a language more than others. That is, a swarm of vehicles or robots interacting with an environment can highly benefit from communicating a language. This can minimize the number of control messages repetitively communicated to achieve a goal. Also, digital twins\cite{hashash2022} require a tight level of synchronization between the physical and the cyber twin. A semantic language here can improve the cooperation between the twins and guarantee achieving a real-time replica of twin in the metaverse.
    \item \textbf{Evolution of \ac{AI}:} On top of the evolution needed on the network's architecture, semantic communication requires a major evolution of current \ac{AI} frameworks. As of today, causal representation learning, as well as frameworks that deploy associational and causal logic remain in their infancy. Thus, a suite of novel \ac{AI} mechansims that can efficiently execute the causal ladder we previously described is needed to ensure the deployment of future semantic networks.
\end{itemize}
\indent Hence, based on our previous observations, we expect that the use case of \emph{goal-oriented communications} will be the first major application of semantic communications , which somewhat explains the high interest that it received from early-on semantic communication papers~\cite{farshbafan2022curriculum, zhang2022goal}. Here, services that could highly benefit from goal-oriented communications like cyber-physical systems, digital twins, and \ac{CRAS} could be the first adopters of semantic communications given the: a) The abundance of computing resources as a result of their infrastructure, and b) The high level of cooperation that intrinsically exists in the operation of such services. For instance, for digital twins, transforming the communication between the physical twin and the cyber twin to a semantic-based one enables them reaching a high level of synchronization. Here, this can be done by majorly relying on computing resources, and thus \emph{minimizing the need} for a highly reliable and low latency channel to achieve the overarching synchronization goal.\\
\indent Next, we investigate the opportunities and challenges of semantic communications in the large-scale cellular network context.
\subsection{\ac{E2E} Semantic Large-Scale Wireless Networks}
In this subsection, we further expand our study on semantic communication networks, by examining how their corresponding networking protocols could look like. Then, we investigate the techniques needed to distribute computing resources in a network, which showcases the true convergence of communication and communication resources in semantic systems. Finally, we scrutinize how semantic communication systems transform today's 5G \ac{O-RAN}.  
\begin{figure*}
    \centering
    \includegraphics[width=0.80\textwidth]{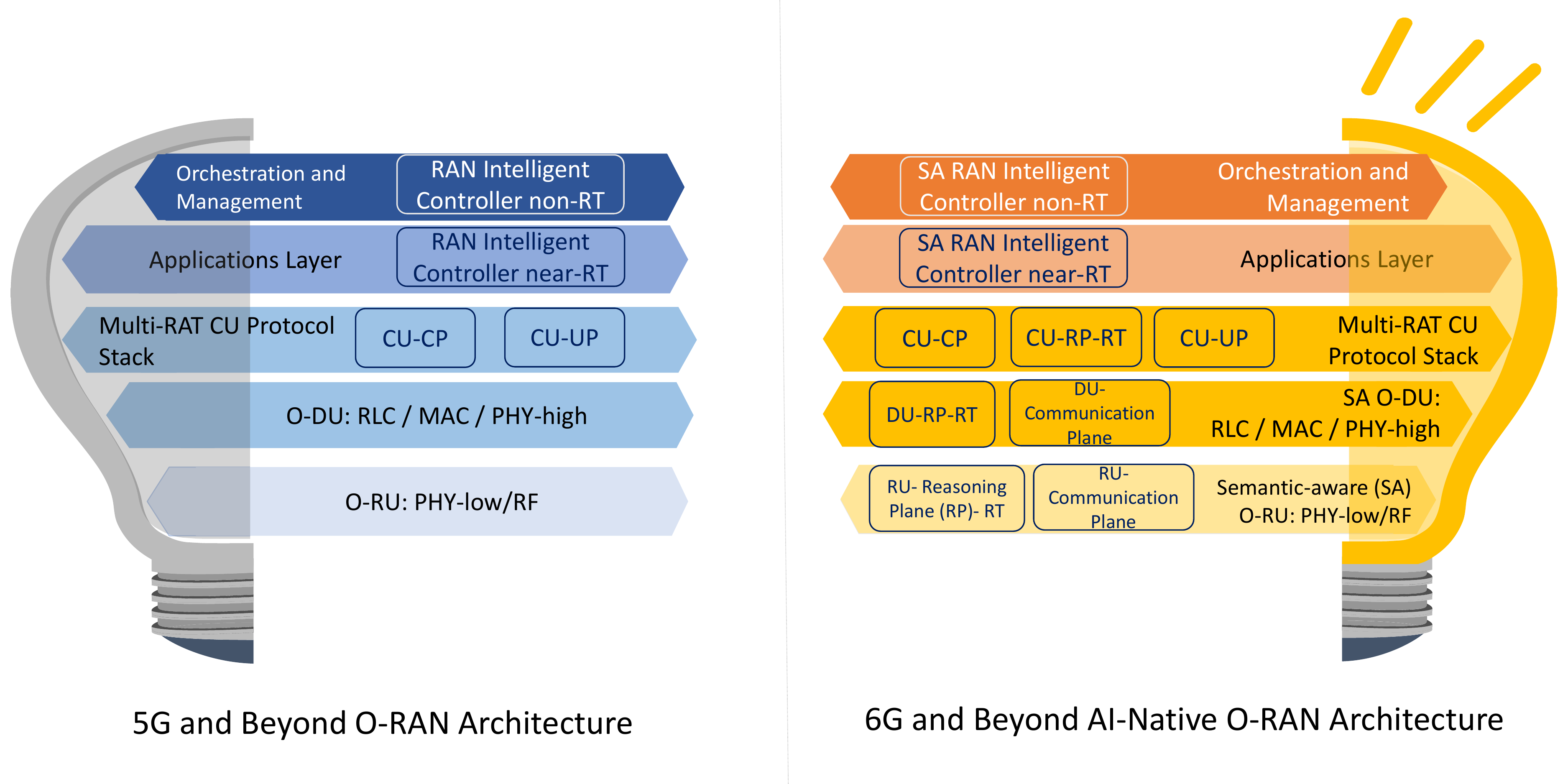}
    \caption{\small{An illustrative figure showcasing the transition of the current O-RAN architecture to a future AI-native O-RAN architecture.}}
    \label{fig:oran_architecture}
\end{figure*}
\subsubsection{Networking and Computing Considerations}
\indent \\$\bullet$ \textbf{On Semantic-Based Networking:}
In Section~\ref{reasoning_via_causality}, we investigated how interventions and counterfactuals enable the apprentice to learn about the structural causal model that the teacher is trying to convey. Mathematically, such queries are \acp{PDF} with ``do operators" that enable the apprentice to better understand the causal model. That said, the apprentice will be communicating such ``do operators"  in the uplink in contrast to classical control acknowledgements and non-acknowledgements. In essence, classically, the ``networking task" was to signal the success or failure in the reception of a particular message (or a part of a message). Meanwhile, in semantic communication systems, the mere reception of a message does not have the same significance anymore. This is why, interventions and countefactuals should be key features of the semantic-based control messages. That is, the goal of semantic networking relies on: a) confirming the ``understanding" of a semantic representation, and b) the ability to build or compile a semantic language with a similar view-point as the teacher. \\
\indent Moreover, in current 5G  cellular systems, the control and user planes are separated as shown in Fig.~\ref{fig:semantic_planes}. This has been designed to enable a high flexibility and interoperability. On the one hand, the control plane can interact with multiple user planes. On the other hand, the user plane function can be shared by multiple control plane functions. This separation is also aligned with the \ac{O-RAN}'s initiative to have an \emph{open, intelligent, virtualized and fully interoperable RAN}~\cite{gavrilovska2020cloud}. In contrast, in semantic cellular systems, and given the full seamless convergence of computing and communication resources, a novel \emph{reasoning} plane is in operation. This reasoning plane is separated from the other planes, yet is \emph{sandwiched} between the control and user plane. The main functions of this reasoning plane are centered around:
\begin{enumerate}
    \item Deliver the user plane the extracted semantic representations that correspond to the source message.
    \item Tune the control messages so as to enable building a common semantic language between the teacher and the apprentice, and asymptotically reaching a symmetry between the teacher and the apprentice.
\end{enumerate}
In other words, the reasoning plane will act as a master to guide the user and control plane. The operation of these two planes would be further reinforced with intent and enable reaching the overall system goals more efficiently. Based on these insights we can make the following observations on the evolution of the physical layer and networking layer in semantic networks:
\begin{itemize}
    \item \emph{On the Physical Layer:} We can clearly see that semantic communications is not a substitute physical layer. In essence, semantic communications re-engineers the physical layer to be viewed as the language medium, rather than the bit-to-symbol mappers. 
    \item \emph{On Data-link and Networking Functions:} Networking functions like multiple access, multiplexing, resource allocation, and scheduling can conserve its current classical structure in semantic networks. Nonetheless, to improve the efficiency of the overall system, one can implement them via queries (interventions or counterfactuals) or transmit them via a semantic language. That is, if end nodes have reached a higher level of maturity in terms of the language used, the reasoning plane will require the user plane and control plane the semantic representations \emph{that describe the next multiple access command.} This also improves the radio nodes intelligence and autonomy in making control and networking decisions. Moreover, novel semantic based networking functions can enhance the performance of time-critical communication. For instance, association, beamforming, and scheduling can be tuned based on the needs of the users as well as the changes in the environment that can be concluded from the radio nodes' knowledge bases. Essentially, all of these protocols can now be revisited with the lense of the presence of semantics, reasoning faculties, and other intrinsic features of a reasoning-driven, AI-native semantic wireless system.
\end{itemize}
\indent \\$\bullet$ \textbf{On Reasoning and Computing-based Optimization:}
In Section~\ref{co-evolution}, we have highlighted that the paucity of computing resources may restrain the potential for future semantic communication networks. In fact, similarly to the chase of bandwidth that was observed in the migration from 4G to 5G, the evolution of semantic communications will be governed by a \emph{chase behind computing}. Hence, optimizing computing resources, orchestrating them, and distributing them in an efficient way is an important practice of future semantic communications in 6G and beyond. The following key open-problems can be investigated:
\begin{itemize}
    \item \emph{Ubiquitous Distributed Computing:} In case the network has a high heterogeneity of radio nodes (with respect to the computing capability, e.g., \ac{IoT} sensor communicating with digital twin), one can consider distributing the computing resources. That is, to ensure that reasoning-based services can benefit from a semantic language, one can distribute the computing resources around local low-power devices. For example, if a device cannot be augmented with more computing, one can move part of the reasoning mechanism to the edge (of the associated access point or base station, instead of the end-device). Hence, from the \ac{IoT} sensor's perspective, the communication is being carried out classically, since that link has remained intact. Meanwhile, from the edge, all the way to the digital twin, a semantic-based communication scheme is performed on that end of the link. This enables achieving fairness in the network despite the existing heterogeneity in \acp{UE}. Clearly, optimizing the distribution of computing resources for semantic communication systems is an open problem that can be carried out using distributed \ac{AI} mechanisms or game theory \cite{chen2019artificial}.
    \item \emph{Competitive and Cooperative Languages:} Throughout this tutorial, we have carried out the task of ``building and learning" the language from the teacher and apprentice's side in a fully \emph{cooperative} way. Nonetheless, when the setting expands to multiple radio nodes, the goal of each might be distinct. Consequently, this might result in a noncooperative mechanism in building the language. Also, a subset of the radio nodes might share some common system goals, and thus might have consensus on part of the language. To examine all of these considerations, one can resort to game theory~\cite{han2012game}, in general, and \emph{hypergame theory} \cite{kovach2015hypergame}, in particular. Essentially, hypergame theory is the confluence of game theory and decision theory, and it provides a set of tools that can be used to characterize the interaction between different radio nodes so as to model: a) The final ``common" language built with respect to all nodes, b) The exact and specific view-point of every radio node with respect to the language, and c) The evolution of the ``emergent" language versus the system goals of each node. Here, many other game-theoretic tools can also be explored, such as the use of signaling games and their extension (see our key results in this space \cite{thomas2022neuro}), referential games, and even a more complex mix of cooperative and noncooperative games. Indeed, this is an essential area of research to guarantee the successful evolution of an emergent language, as well as the scalability of semantic communications over a large scale network.
\end{itemize}
\subsubsection{Semantic Communications in ORAN}
Recently, there has been a concerted global effort from academia, industry, and governmental agencies towards the principles of \emph{openness} and \emph{intelligence} \cite{niknam2020intelligent}. As a result, the \ac{O-RAN} Alliance was formally defined to reach these goals. The main objective of this alliance is to move cellar networks architectures towards disaggregated, intelligent, virtualized, and fully interoperable RAN. We can see on the left hand side of Fig.~\ref{fig:oran_architecture} the current proposed architecture of \ac{O-RAN} for 5G systems. Remarkably, the \ac{O-RAN} architecture is notorious for its \ac{RIC} which consists of two main units \cite{ericssonoran}: 
\begin{itemize}
    \item \emph{The non-\ac{RT}-\ac{RIC}} which operates in the orchestration and management plane and is used to run \ac{AI} tasks that can tolerate a considerable execution time. For example, network benchmarking, \ac{AI}/\ac{ML} lifecycle management, and orchestration. 
    \item \emph{The near-\ac{RT}-\ac{RIC}} which operates in the applications layer and can be used for tasks that require a faster decision making (e.g. handover decisions, \ac{QoS} control, and load balancing).
\end{itemize}  
In fact, the ``virtualized and intelligent" aspects of \ac{O-RAN} have already been poised for a convergence in the computing and communication resources, however this was observed in the upper layers: applications layer and orchestration and management layer. With semantic communications, this convergence will become seamless and take place in every layer of the network stack. As a result of the introduction of the reasoning plane, we can see in Fig.~\ref{fig:oran_architecture} that novel \emph{real-time} \ac{AI}-oriented blocks have been proposed in the \ac{O-RU}, \ac{O-DU}, and \ac{CU} layers. Moreover, we can see three planes on the \ac{CU} layer, this is where the majority of the interactions take place between the control, reasoning, and user plane we have previously highlighted in Fig.~\ref{fig:semantic_planes}. We can further make the following observations with regards to the 6G and Beyond \ac{AI}-Native \ac{O-RAN} architecture:
\begin{itemize}
    \item Semantic communications not only is the path for \ac{AI}-nativeness, but it is also a driving force that aligns with the initiative of \ac{O-RAN}. That is, with semantic communications, the intelligence chased will become operable in every aspect of the network and in three time-scales: a) real-time, b) near-real time, and c) non-real time.
    \item The reasoning plane introduced by semantic communications will slightly modify the non-\ac{RT} \ac{RIC} and \ac{RT}-\ac{RIC} functions, whereby their operation will be more intent-based. This intent will be formulated as a feedback from the reasoning plane and results from: a) the context of information, b) the overall system goal. For instance, the xApps (applications running on the near-\ac{RT} \ac{RIC}) and rApps (applications running on the non-\ac{RT} \ac{RIC}) can benefit from the semantic content elements identified to make better \ac{AI} decisions. 
    \item Manual requests that are still driven by the network operator can be automated and inferred from the identified root-causes of semantic content elements. 
\end{itemize}
\indent Henceforth, in this section, we have investigated the key techniques that enable a nascent teachers to get accustomed to a language, then we have highlighted the use-cases that will prospectively be the early adopters of semantic communications, as well as the key facilitators that enable expediting the maturity of semantic systems. Subsequently, we have investigated the networking and computing considerations that must take place to expand semantic communications to large scale networks. Finally, we have emphasized the role of semantic communications in \ac{O-RAN}.
\section{Conclusions and Recommendations}\label{Sec:Conclusion}
Semantic communications can potentially revolutionize the wireless industry, and provide a fundamentally novel way to design and operate communication systems. However, as discussed in this tutorial, in order to reap the real benefits of semantic communications, it is necessary not to reduce this area into yet another incremental extension of existing techniques such as source coding, data compression, application-aware scheduling, and natural language processing. In contrast, we advocate for creating new, rigorous mathematical foundations for semantic communication systems that lie at the intersection of \ac{AI}, communication theory, networking, causal reasoning, information theory, transfer learning, and minimum description length theory. In particular, we have identified the five main tenet of semantic communication systems that include: a) minimally sufficient representations, b) semantic language, c) reasoning via causality, d) semantic-based \acp{KPI}, and e) judicious use of computing resources. We have then developed a comprehensive roadmap towards designing these pillars and building next-generation semantic communication networks that are grounded on rigorous, concrete, and flexible knowledge-driven AI frameworks. In doing so, we have also showed that there is a need to revisit the fundamentals of information theory, and extend it to a semantic information theory that hinges on a \emph{minimal, generalizable, and efficient} semantic language which ensures a symmetrical communication. Through the proposed frameworks, concepts, and vision, this tutorial provides, for the first time, holistic and technically-grounded answers to the following key questions:
\begin{itemize}
    \item What is a semantic communication system, and how is it different from what we already know?
    \item What are the fundamental building blocks of a semantic communication system?
    \item How do we build the reasoning faculty of a semantic system, and how does it communicate via a \emph{minimal, generalizable, and efficient} semantic language?
    \item How do we evaluate the performance of semantic communication networks, and what are the major influencers of the performance compared to classical communications?
    \item How do we expand semantic communications to existing and future large scale wireless networks?
\end{itemize}
In light of our panoramic investigation, we conclude with several \emph{recommendations} to ensure a proper deployment of future semantic communication networks:  
\begin{enumerate}
    \item \textbf{Semantic communications is beyond goal-oriented communications:} We acknowledge that goal-oriented communication is fundamental, particularly for use-cases with common radio nodes attempting to achieve a common goal. Also, we acknowledge that such use-cases might be the first to benefit from the concept of semantic communications. Nonetheless, semantic communications is beyond goal-oriented communications and is the path to creating a fundamentally novel type of networks that we called reasoning-driven \ac{AI}-native wireless networks. These new reasoning-driven, \ac{AI}-native systems will be able to cater to the complex requirements of services like the metaverse, \ac{XR}, and the internet of senses.
    \item \textbf{Advances in AI and computing:} Indeed there are many wireless communication challenges in deploying semantic communications. Nonetheless, given that reasoning is the central pillar of semantic communications, various advances and developments must occur in \ac{AI} so as to develop radio nodes that can build comprehensive and organized knowledge bases. Also, further computing advances are needed so that logical conclusions performed by radio nodes can meet the time-critical needs of beyond 6G applications.
    \item \textbf{On the relationship between semantic communication systems and classical communications:} In many recent works, the concept of semantic communication systems has been touted as the ultimate and only replacement of classical communication systems, and the solution to every wireless communication challenge of the next decade. While we agree that reasoning-driven, AI-native semantic communication systems, if built correctly, can fundamentally change the way we design wireless networks, the reality is that semantic networks and classical networks will have to co-exist and work hand-in-hand. As we outlined in Section~\ref{section:howto}, raw datastream is not entirely structured into semantic content elements and contains a lot of random information. Such random information are \emph{better memorized than learned}, and thus must be transmitted via classical communication channels. Meanwhile, learnable data will be the key input that will be transformed into a semantic language. Therefore, we recommend that research in this space heed and acknowledge the differences between a) the important short-term needs of wireless systems (e.g., better managing \ac{mmWave} and \ac{THz} links, enhancing classical reliability, taming the \ac{E2E} communication latency), b) a medium-term milestone during which both classical and reasoning-driven semantic networks will harmoniously coexist, serving different applications and use cases, and c) the longer-term vision of pure reasoning-driven \ac{AI} systems in which the majority of radio nodes will be able to leverage their accumulated and organized knowledge base to perform versatile and logical decisions across the networking stack. For the next decade, research along all three lines must concurrently take place in order for us to truly usher in a revolution in wireless systems
    \item \textbf{Less spectrum use through convergence of computing and communication:} The deployment of semantic communications, which will be crowning of the convergence of communications and computing, will help alleviate the technical and regulatory burdens associated with the need to open up new spectrum bands each time a new wireless cellular system generation must be deployed or a new use case of the spectrum emerges. Therefore, it is necessary for the community to further think about this convergence and its implications to current and future spectrum-related roadmaps and challenges.
\end{enumerate}
In a nutshell, by answering these questions and concretely laying the foundations of semantic communication networks through a unified and systematic treatment of the underlying challenges, this tutorial is poised to become a primary reference in this burgeoning field.

\bibliographystyle{IEEEtran}
\def\baselinestretch{0.80}
\bibliography{bibliography}
\end{document}